\documentclass[draft]{agujournal2019}
\usepackage{url} 
\usepackage{lineno}
\usepackage[inline]{trackchanges} 
\usepackage{soul}

\usepackage{amsmath}
\usepackage{amssymb}

\usepackage{mathtools} 

\usepackage{bm}
\usepackage{bbm}        
\usepackage{amsmath}    
\usepackage{mathtools}  

\newcommand{\state}{\mathbf{v}}
\newcommand{\State}{\mathbf{V}}
\newcommand{\hidden}{\mathbf{r}}
\newcommand{\Hidden}{\mathbf{R}}
\newcommand{\inputstate}{\mathbf{u}}
\newcommand{\hiddenlayer}{f}
\newcommand{\outputlayer}{g}

\newcommand{\hyperparameters}{\bm{\theta}}
\newcommand{\esnparams}{\hyperparameters_{ESN}}
\newcommand{\nvarparams}{\hyperparameters_{NVAR}}
\newcommand{\Wout}{\mathbf{W}_\text{out}}
\newcommand{\localWout}{\Wout^{k}}
\newcommand{\cf}{\mathcal{J}}
\newcommand{\macrocost}{\cf_\text{macro}}
\newcommand{\tikhonov}{\beta}

\DeclareMathOperator*{\argmin}{argmin}

\newcommand{\leak}{\alpha}
\newcommand{\adjacency}{\mathbf{A}}
\newcommand{\inputmatrix}{\mathbf{W}_\text{in}}
\newcommand{\inputscaling}{\sigma}
\newcommand{\bias}{\sigma_{b}}
\newcommand{\biasvector}{\mathbf{b}}
\newcommand{\biasvectorelement}{b}

\newcommand{\spectralradius}{\rho}

\newcommand{\maxpolynomial}{p}
\newcommand{\maxlag}{N_\text{lag}}

\newcommand{\localstate}{\state_{k}}
\newcommand{\localhidden}{\hidden_{k}}
\newcommand{\localinputstate}{\inputstate_{k}}
\newcommand{\localoutput}{\hat{\state}_{k}}

\newcommand{\nstate}{N_{\state}}
\newcommand{\nhidden}{N_{\hidden}}
\newcommand{\ninputstate}{N_{\inputstate}}
\newcommand{\ntrain}{N_\text{train}}
\newcommand{\ntime}{N_\text{time}}

\newcommand{\ngroups}{N_{g}}
\newcommand{\nx}{N_{x}}
\newcommand{\ny}{N_{y}}
\newcommand{\nvertical}{N_{z}}
\newcommand{\noverlap}{N_{o}}
\newcommand{\nlocalstate}{\nstate^\text{loc}}
\newcommand{\nlocalinputstate}{\ninputstate^\text{loc}}
\newcommand{\nlocalx}{\nx^\text{loc}}
\newcommand{\nlocaly}{\ny^\text{loc}}
\newcommand{\nneighbor}{N_{b}}
\newcommand{\nsub}{N_\text{sub}}
\newcommand{\nlag}{\maxlag}
\newcommand{\nk}{N_K}
\newcommand{\nmacro}{N_\text{macro}}

\newcommand{\statespace}{\mathbb{R}^{\nstate}}
\newcommand{\hiddenspace}{\mathbb{R}^{\nhidden}}
\newcommand{\Woutspace}{\mathbb{R}^{\nstate\times\nhidden}}

\newcommand{\norm}[1]{\left\lVert#1\right\rVert}
\usepackage[capitalise,noabbrev]{cleveref}

\crefname{appendix}{}{} 

\newcommand{\citep}{\cite}
\newcommand{\citet}{\citeA}

\draftfalse
\journalname{Journal of Advances in Modeling Earth Systems (JAMES)}

\begin{document}

\title{Temporal Subsampling Diminishes Small Spatial Scales in
    Recurrent Neural Network Emulators of
    Geophysical Turbulence}

%
%


\authors{
Timothy A. Smith\affil{1,2},
Stephen G. Penny\affil{1,3},
Jason A. Platt\affil{4},
Tse-Chun Chen\affil{5}
}
\affiliation{1}{Cooperative Institute for Research in Environmental Sciences
    (CIRES) at the University of Colorado Boulder, Boulder, CO, USA
}
\affiliation{2}{Physical Sciences Laboratory (PSL), National Oceanic and
    Atmospheric Administration (NOAA), Boulder, CO, USA
}
\affiliation{3}{Sofar Ocean, San Francisco, CA, USA}
\affiliation{4}{University of California San Diego (UCSD), La Jolla, CA, USA}
\affiliation{5}{Pacific Northwest National Laboratory, Richland, WA, USA }

\correspondingauthor{Timothy A. Smith}{tim.smith@noaa.gov}

\begin{keypoints}
    \item Reducing training data temporal resolution by subsampling leads to
        overly dissipative small spatial scales in neural network
        emulators
    \item A quadratic autoregressive architecture is shown to be inadequate at capturing
        small scale turbulence, even when data are not subsampled
    \item Subsampling bias in Echo State Networks is mitigated but not
        eliminated by prioritizing kinetic energy spectrum during training
\end{keypoints}

%
%

\begin{abstract}
    The immense computational cost of traditional numerical weather and climate
    models has sparked the development of machine learning (ML) based emulators.
    Because ML methods benefit from long records of training data,
    it is common to use datasets that are temporally subsampled relative to the
    time steps required for the numerical integration of differential equations.
    Here, we investigate how this often overlooked processing step affects
    the quality of an emulator's predictions.
    We implement two ML architectures
    from a class of methods called reservoir computing: (1) a form of Nonlinear
    Vector Autoregression (NVAR), and (2) an Echo State Network (ESN).
    Despite their simplicity, it is well documented that these architectures
    excel at predicting low dimensional chaotic dynamics.
    We are therefore
    motivated to test these architectures in an idealized setting of predicting
    high dimensional geophysical turbulence as represented by Surface
    Quasi-Geostrophic dynamics.
    In all cases, subsampling the training data consistently leads
    to an increased bias at small spatial scales that resembles numerical diffusion.
    Interestingly, the NVAR architecture becomes unstable when the temporal
    resolution is increased, indicating that the polynomial based interactions
    are insufficient at capturing the detailed nonlinearities of the turbulent
    flow.
    The ESN architecture is found to be more robust, suggesting a benefit to the
    more expensive but more general structure.
    Spectral errors are reduced by including a penalty on the
    kinetic energy density spectrum during training, although the subsampling
    related errors persist.
    Future work is warranted to understand how the temporal resolution
    of training data affects other ML architectures.
\end{abstract}

\section*{Plain Language Summary}

The computer models that govern weather prediction and climate projections
are extremely costly to run, causing practitioners to make unfortunate
tradeoffs between accuracy of the physics and credibility of their statistics.
Recent advances in machine learning have sparked the development of neural
network-based emulators, i.e., low-cost models that can be used as drop-in
replacements for the traditional expensive models. Due to the cost of storing
large weather and climate datasets, it is common to subsample these fields in
time to save disk space.
This subsampling also reduces the computational expense of training emulators.
Here, we show that this pre-processing step hinders the fidelity of the emulator. We
offer one method to mitigate the resulting errors, but we suggest that more
research is needed to understand and eventually overcome them.

%
%

\section{Introduction}
\label{sec:intro}

Weather and climate prediction requires the numerical integration of one or more computational models derived from the fundamental equations of motion and initialized
with an estimate of the present-day system state (e.g., temperature, wind speeds,
etc.).
Due to the high cost of these computational models, prediction systems
typically require suboptimal tradeoffs.
On one hand, it is desirable to increase the credibility of the underlying
numerical model as much as possible,
for instance by increasing model grid resolution
\citep <e.g.,>[]{hewitt_impact_2016}
or by explicitly
simulating as many coupled components (e.g., atmosphere, land, ocean, ice) as
possible \citep <e.g.,>[]{penny_coupled_2017}.
On the other hand, knowledge of the model initial conditions is imperfect and
the governing equations will always contain necessary, inexact approximations of
reality.
As a result, prediction systems employ statistical methods like ensemble based forecasting in order to represent this
uncertainty.
Producing an ensemble with statistical significance
requires integrating the underlying numerical model many times; usually
$\mathcal{O}(10) - \mathcal{O}(100)$ in practice, but ideally $\mathcal{O}(1000)$ or greater
\citep{evensen_data_2022}.
Therefore, the resulting computational costs require practitioners to compromise between the
fidelity of the numerical model and credibility of the statistical method.

An ongoing area of research that aims to enable statistical forecasting subject
to the dynamics of an expensive numerical model is \textit{surrogate modeling}.
The general approach relies on using a model that represents or ``emulates'' the
dynamics of the original numerical model with ``sufficient accuracy'' for the given
application, while being computationally inexpensive to evaluate.
Historically, surrogate models have been an important tool for nonlinear
optimization \citep<e.g.,>[]{li_data-based_2019,bouhlel_scalable_2020},
and in the Earth sciences have been developed with techniques such as
Linear Inverse Models
\citep <e.g.,
principal oscillation or interaction patterns;>[]{hasselmann_pips_1988,penland_random_1989, moore_linear_2022},
kriging \citep{cressie_statistics_1993},
or
polynomial chaos techniques \citep{najm_uncertainty_2009},
to name only a few.
More recently, advances in computing power, the rise of general purpose graphics processing units,
and the explosion of freely available data
has encouraged the exploration of more expensive machine learning methods like neural networks for the emulation task \citep{schultz_can_2021}.
A number of data-driven, neural network architectures have been developed to
generate surrogate models for weather forecasting and climate projection applications.
Some examples include models based on feed forward neural networks
\citep{dueben_challenges_2018},
convolutional neural networks
\citep <CNNs;>[]{scher_toward_2018,scher_weather_2019,rasp_data-driven_2021,weyn_can_2019,weyn_improving_2020,weyn_sub-seasonal_2021},
recurrent neural networks
\citep <RNNs;>[]{arcomano_machine_2020,chen_predicting_2021,nadiga_reservoir_2021},
graph neural networks \citep{keisler_forecasting_2022,lam_graphcast_2022},
Fourier neural operators \citep{pathak_fourcastnet_2022},
and encoder-transformer-decoder networks
\citep{bi_accurate_2023}.

A significant advancement in surrogate modeling for weather and climate prediction
has been the rapid increase in spatial resolution.
To the best of our knowledge, the current highest resolution neural network
emulators for
global atmospheric dynamics is $\sim0.25^\circ$ ($\sim$31~km)
\citep{pathak_fourcastnet_2022,bi_accurate_2023,lam_graphcast_2022},
which is the same
resolution as the ERA5 Reanalysis \citep{hersbach_era5_2020} used to train these models.
At this resolution, General Circulation Models (GCMs) of the atmosphere are
capable of explicitly capturing important small scale processes like low-level
jets and interactions with mountainous topography \citep{orlanski_rational_1975}.
However, it is not yet clear that neural networks are able to
represent the same dynamical processes as the training data.
Instead, based on our own experimentation, we hypothesize that without careful
architectural modifications, neural network emulators will effectively operate at a coarser
resolution than the original dataset used in training.

\begin{figure}
    \centering
    \includegraphics[width=.8\textwidth]{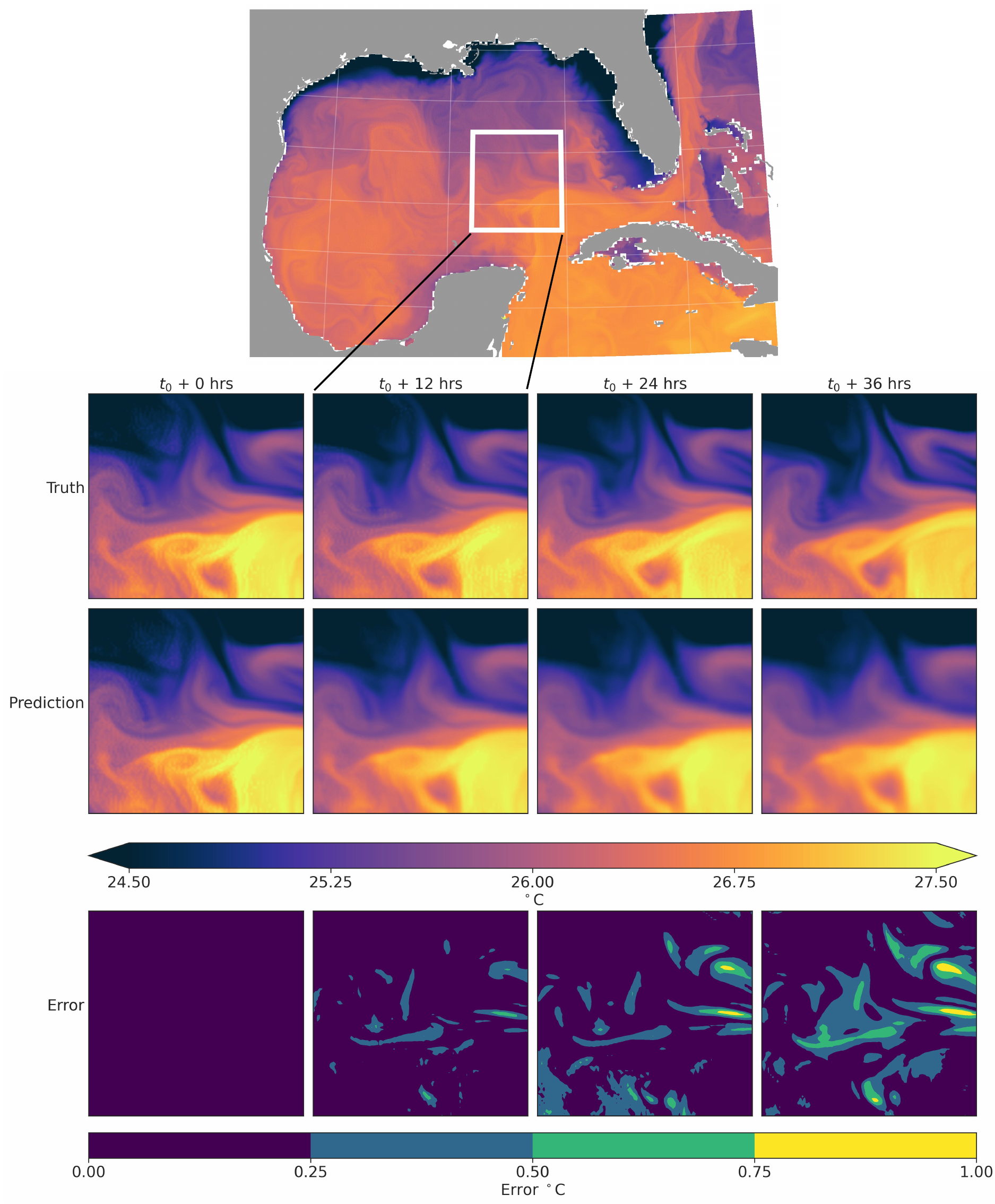}
    \caption{A sample prediction of sea surface temperatures in the Gulf of Mexico at 1/25$^\circ$
        horizontal resolution.
        The upper row (Truth) shows the evolution of unseen test data from the
        Navy HYCOM reanalysis product, and the middle row shows a prediction
        from the Echo State Network architecture described in
        \cref{subsec:rc}.
        The bottom row (Error) shows the absolute value of the difference between the two.
        See \cref{sec:gom} for a description of the dataset.
    }
    \label{fig:gom_sst}
\end{figure}

To make the discussion concrete, we present a sample prediction from our own surrogate model
in \cref{fig:gom_sst}.
The panels show the time evolution of Sea Surface Temperature (SST) in the Gulf of Mexico at 1/25$^\circ$ horizontal resolution, using data from a Navy HYCOM, 3D-Var-based reanalysis product as ``Truth'' (upper row; see \cref{sec:gom} for data details).
We generate the prediction (middle row) with an RNN architecture
described more fully in \cref{subsec:rc}.
Generally speaking, the RNN captures the largest scales of the SST pattern over
a 36~hour window.
However, as time progresses, the SST pattern becomes overly smooth.
The RNN is unable to capture the spatial details that are well resolved in the reanalysis
dataset, with the largest errors evolving along sharp SST fronts.
We note that a similar smoothing behavior can be observed in other neural network based emulators, see for example
\citep<>[Figure 3]{bi_accurate_2023},
\citep<>[Figure 4c \& 4d]{pathak_fourcastnet_2022},
\citep<>[Figure 5]{keisler_forecasting_2022}.

There are a number of reasons that could cause this smoothing behavior to manifest in the
predictions.
As we show in \cref{sec:nvar-results,sec:esn-results}, the blurring of small scale features is
a high frequency spectral bias, which has been studied in relation to the
training of feedforward neural networks \citep{xu_overview_2022} and numerical
instabilities of neural network predictions for turbulent flows
\citep{chattopadhyay_long-term_2023}.
One potential reason that we observe spectral bias in our predictions is that
the training uses a mean-squared error loss function, which is known to prioritize large over
small scale features \citep{rossa_overview_2008}.
Here, we suggest that any blurring effect from such a loss function is exacerbated by more fundamental decisions in the experimental design.
Our primary goal is to explore how temporal subsampling in the training dataset adds to this blurring effect.
We are motivated to study the impact of this subsampling because many existing emulators, including our example in \cref{fig:gom_sst}, rely on reanalysis products as training data
\citep<e.g.>[]{lam_graphcast_2022,bi_accurate_2023,pathak_fourcastnet_2022,keisler_forecasting_2022,weyn_sub-seasonal_2021,arcomano_machine_2020}.
While there are excellent reasons to leverage the existence of reanalysis products, namely that they are constrained to observational data, the shear size of the data requires some
degree of temporal subsampling.
We suggest that it is important to understand how this highly routine data reduction step impacts the performance of data-driven prediction methods when used for training.

In our work, we explore the degree to which temporal subsampling impedes single
layer autogregressive and recurrent neural network emulators from learning the true underlying
dynamics of the system.
In order to isolate this effect from the potential impacts of a data
assimilation system and multivariate interactions,  we do not rely on the Gulf
of Mexico reanalysis data.
Instead, we use a model for Surface Quasi-Geostrophic (SQG) turbulence
\citep{held_surface_1995,blumen_uniform_1978}, which additionally
gives us direct control over the datasets used for training, validation, and testing.
The SQG model and dataset generation is described more fully in
\cref{sec:sqg}.

The architectures that we use in this study stem from a broad class of machine learning techniques termed as reservoir computing (RC), which was independently discovered as
Echo State Networks \citep<ESNs;>[]{jaeger_echo_2001},
Liquid State Machines \citep{maass_real-time_2002},
and the Decorrelation Backpropagation Rule
\citep{steil_backpropagation-decorrelation_2004}.
One defining characteristic of RC models is that all internal connections are adjusted by global or ``macro-scale'' parameters, significantly reducing the
number of parameters that need to be trained.
The relatively simplified structure and training requirements of RC make it an
attractive architecture for large scale prediction because it enables rapid development, and could be useful in situations requiring online learning.
More importantly though, we are motivated to use RC because past studies have repeatedly shown that it can emulate low dimensional chaotic systems while often outperforming more complex RNNs such as those with Long Short-Term Memory units (LSTMs)
\citep<e.g.>[]{platt_systematic_2022,vlachas_backpropagation_2020,griffith_forecasting_2019,lu_attractor_2018,pathak_model-free_2018}.
Additionally, \citet{penny_integrating_2022} showed that RC can be
successfully integrated with a number of data assimilation algorithms, either
by generating samples for ensemble based methods like the Ensemble Kalman Filter,
or by generating the tangent linear model necessary for 4D-Var.
Finally, we note that \citet{gauthier_next_2021} proposed a further
simplification to the RC architecture based on insights from
\citet{bollt_explaining_2021} that unifies versions of RC with nonlinear vector autoregression (NVAR).
For a variety of chaotic systems, this architecture has shown excellent prediction skill
even with low order, polynomial-based feature vectors
\citep{chen_next_2022,barbosa_learning_2022,gauthier_next_2021}, despite requiring a much
smaller hidden state and less training data.
Considering all of these advancements, we are motivated to use these simple yet
powerful single layer NVAR and ESN architectures to emulate turbulent
geophysical fluid dynamics, and study how they are affected by temporal subsampling
(see \cref{sec:rnn-architecture} for architecture details).

\section{Surface Quasi-Geostrophic Turbulence}
\label{sec:sqg}

\begin{figure}
    \centering
    \includegraphics[width=\textwidth]{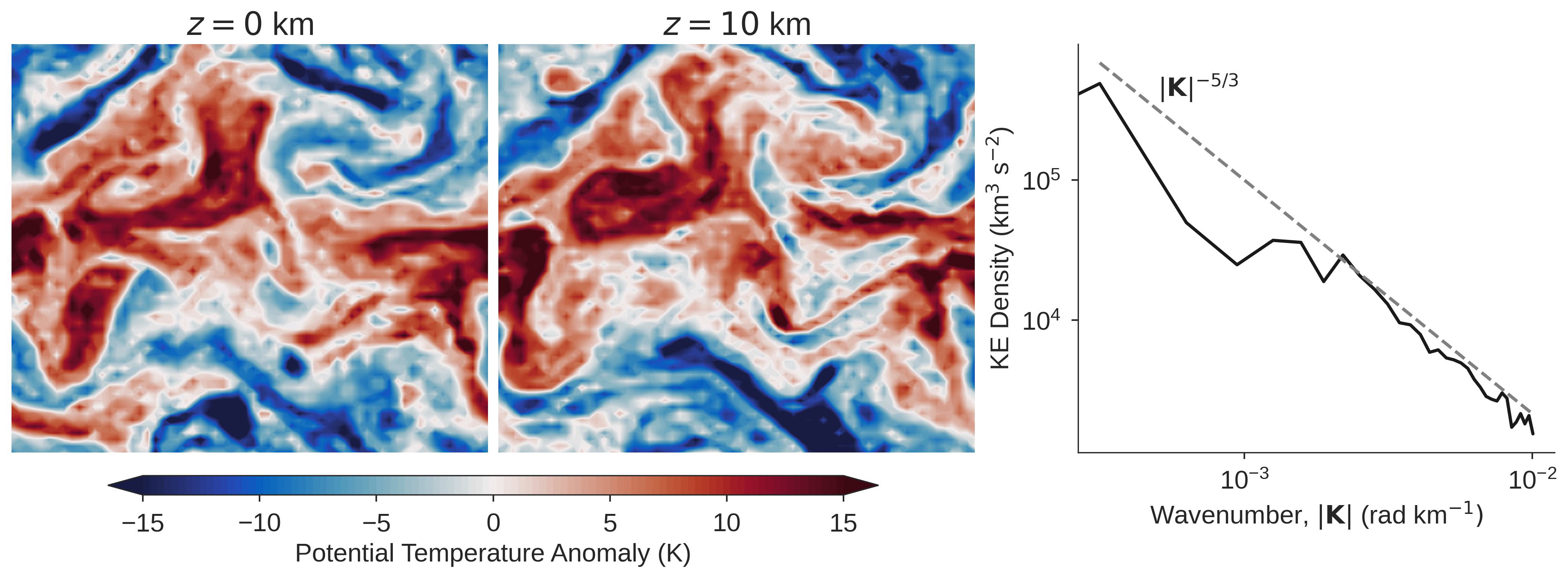}
    \caption{A reference snapshot from the SQG dataset. The left and middle panels
        show snapshots of potential temperature anomaly at the surface and
        top-of-troposphere layers, respectively.
        The right panel shows the kinetic energy density spectrum associated
        with this snapshot (black line), compared to
        $|\mathbf{K}|^{-5/3}$ (dashed line).
    }
    \label{fig:sqg-reference}
\end{figure}

Our goal in this study is to emulate turbulent motions relevant to realistic
geophysical fluid dynamics, while avoiding the complications associated with the
data assimilation system used to produce reanalysis datasets, including
observational noise and error covariance estimates, and the intricate
multivariate interactions inside atmosphere or ocean GCMs.
Therefore, we aim to emulate a numerical model for SQG turbulence
\citep{held_surface_1995,blumen_uniform_1978}
as outlined by \citet{tulloch_note_2009}.
The model is formulated to represent the nonlinear Eady problem
\citep{eady_long_1949}, following \citet{blumen_uniform_1978-1}.
The model simulates turbulence
on an $f$ plane with uniform stratification and shear, bounded by rigid surfaces $H=10$~km apart.
The motion is determined entirely by temperature advection on the boundaries
$z=\{0\text{~km},10\text{~km}\}$ as follows,
\begin{linenomath*}\begin{equation*}
    \dfrac{\partial \hat{\theta}}{\partial t} +
    \hat{J}(\hat{\psi}, \hat{\theta}) + ik\left(U \hat{\theta} +
        \hat{\psi}\dfrac{\partial \Theta}{\partial y}\right)
    = 0 \qquad z = 0, 10\,\text{km} \, ,
\end{equation*}\end{linenomath*}
where $z=0$~km is the surface layer of the atmosphere, and $z=10$~km is
approximately at the top of the troposphere.
Here, hatted variables denote spectral components, $\hat{J}$ is the Jacobian in spectral space, and the temperature streamfunction is
\begin{linenomath*}\begin{equation*}
    \hat{\psi}(z,t) = \dfrac{H}{\mu\sinh\mu}
    \left[ \cosh\left(\mu\dfrac{z}{H}\right) \hat{\theta}(H,t)
        - \cosh\left(\mu\dfrac{z-H}{H}\right) \hat{\theta}(0,t)
    \right]\, ,
\end{equation*}\end{linenomath*}
with $\mu = |\mathbf{K}| NH/f$ as the nondimensional wavenumber.
We note that this model produces an approximate spectrum of
$|\mathbf{K}|^{-5/3}$ without any break (\cref{fig:sqg-reference}),
as is expected in Eady turbulence.
For more details on this model, see \citet{tulloch_note_2009}.

Our model configuration is discretized in space with $N_x = N_y = 64$ and $\nvertical=2$,
uses a periodic boundary in both horizontal directions,
and uses a timestep of $\Delta t=5$~minutes.
To generate datasets for the neural networks, we initialize the model with Gaussian i.i.d.
noise and spinup for 360~days, which we define as one model year.
The spinup period is discarded, and we then generate a 25~year dataset that we partition
into training (first 15~years), validation (next 5~years), and testing
(final 5~years).
For validation and testing, we randomly select 12~hour time windows from each
respective dataset.

\section{Single Layer Autoregressive and Recurrent Neural Networks}
\label{sec:rnn-architecture}

Our goal is to develop an emulator that can reproduce the time evolution of a chaotic dynamical system, such that its future state can be predicted from an initial
state estimate.
Therefore we use the following generic, discrete-time equations for our recurrent and autoregressive models,
\begin{linenomath*}\begin{equation}
    \begin{aligned}
        \hidden(n+1) &= \hiddenlayer\left(
            \hidden(n), \state(n); \hyperparameters
            \right) \\
        \hat{\state}(n+1) &= \outputlayer \left( \hidden(n+1) \right) \, ,
    \end{aligned}
    \label{eq:rnn}
\end{equation}\end{linenomath*}
as by \citet{goodfellow_sequence_2016}.
Here $n\in\mathbb{Z}$ denotes a particular timestep $t = n\Delta \tau$, where $\Delta\tau = \nsub\Delta t$ is the timestep size of the neural network,
which may be larger than $\Delta t=5$~minutes, the step size of the original model described in
\cref{sec:sqg}.
Here $\state(n)\in\statespace$ is the state of the dynamical system and $\hidden(n)\in\hiddenspace$ is the hidden or internal state of the network, which is also referred to as the ``reservoir'' in RC or ``feature vector'' in
NVAR.
The generic function $\hiddenlayer(\cdot)$ evolves this hidden state forward in
time subject to the explicit
influence of the current hidden and system states, as well as the macro-scale
parameters $\hyperparameters$.
The output layer,
$\outputlayer(\cdot)$, or ``readout'' operation,
maps the hidden state back to the original state space, giving an
approximation of the target system.

During the training phase, $\state(n)$ is provided to the model at each timestep
and the misfit between the approximation and data,
$\hat{\state}(n+1) - \state(n+1)$, is used to train the weights in the output
layer.
After training, during the prediction phase, the network becomes an autonomous
system:
\begin{linenomath*}\begin{equation*}
    \hidden(n+1) = \hiddenlayer\left(
        \hidden(n), \hat{\state}(n); \hyperparameters\right) \, .
\end{equation*}\end{linenomath*}

The neural network architectures that we use employ a common
structure that is relevant to the readout operator and training
procedure; we discuss these details in \cref{subsec:readout}.
Additionally, we employ a similar strategy to parallelize the architecture for
high dimensional systems, and this is discussed in
\cref{subsec:parallelization}.
Finally, the specific form of $\hiddenlayer(\cdot)$ for the ESN and NVAR architectures
is provided in \cref{subsec:rc,subsec:nvar},
respectively.

\subsection{Linear Readout and Training}
\label{subsec:readout}

The neural networks that we use employ two
simplifications relative to the generic form presented in
\cref{eq:rnn}.
First, any internal relationships encapsulated within
$\hiddenlayer(\cdot)$ are pre-defined by the macro-scale parameters,
$\hyperparameters$.
Therefore, no internal weights contained within $\hiddenlayer(\cdot)$
are learned during the formal training process.
Secondly, the readout operator is linear, such that
\begin{linenomath*}\begin{equation*}
    \outputlayer(\hidden(n)) \coloneqq \Wout \hidden(n) \, ,
\end{equation*}\end{linenomath*}
where $\Wout \in \Woutspace$ is a matrix.
The result of these two assumptions is a cost function that is quadratic with
respect to the elements of $\Wout$,
\begin{linenomath*}\begin{equation}
    \cf(\Wout) =
        \dfrac{1}{2\ntrain}\sum_{n=1}^{\ntrain}
        \norm{\Wout \hidden(n) - \state(n)}^2_2
        +
        \dfrac{\tikhonov}{2}\norm{\Wout}_\text{F}^2 \, .
    \label{eq:cost}
\end{equation}\end{linenomath*}
Here
$\norm{\mathbf{A}}_\text{F} \coloneqq
\sqrt{\text{Tr}\left(\mathbf{A}\mathbf{A}^T\right)}$
is the Frobenius norm,
$\ntrain$ is the number of time steps used for training,
$\tikhonov$ is a Tikhonov regularization parameter \citep{tikhonov_solution_1963}, chosen to improve
numerical stability and prevent overfitting.

The hidden and target states can be expressed in matrix form by concatenating
each time step ``column-wise'':
$\Hidden \coloneqq (\hidden(1) \, \hidden(2) \, \cdots \, \hidden(\ntrain))$,
and similarly\\
\noindent$\State \coloneqq (\state(1) \, \state(2) \, \cdots \, \state(\ntrain))$.
With this notation, the elements of $\Wout$ can be compactly written as the
solution to the linear ridge regression problem
\begin{linenomath*}\begin{equation}
    \Wout = \State \Hidden^T \left(\dfrac{1}{\ntrain}\Hidden\Hidden^T +
    \tikhonov\mathbf{I}\right)^{-1} \, ,
    \label{eq:ridge_regression}
\end{equation}\end{linenomath*}
although we do not form the inverse explicitly.
We instead use the \texttt{solve} function from SciPy's linear algebra module
\citep{scipy_2020}, based on testing shown in
\citet<Appendix C of>[]{platt_systematic_2022}.

\subsection{Parallelization Strategy}
\label{subsec:parallelization}

The model architectures that we use inherit the gridded structure of the target
state being emulated, and often require hidden states that are
$\mathcal{O}(10)$ to $\mathcal{O}(100)$ times larger than the target system dimension.
Atmosphere and ocean GCMs typically propagate high dimensional state vectors, ranging from $\mathcal{O}(10^6)$ to $\mathcal{O}(10^9)$,
so representing the system with a single hidden state would be intractable.
Thus, we employ a parallelization strategy to distribute the target and hidden
states across many semi-independent networks.
Our strategy follows the algorithm introduced by \citet{pathak_model-free_2018},
and follows a similar construction as \citet{arcomano_machine_2020}.
We outline the procedure here and note an illustration of the process for
the ESN architecture in \cref{fig:esn-diagram}.

We subdivide the domain into $\ngroups$ rectangular groups based on horizontal location,
akin to typical domain decomposition techniques for atmosphere and ocean
GCMs on structured grids.
Each group contains
$\nlocalx\times\nlocaly$ horizontal grid cells, and all $\nvertical$
vertical grid cells at each horizontal location.
The global state vector, $\state$, which consists of all state variables to be
emulated at all grid cells, is partitioned into $\ngroups$ local state vectors,
$\localstate$.
For example, \cref{fig:esn-diagram} shows a field $\state$ decomposed into nine
groups, where each group is delineated by white lines.
In our SQG predictions, we set $\nlocalx = \nlocaly = 8$, resulting in
$\ngroups=64$.

In order to facilitate interactions between nearby groups, each group has a designated overlap, or ``halo'', region which consists of $\noverlap$ elements
from its neighboring groups.
The local group and overlapping points are illustrated in \cref{fig:esn-diagram}
with a black box.
The local state vectors, plus elements from the overlap region, are concatenated
to form local input state vectors,
$\localinputstate\in\mathbb{R}^{\nlocalinputstate}$.
The result from the network is the local output state vector,
$\localstate\in\mathbb{R}^{\nlocalstate}$,
which is expanded to fill the target group as illustrated by the
white box on the prediction shown in \cref{fig:esn-diagram}.
Here we set $\noverlap=1$, so that $\nlocalinputstate=200$ and
$\nlocalstate=128$, given that $\nlocalx=\nlocaly=8$ and $\nvertical=2$.

The local input vectors drive separate networks at each group, thereby generating distinct hidden states for each group as follows
\begin{linenomath*}\begin{equation}
    \begin{aligned}
        \localhidden(n+1)
        &= \hiddenlayer\left(
            \localhidden(n), \localinputstate(n); \hyperparameters
        \right) \\
        \localoutput(n+1)
        &= \localWout \localhidden(n+1) \, .
    \end{aligned}
    \label{eq:local-rnn}
\end{equation}\end{linenomath*}
We make the assumption that the macro-scale parameters which determine internal
connections within $\hiddenlayer(\cdot)$ are globally fixed.
Therefore, the only components
that drive unique hidden states in each group are the local input vector
$\localinputstate$ and the local readout matrix, $\localWout$.

During the training phase, each group acts completely independently from one
another.
Therefore, the training process is embarrassingly parallel and allows us to
scale the problem to arbitrarily large state vectors across a distributed
computing system, subject to resource constraints.
During the prediction phase, neighboring elements must be passed between
groups in order to fill each overlap region at each time step with the most accurate state estimate possible, to ensure spatial consistency across the domain.

\subsection{Nonlinear Vector Autoregression Design}
\label{subsec:nvar}

Following \citet{gauthier_next_2021} and \citet{chen_next_2022}, we consider forming the hidden state by using polynomial combinations of the time-lagged input state.
We explain this process with a simple example using a two variable system,
$\inputstate(n) = [u_0(n), u_1(n)]^T$,
a maximum polynomial degree
$\maxpolynomial=2$, and a generic maximum number of lagged states $\maxlag$:
\begin{linenomath*}\begin{equation}
    \begin{aligned}
        \localhidden(n+1)
        =
        [&1, \\
         &u_0(n), \,\, u_1(n), \,\,
        u_0(n-1),\,\, u_1(n-1), \,\,
        \cdots \,\,
        u_0(n-\maxlag), \,\, u_1(n-\maxlag), \\
         &u_0^2(n), \,\, u_1^2(n), \,\, u_0(n)u_1(n), \,\,
        u_0^2(n-1), \,\, \cdots \,\, u_1^2(n-\nlag) \\
         &u_0(n)u_0(n-1), \,\,
        u_0(n)u_1(n-1), \,\, \cdots \,\, u_0(n-\nlag)u_1(n), \,\,\cdots
        ] \\
        \localoutput(n+1) = &\localWout \localhidden(n+1) \, .
    \end{aligned}
\end{equation}\end{linenomath*}
Clearly, the size of the hidden state vector grows rapidly with
$\maxpolynomial$ and $\nlag$,
even for relatively low dimensional systems
\citep<see supplemental material of>[for explicit calculations]{chen_next_2022}.
We therefore make a simplification to the generic polynomial NVAR model.
That is, we only represent nonlinear interactions between points that lie
within a given radius between one another, defined by the number of neighboring
points, $\nneighbor$.
As a simple example, with $\nneighbor=1$ and $\maxlag=0$, the quadratic elements of a periodic, four variable
system would be
\begin{linenomath*}\begin{equation*}
    u_0^2, \,\, u_1^2, \,\, u_2^2, \,\, u_3^2, \,\,
    u_0u_1, \,\, u_0u_3, \,\, u_1u_2, \,\, u_2u_3
\end{equation*}\end{linenomath*}
ignoring ``non-local'' interactions such as $u_0u_2$.
In order to make this parameter consistent with the overlap region in the parallelization scheme (\cref{subsec:parallelization}),
we set $\nneighbor = \noverlap = 1$.
Note, however, that we do model ``non-local'' linear interactions, up to the
number of grid cells in each local group, i.e., containing
$(\nlocalx+2\noverlap)\times(\nlocaly+2\noverlap)\times\nvertical$ points.

All of the remaining macro-scale parameters that determine the NVAR performance are
\begin{linenomath*}\begin{equation*}
    \nvarparams =
    \{ \maxpolynomial, \maxlag, \tikhonov \} \, .
\end{equation*}\end{linenomath*}
By using the preconditioning scheme introduced by \citet{chen_next_2022},
we found results to be insensitive to the Tikhonov parameter $\tikhonov$, and so
we fix this to $\tikhonov = 10^{-4}$.
As noted earlier, we set $\maxpolynomial = 2$.
Our assumption behind this decision is that the NVAR model will be able to learn local
quantities like gradients and fluxes between neighboring grid cells.
Based on the results from \citet{chen_next_2022},
the NVAR model should then be able to use this information to construct
arbitrarily complex time stepping schemes as a function of $\nlag$.
Because of its explicit nature, we manually vary $\nlag$
to understand how memory impacts NVAR prediction skill.

\subsection{Echo State Network Design}
\label{subsec:rc}

\begin{figure}
    \centering
    \includegraphics[width=\textwidth,
    trim={1in 2.4in 1in 2.4in}, clip
    ]{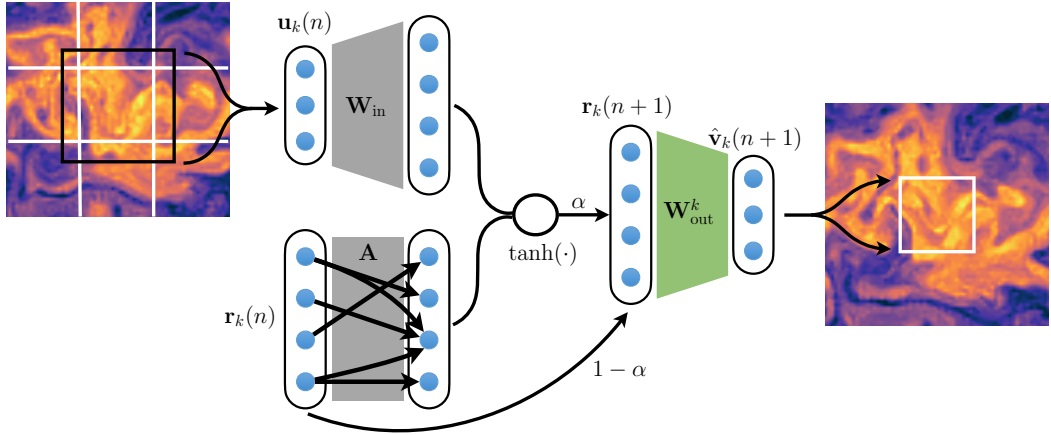}
    \caption{An illustration of the ESN architecture used, as it
        is applied to each local group throughout the domain.
        The domain is decomposed purely based on horizontal location, so the
        illustration shows a single horizontal slice, but note that each group
        contains all $\nvertical$ vertical levels.
        In this example, there are nine groups delineated by the white lines on
        the 2D slice on the left.
        The black box denotes the group being operated on, which includes a
        region of width $\noverlap$ that overlaps with neighboring groups.
        At timestep $n$, the group is flattened to make the input vector
        $\localinputstate(n)$, which is
        mapped into the ESN via $\inputmatrix$.
        The output $\localoutput($$n$$+$$1$$)$ is expanded to fill its position in the global
        domain.
        In ESNs, the matrices $\adjacency$ and $\inputmatrix$ (gray) are fixed, and only
        the readout matrix, $\Wout$ (green), is estimated from the training data.
    }
    \label{fig:esn-diagram}
\end{figure}

Our ESN architecture is illustrated in \cref{fig:esn-diagram}, and is defined as
follows
\begin{linenomath*}\begin{equation}
    \begin{aligned}
        \localhidden(n+1)
        &=
        \left(1-\leak\right)\localhidden(n)
        +
        \leak \tanh\left(
            \adjacency \localhidden(n) + \inputmatrix \localinputstate(n) + \biasvector
            \right)
             \\
        \localoutput(n+1)
        &= \localWout \localhidden(n+1) \, .
    \end{aligned}
    \label{eq:rc}
\end{equation}\end{linenomath*}
Here
$\leak\in[0,1]$ is a leak parameter,
$\adjacency \in \mathbb{R}^{\nhidden\times\nhidden}$ is an adjacency matrix that
determines the internal connections between the nodes of the hidden state,
$\inputmatrix \in \mathbb{R}^{\nhidden\times\ninputstate}$ maps the input vector
into the higher dimensional hidden state,
and $\biasvector\in\mathbb{R}^{\nhidden}$
is the bias vector with elements
$\biasvectorelement_i \sim \mathcal{U}(-\bias,\bias)$.
Unless otherwise specified, each ESN model uses a hidden layer width of
$\nhidden=6,000$.
Finally, we note that ESNs require a spinup period before generating
predictions, so we specify a 10~day spinup period for all validation and testing
samples.

Two scalar parameters, $\spectralradius$ and $\inputscaling$,
are used to control the scaling of the adjacency and input matrices,
respectively.
These parameters have a dramatic influence on ESN prediction skill, since their values influence the network's memory and stability
\citep{lukosevicius_practical_2012,hermans_memory_2010}.
Here we first normalize the matrices by their largest singular value, and then
apply the scaling parameters as follows
\begin{linenomath*}\begin{equation*}
    \adjacency \coloneqq
    \dfrac{\spectralradius}{\sigma_{max}\left(\hat{\adjacency}\right)}
    \hat{\adjacency}
    \qquad
    \inputmatrix \coloneqq
    \dfrac{\sigma}{\sigma_{max}\left(\hat{\mathbf{W}}_\text{in}\right)}
    \hat{\mathbf{W}}_\text{in} \,
\end{equation*}\end{linenomath*}
where the elements of $\hat{\mathbf{W}}_\text{in}$
are initialized with elements $\hat{w}_{i,j}\sim\mathcal{U}(-1,1)$.
The initial adjacency matrix is generated similarly, except that the indices
$i,j$ are randomly chosen such that $\hat{\adjacency}$ attains a specified
sparsity.
Here we set the matrix sparsity to $1 - \kappa / \nhidden$, with $\kappa=6$,
following the success of very sparsely connected adjacency matrices as shown by
\citet{griffith_forecasting_2019}.
By first normalizing the matrices by the largest singular value, the parameters
$\spectralradius$ and $\inputscaling$ re-scale the induced 2-norm of
the matrix.
This normalization is not standard in the ESN literature, but we found that it
helped improve prediction skill.
We provide further discussion of this process in \cref{sec:new_methods}.

In summary, the macro-scale parameters that determine the overall
characteristics of the ESN are

\begin{linenomath*}\begin{equation}
    \esnparams =
    \{ \spectralradius, \inputscaling, \bias, \leak, \tikhonov \} \,
    ,
    \label{eq:rc-hyperparameters}
\end{equation}\end{linenomath*}
which are globally fixed for all groups.
Due to the high sensitivity of ESN prediction skill to these parameter values,
we follow the general optimization framework described by
\citet{platt_systematic_2022} to determine approximately optimal values.
We use the Bayesian Optimization algorithm outlined by
\citet{jones_efficient_1998} and implemented by \citet{bouhlel_python_2019} to tune them.
This process is discussed in \cref{sec:esn-results}.
However, we first focus on prediction skill using the NVAR architecture in
\cref{sec:nvar-results}.

\section{Nonlinear Vector Autoregression Prediction Skill}
\label{sec:nvar-results}

In this section we show the prediction skill of the polynomial based NVAR architecture
described in \cref{subsec:nvar}.
Note that we show the prediction skill of the ESN architecture in
\cref{sec:esn-results}.
To quantitatively evaluate each forecast, we compute
the normalized root-mean-square error (NRMSE)
\begin{linenomath*}\begin{equation}
    \text{NRMSE}(n) = \sqrt{\dfrac{1}{\nstate}\sum_{i=1}^{\nstate}\left(
        \dfrac{\hat{v}_i(n) - v_i(n)}{SD}
        \right)^2 } \, ,
    \label{eq:nrmse}
\end{equation}\end{linenomath*}
which is averaged over each spatial dimension, succinctly represented as a
summation over $\nstate$, and normalized by the standard deviation, $SD$,
computed from the true trajectory over time and all spatial dimensions.
Additionally, we compute the relative error in terms of the kinetic energy
(KE) density spectrum,
\begin{linenomath*}\begin{equation}
    \text{KE Relative Error}(n, k) =
    \dfrac{\hat{E}(n, k) - E(n, k)}{|E(n,k)|} \, ,
    \label{eq:ke_relerr}
\end{equation}\end{linenomath*}
where $E(n,k)$ and $\hat{E}(n,k)$ are the true and predicted KE density coefficients
for each timestep $n$ and wavenumber $k$, respectively (e.g., as in the right
panel of \cref{fig:sqg-reference}).
Note that $|\cdot|$ denotes the absolute value operation,
and we retain the sign of the error in order to show a sense of the
spectral error in each prediction.

We compute these quantities based on 50 twelve-hour predictions initialized from a random
set of initial conditions taken from an unseen test dataset.
To compactly visualize the skill over all samples, each lineplot in the
following subsections shows a sample-average value with a solid line, and the
99\% confidence interval with shading.
We note that in some cases the model trajectory becomes unstable to the point
that infinite values are produced.
In the event that any single sample from a distribution has produced infinity,
we take the more conservative approach and cut off any statistical averaging or
confidence interval computation at that point in time and carry it no further.
Therefore, some plots of NRMSE over time do not extend over the full 12~hour
window, even though some sample trajectories are still valid, e.g.,
\cref{fig:nvar_nrmse} (left).

\subsection{Temporal Subsampling}
\label{subsec:nvar-subsampling}

\begin{figure}
    \centering
    \includegraphics[width=\textwidth]{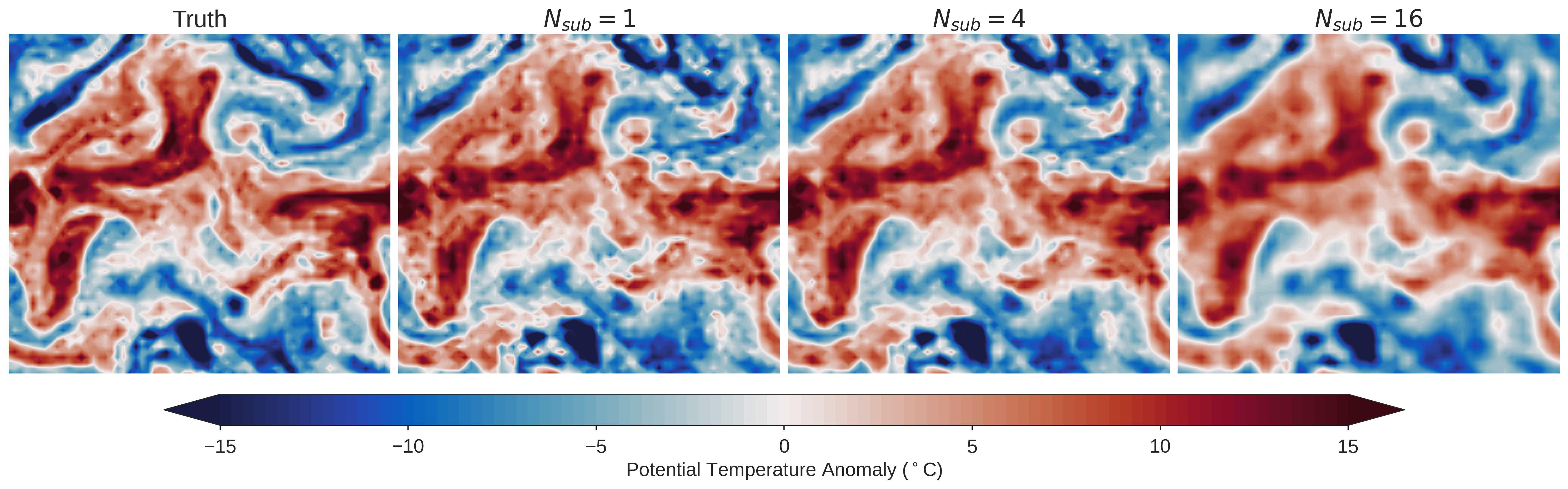}
    \caption{One sample NVAR prediction from the test dataset for
        $\nsub$~=~1,~4,~16;
        shown in the second, third, and fourth panels at a lead time of 4~hours.
        The corresponding truth is shown in the far left panel.
        As the temporal subsampling factor is increased,
        the small spatial scale features are diminished and predictions become
        blurrier.
        Here $\maxlag$~=~1 and only the surface level is shown.
    }
    \label{fig:nvar_qualitative}
\end{figure}

\begin{figure}
    \centering
    \includegraphics[width=.8\textwidth]{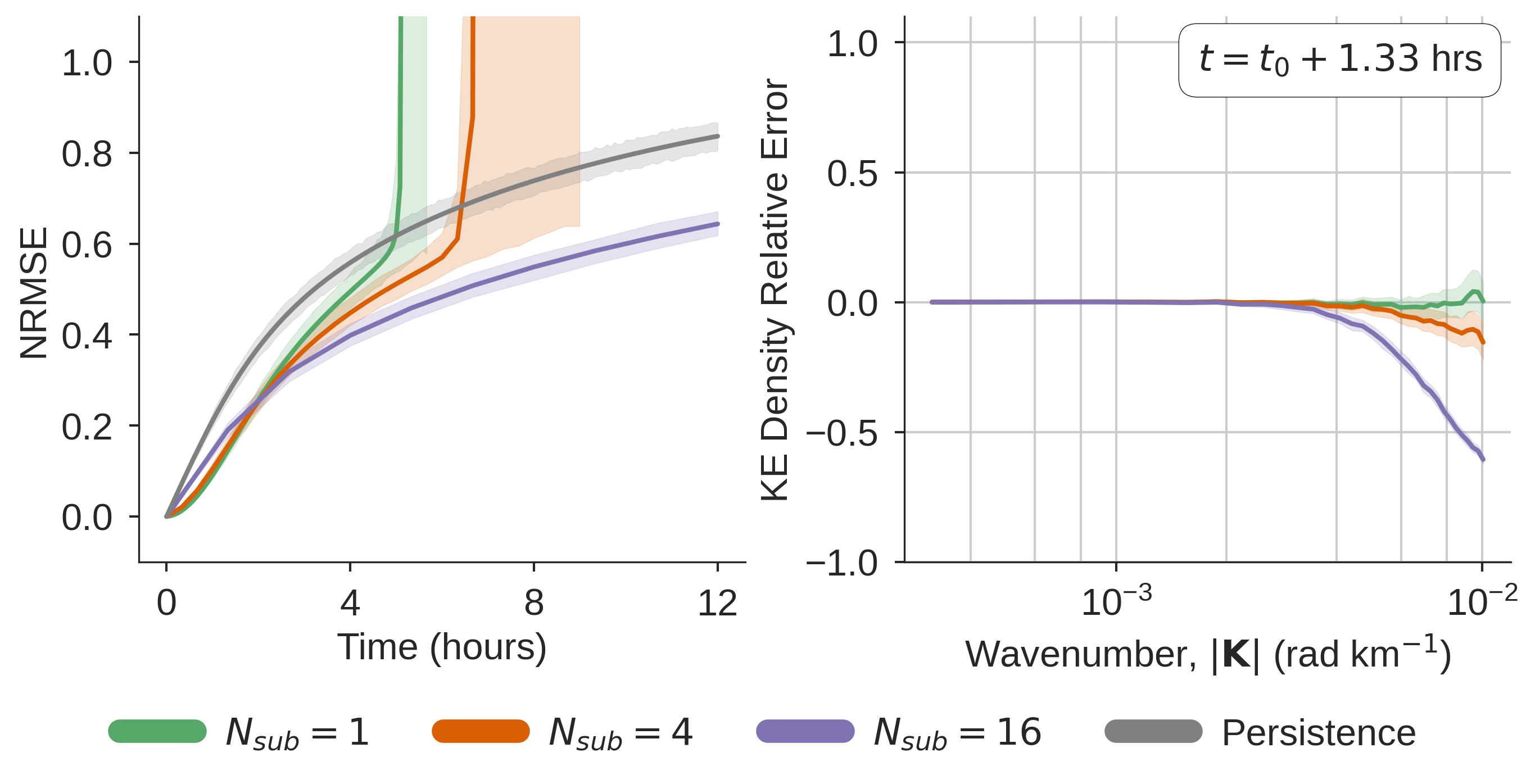}
    \caption{NRMSE (\cref{eq:nrmse}; left) and KE density
        relative error (\cref{eq:ke_relerr}; right)
        indicating prediction skill of the NVAR
        architecture using 50~samples from the test dataset.
        Solid lines indicate averages and shading indicates 99\% confidence
        interval.
        Here $\nlag$~=~1, and the gray line indicates prediction skill of a
        persistent forecast, i.e., where the initial condition does not change.
    }
    \label{fig:nvar_nrmse}
\end{figure}

\cref{fig:nvar_qualitative} shows a qualitative comparison of NVAR predictions
as a function of $\nsub$, i.e., how frequently the training data are sampled and
the model makes predictions.
For this figure, we set $\nlag=1$, and note that both the NRMSE and a snapshot
of the KE density relative error corresponding to this configuration are shown in
\cref{fig:nvar_nrmse}.

At the model timestep ($\Delta \tau = \Delta t = 5$~min; $\nsub=1$), the NVAR predictions are
qualitatively similar to the truth for short forecast lead times.
That is, the NRMSE is near 0, and
many of the small scale features that exist in the truth are also evident
in the predictions.
However, at longer lead times the predictions become unstable.
NRMSE spikes rapidly at about 4~hours after numerical instabilities are
generated, which causes the NVAR model to produce physically unrealistic results.
For reference, Figure S1 shows a view of what these
numerical instabilities look like at their onset.

As the temporal resolution of the data is reduced, i.e., as $\nsub$ increases,
the predictions are generally stable for a longer period of time.
\cref{fig:nvar_nrmse} shows that for $\nsub=4$, predictions are stable for roughly
6~hours, and for $\nsub=16$ no predictions generate numerical instabilities over the
12~hour window.
However, this stability comes with a cost: as the temporal resolution is
reduced, the model's representation of small scale features diminishes as these
features become more blurry or smoothed. This blurring effect is apparent in \cref{fig:nvar_qualitative}, where the prediction is qualitatively more blurry as $\nsub$ increases in each panel from left to right.

This smoothing behavior is captured quantitatively in the right panel of
\cref{fig:nvar_nrmse},
which shows the KE relative error as in \cref{eq:ke_nrmse}.
Here, we show the KE relative error after only 1.33~hours to show the behavior
before instabilities dominate the $\nsub=1$ predictions.
The plot indicates the degree of spectral bias in each solution, which is
largest at the smaller spatial scales, corresponding to higher wave numbers.

At $\nsub=1$ there is a small positive bias at the smallest resolved spatial
scales, indicating that this is when numerical instabilities are starting to
generate.
The subsampled runs, $\nsub=\{4,16\}$, show a negative bias, which corresponds
to a dampened energy spectrum at the scales that are not resolved in the
qualitatively smooth predictions shown in \cref{fig:nvar_qualitative}.
This negative bias is clearly larger with higher subsampling, or reduced
temporal resolution, suggesting that as the data are subsampled, the network
becomes incapable of tracking the small scale dynamics.
The result is an averaged view of what may be occurring in
between each time stamp.

\subsection{Prediction Skill as a Function of Memory}
\label{subsec:nvar-memory}

\begin{figure}
    \centering
    \includegraphics[width=\textwidth]{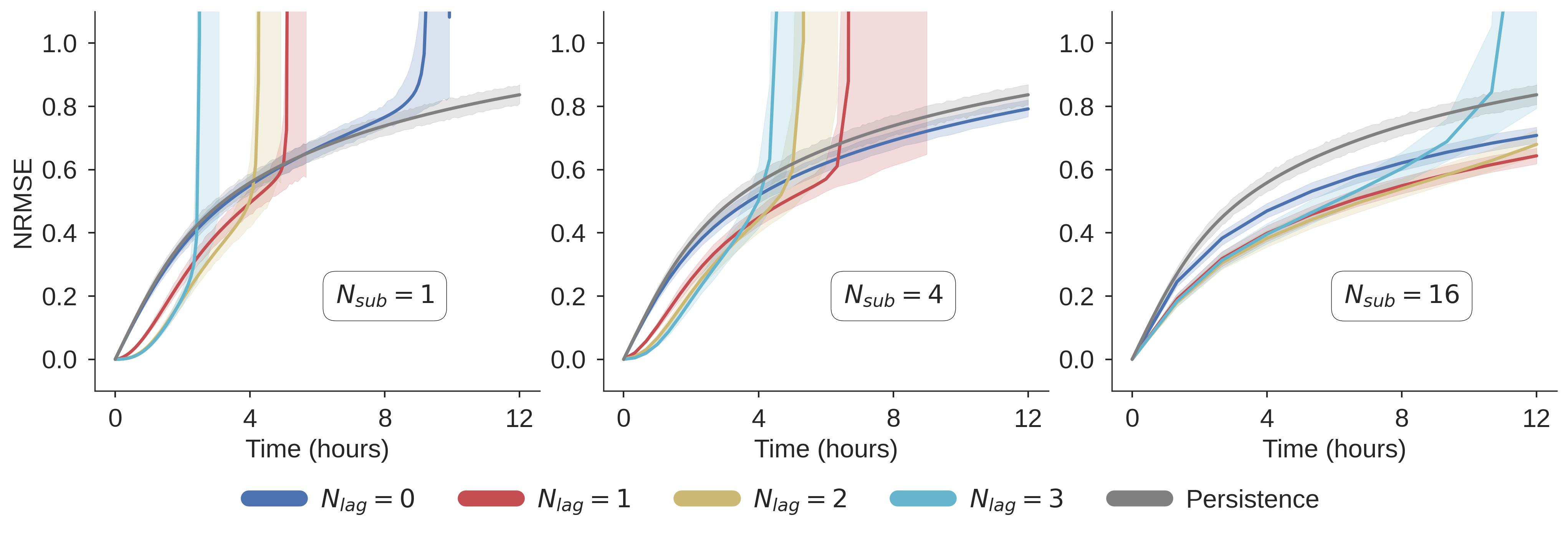}
    \caption{NRMSE computed using NVAR at various temporal resolutions
        ($\nsub$; columns) and with variable memory capacities ($\nlag$;
        colors).
        Decreasing the subsampling factor shows a similar effect as adding
        memory: error is at first reduced, but tends to produce more unstable
        forecasts.
    }
    \label{fig:nvar_nrmse_vs_lag}
\end{figure}

A key feature of RNNs and autoregressive models is that they retain memory of
previous system states.
Given the explicit nature of the NVAR architecture, we explore the effect of
adding memory by increasing $\nlag$, the number of lagged states used to create the
feature vector.
We first summarize how memory impacts prediction skill in
\cref{fig:nvar_nrmse_vs_lag}, which shows the NRMSE as a
function of $\nlag$ (colors) for each
subsampling factor $\nsub = \{1, 4, 16\}$ (panels).
For any value of $\nsub$, adding memory (increasing $\nlag$) reduces
the short term error.
However, adding memory also tends to increase error by the end of the forecast,
often leading to the development of numerical instabilities and an
incoherent solution.
Similarly, for any fixed value of $\nlag$, increasing the temporal resolution
(decreasing $\nsub$) shows the same behavior.

\begin{figure}
    \centering
    \includegraphics[width=\textwidth]{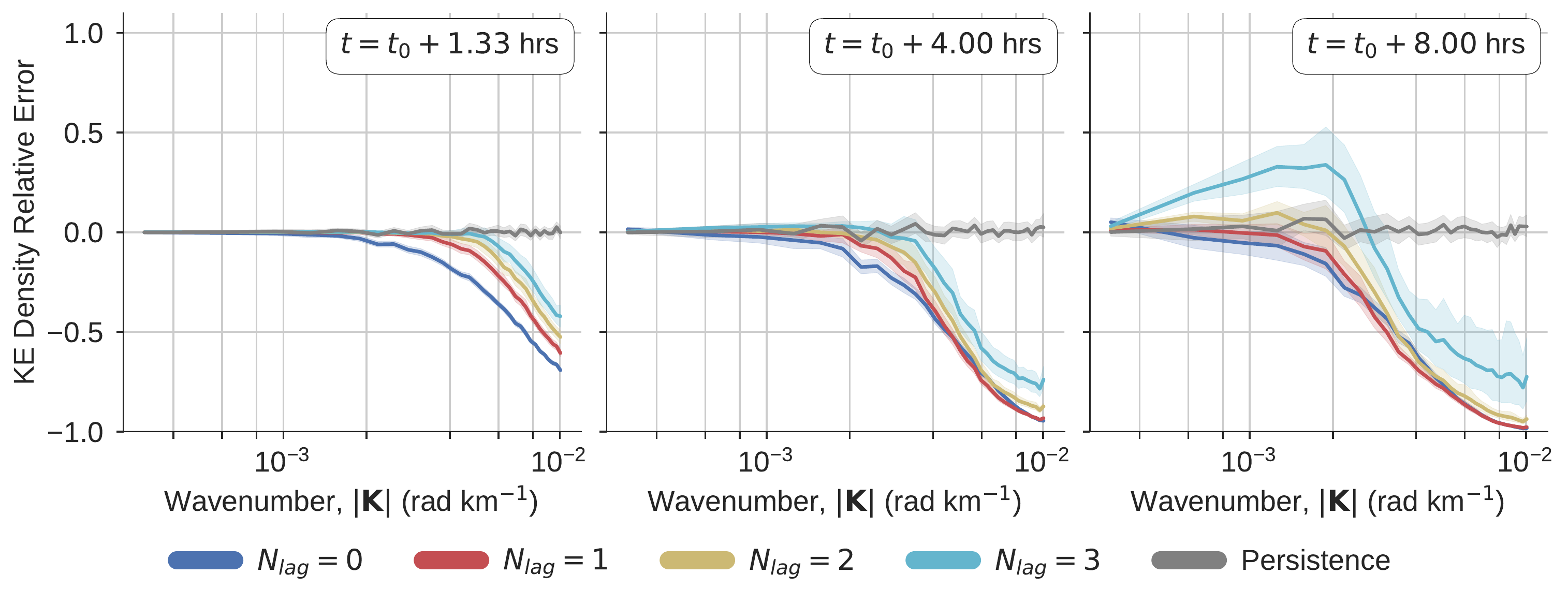}
    \caption{Kinetic energy density relative error with $\nsub$~=~16 at various
        timesteps (columns) and memory capacity ($\nlag$; colors).
        Increasing memory at first reduces error at all spatial scales, but
        later on the error propagates more readily into the large scale.
    }
    \label{fig:nvar_ke_vs_lag}
\end{figure}

To shed some light on how this additional memory impacts the solution,
we show the KE relative error
for the case of $\nsub=16$ as a function of time (panels) and $\nlag$ (colors)
in \cref{fig:nvar_ke_vs_lag}.
For about the first 4~hours, increasing memory improves prediction skill at all
spatial scales.
However, beyond this point, the overall NRMSE grows rapidly, the improvement
at small scales ($|\mathbf{K}|>4\cdot10^{-3}$~rad~km$^{-1}$) is more muted,
and error is propagated rapidly into the larger spatial scales.

We surmise that adding memory degrades the long term prediction skill in the
quadratic NVAR because
the relationship between points further back in history are governed by higher
order nonlinear interactions that are incorrectly represented by
the simple local-quadratic relation that is used here.
As more terms are added that are incorrectly represented, the model becomes
more and more unstable.
We make this supposition based on the fact that despite
theoretical similarities between NVAR and
ESNs as highlighted by \citet{bollt_explaining_2021}, we attain stable
predictions using an ESN architecture with a hyperbolic tangent activation
function in \cref{sec:esn-results}.

The question for the NVAR architecture
is therefore how to retain the short term benefit of added memory
capacity throughout the forecast horizon while maintaining a stable trajectory.
While it may seem natural to explore higher order polynomials to
properly represent this history, we do not explore this further because the size
of the feature vector grows dramatically with the polynomial order
\citep{chen_next_2022}.
Another option would be to explore entirely different basis functions.
While this could be a potential option for future work, we note the findings of
\citet{zhang_catch-22_2022}, who show the extreme sensitivity of NVAR to the
form of nonlinearity imposed.
Given that it is an entirely open question on how to represent the smallest
scales of geophysical turbulence, we do not explore other basis functions, and
instead turn to the more general ESN architecture.

\section{Echo State Network Prediction Skill}
\label{sec:esn-results}

In this section we show the prediction skill of the more general ESN architecture outlined in \cref{subsec:rc}.
Here we use similar metrics as in \cref{sec:nvar-results} to evaluate the ESN
skill, except that we show time averaged quantitative metrics because all of the
ESN predictions are stable for the full twelve-hour forecast horizon.
That is, when shown as a single distribution rather than a time series, NRMSE is reported as
\begin{linenomath*}\begin{equation}
    \text{NRMSE} = \sqrt{
            \dfrac{1}{\ntime\nstate}\sum_{n=1}^{\ntime}\sum_{i=1}^{\nstate}\left(
        \dfrac{\hat{v}_i(n) - v_i(n)}{SD}
        \right)^2 } \, ,
    \label{eq:total-nrmse}
\end{equation}\end{linenomath*}
where $\ntime$ consists of the number of timesteps in the trajectory.
In order to characterize spectral error, we show the KE relative error as in
\cref{sec:nvar-results}.
Additionally, we show the NRMSE in terms of the KE density spectrum as follows
\begin{linenomath*}\begin{equation}
    \text{KE\_NRMSE} = \sqrt{
            \dfrac{1}{\ntime\nk}\sum_{n=1}^{\ntime}\sum_{k=1}^{\nk}\left(
            \dfrac{\hat{E}(n, k) - E(n, k)}{SD(k)}
            \right)^2} \, ,
    \label{eq:ke_nrmse}
\end{equation}\end{linenomath*}
where $\nk$ is the number of spectral coefficients and $SD(k)$ is the temporal
standard deviation of each spectral coefficient throughout the test trajectory.
As in \cref{sec:nvar-results}, all distributions and lineplots indicate
prediction skill from 50 randomly selected initial conditions from an unseen
test dataset.

\subsection{Soft Constraints on Spectral Error}
\label{subsec:esn-ego}

It is well known that ESN prediction skill is highly dependent on the global or
``macro-scale'' parameters noted in
\cref{eq:rc-hyperparameters},
\citep<$\esnparams$, e.g.>[]{platt_systematic_2022,lukosevicius_practical_2012}.
Following the success of previous studies in using Bayesian Optimization methods
to systematically tune these parameters
\citep{griffith_forecasting_2019,penny_integrating_2022,platt_systematic_2022},
we use the Bayesian Optimization algorithm outlined by \citet{jones_efficient_1998} and implemented by
\citet{bouhlel_python_2019} to find optimal parameter values.

More recently, \citet{platt_constraining_2023}
showed that constraining these macro-scale
parameters using global invariant properties of the underlying system leads the
optimization algorithm to select parameters that generalize well to unseen test data. In that work, the authors were successful in using the largest positive
Lyapunov exponent, and to a lesser extent the fractal dimension of the system.
Because of the focus on resolved scales in this work, we take a similar approach, but test the effect of constraining the ESN to the KE density spectral coefficients.
Specifically, we implement the following two-stage training process.
At each step, the macro-scale parameters, $\esnparams$, are fixed, and the
``micro-scale'' parameters $\Wout$ are obtained by minimizing \cref{eq:cost}.
This readout matrix is then used to make forecasts from randomly selected
initial conditions from a validation dataset.
The skill of each of these forecasts is captured by the macro-scale cost
function
\begin{linenomath*}\begin{equation}
    \cf_\text{macro}(\esnparams) = \dfrac{1}{\nmacro}
    \sum_{j=1}^{\nmacro}
    \left\{
        \text{NRMSE}(j) + \gamma \text{KE\_NRMSE}(j)
    \right\}
    \label{eq:macro-cost} \, ,
\end{equation}\end{linenomath*}
where NRMSE and KE\_NRMSE are defined in \cref{eq:total-nrmse,eq:ke_nrmse},
$\nmacro$ is the number of forecasts used in the validation set, and $\gamma$ is
a hyperparameter that determines how much to penalize deviations
from the true KE density spectrum.
The value of $\macrocost$ is then used within the Bayesian Optimization algorithm, which reiterates the whole optimization process with new values for
$\esnparams$ until an optimal value is found or the maximum number of iterations is reached.
Here, we use $\nmacro=10$, initialize the optimization with 20~randomly sampled points in the 5~dimensional parameter space, and run for 10~iterations. Note that we run this optimization procedure for each unique ESN configuration
throughout \cref{sec:esn-results} (i.e., for each $\nsub$ and each $\gamma$
value).

\begin{figure}
    \centering
    \includegraphics[width=\textwidth]{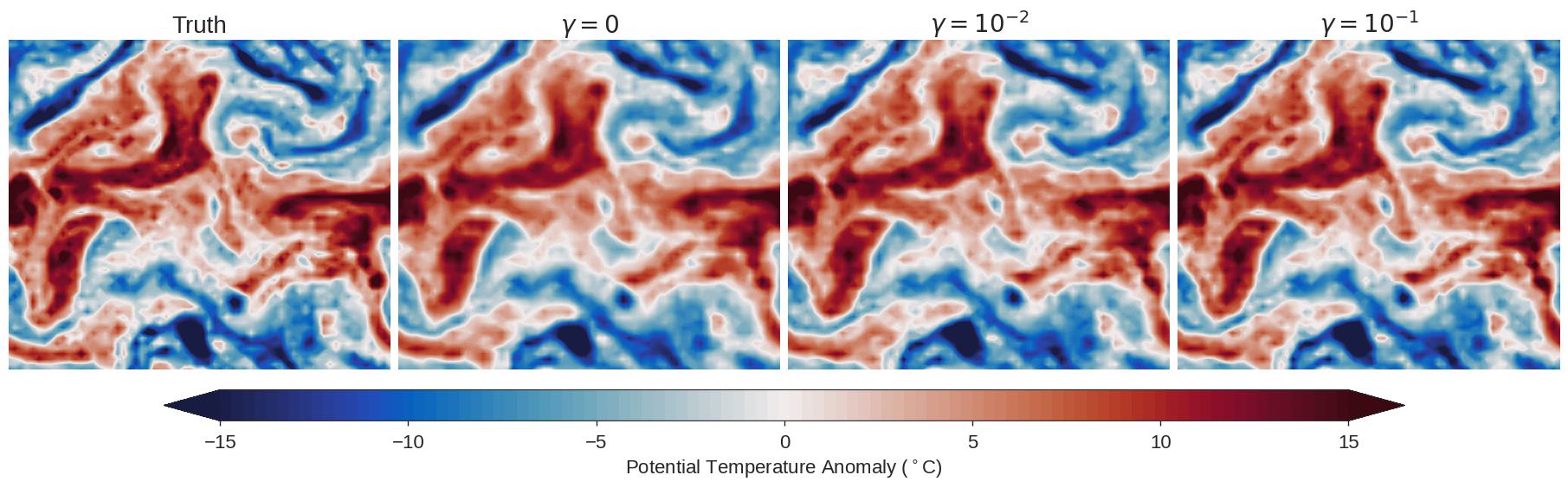}
    \caption{
        One sample prediction from the test dataset, where each panel shows
        potential temperature in the truth (left) and subsequently for
        ESN predictions with parameters optimized using
        $\gamma$~=~\{0,~$10^{-2}$,~$10^{-1}$\} in \cref{eq:macro-cost}.
        Each panel shows the prediction at a forecast lead time of 4~hours,
        using the same initial conditions as in \cref{fig:nvar_qualitative}.
        As $\gamma$ increases from left to right, the prediction becomes sharper
        (i.e., less blurry).
        Here, the ESN is evaluated at the SQG model timestep, i.e., $\nsub$~=~1.
    }
    \label{fig:rc_qualitative_nsub01}
\end{figure}

\cref{fig:rc_qualitative_nsub01} shows a qualitative view of how penalizing the
KE density impacts ESN prediction skill when it operates at the original
timestep of the SQG model (i.e., $\nsub=1$).
At $\gamma=0$, the ESN parameters are selected based on NRMSE alone, and
the prediction is relatively blurry.
However, as $\gamma$ increases to $10^{-1}$, the prediction becomes sharper as
the small scale features are better resolved.

\begin{figure}
    \centering
    \includegraphics[width=\textwidth]{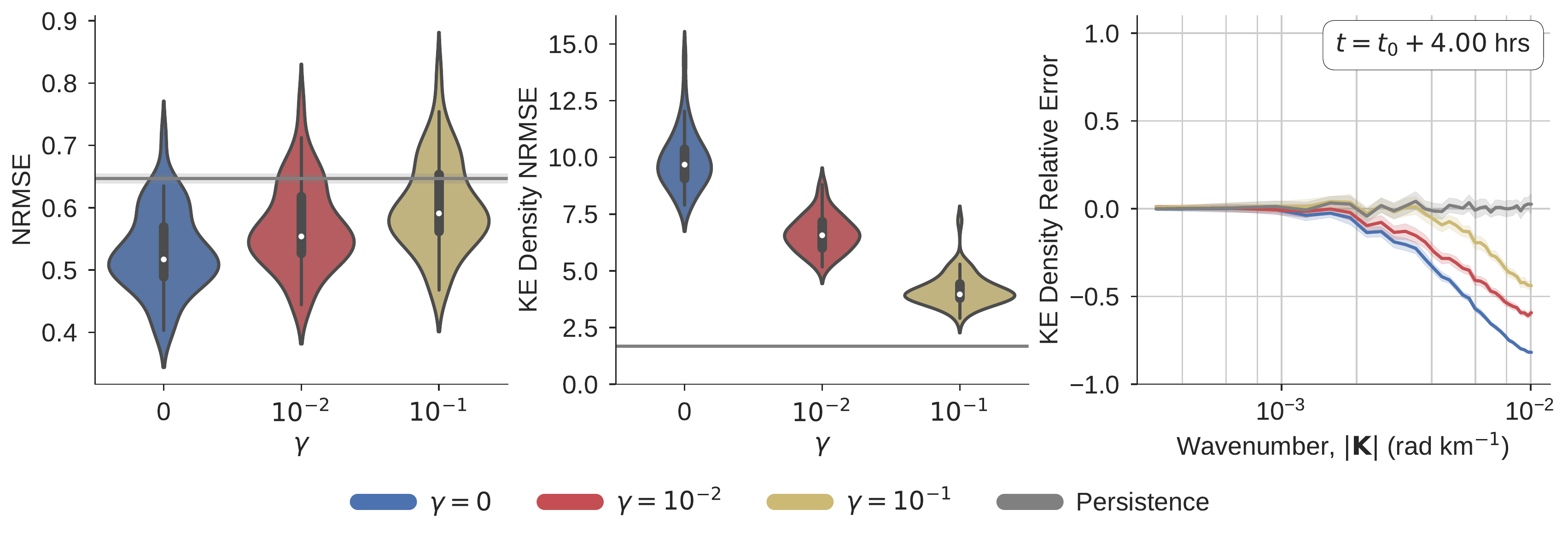}
    \caption{
        Quantitative comparison of ESN predictions at $\nsub$~=~1 with macro-scale
        parameters
        chosen using different values of $\gamma$ in \cref{eq:macro-cost}.
        NRMSE (\cref{eq:total-nrmse}; left), KE\_NRMSE (\cref{eq:ke_nrmse};
        middle), and
        KE relative error (\cref{eq:ke_relerr}; right) highlight the tradeoff
        between minimizing NRMSE and spectral error: as $\gamma$ increases
        spectral error is reduced, but NRMSE increases.
        Note that the KE relative error is shown at 4~hours to provide
        direct comparison to the snapshots in \cref{fig:rc_qualitative_nsub01}.
        In each plot, the solid gray line indicates the median skill of a persistent
        forecast.
    }
    \label{fig:rc_quantiative_nsub01}
\end{figure}

\cref{fig:rc_quantiative_nsub01} gives a quantitative view of how the KE density
penalty changes ESN prediction skill, once again with $\nsub=1$.
The first two panels show that there is a clear tradeoff between NRMSE and KE error:
as $\gamma$ increases the NRMSE increases but the spectral representation improves.
The final panel in \cref{fig:rc_quantiative_nsub01}
shows that the spatial scales at which the spectral error manifests in these
different solutions.
When $\gamma=0$, the macro-scale parameters are chosen to minimize NRMSE,
leading to blurry predictions and a dampened spectrum at the higher wavenumbers,
especially for $|\mathbf{K}| > 2\cdot10^{-3}$~rad~km$^{-1}$.
We note that \citet{lam_graphcast_2022} report the same behavior when using a cost function that is purely based on mean-squared error.
On the other hand, when $\gamma = 10^{-1}$, the global parameters are chosen to
minimize both NRMSE and KE density error, where the latter treats all spatial
scales equally.
In this case, KE relative error is reduced by more than a factor of two and the
spectral bias at higher wavenumbers is much more muted.

Of course, the tradeoff for the reduced spectral error is larger NRMSE, resulting
from slight mismatches in the position of small scale features in the forecast.
However, our purpose is to generate forecasts that are as representative
of the training data as possible.
Overly smoothed forecasts are not desirable, because this translates to losing local extreme values,
which are of practical importance in weather and climate.
Additionally, a key aspect of ensemble forecasting is that the truth remains a
plausible member of the ensemble \citep{kalnay_ensemble_2006}.
Therefore, representing the small scale processes, at least to some degree,
will be critical for integrating an
emulator into an ensemble based prediction system.

Finally, we note that using a cost function with only KE\_NRMSE produced inconsistent
results.
Therefore, we consider it important to keep the NRMSE term in the
cost function, as this prioritizes the position of small scale features, i.e.,
maintains phase information.
Additionally, we note that there is some irreducible high wavenumber error,
which is most clearly seen by comparing the prediction skill to a persistent
forecast.
While the sample median NRMSE for each $\gamma$ value beats persistence, the
KE\_NRMSE is more than double, due to this error at the small spatial scales.
Ideally, our forecasts would beat persistence in both of these metrics, but
obtaining the ``realism'' in the small spatial scales necessary to
dramatically reduce this spectral error should be addressed in future work.


\subsection{Temporal Subsampling}
\label{subsec:esn-subsampling}

The NVAR predictions shown in \cref{subsec:nvar-subsampling} indicate that
subsampling the training data systematically increases error at small spatial
scales.
However, the architecture was not specifically designed or constrained to
have a good spectral representation of the underlying dynamics.
On the other hand, the previous section (\cref{subsec:esn-ego})
showed that the spectral bias at high wavenumbers
can be reduced by optimizing the global ESN parameters
to the true KE density spectrum.
Given these two results, we explore the following question: does temporal subsampling still
increase spectral bias in the more general ESN framework, even when parameters
are chosen to minimize this bias?

\begin{figure}
    \centering
    \includegraphics[width=\textwidth]{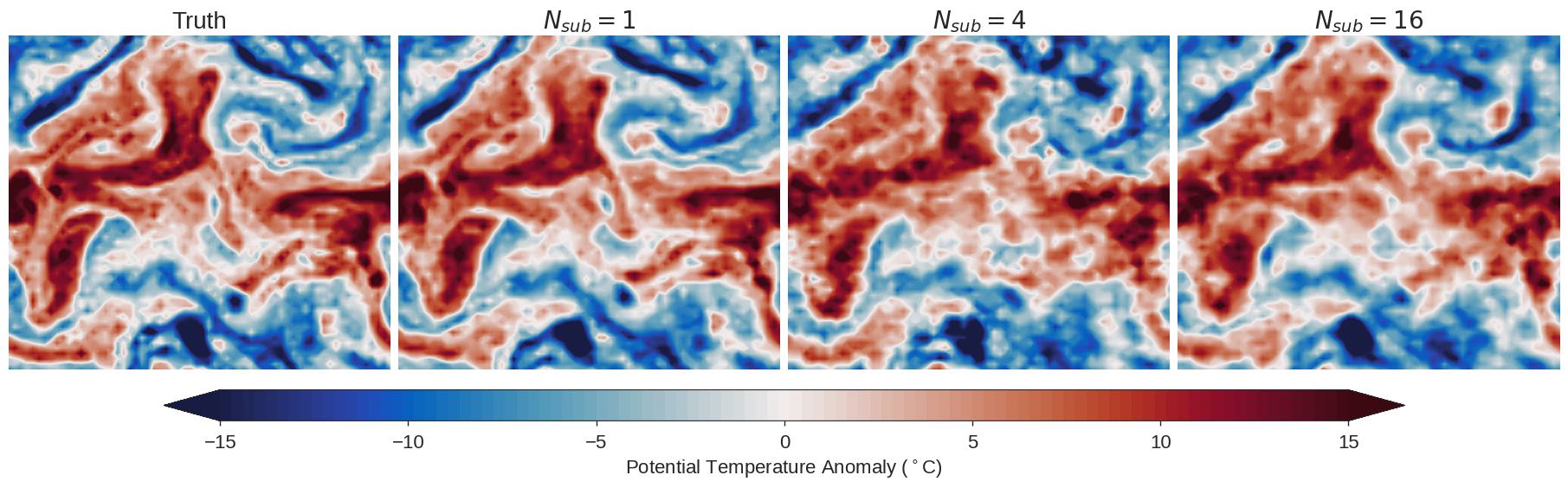}
    \caption{One sample prediction from the test dataset, exactly as in
        \cref{fig:rc_qualitative_nsub01}, except here $\gamma$~=~$10^{-1}$ is fixed, and
        the temporal subsampling factor is varied: $\nsub$~=~\{1,~4,~16\}.
        As the temporal subsampling factor increases, the small spatial scale
        features are lost and the prediction becomes blurrier.
    }
    \label{fig:rc_qualitative_gamma0.1}
\end{figure}

\begin{figure}
    \centering
    \includegraphics[width=\textwidth]{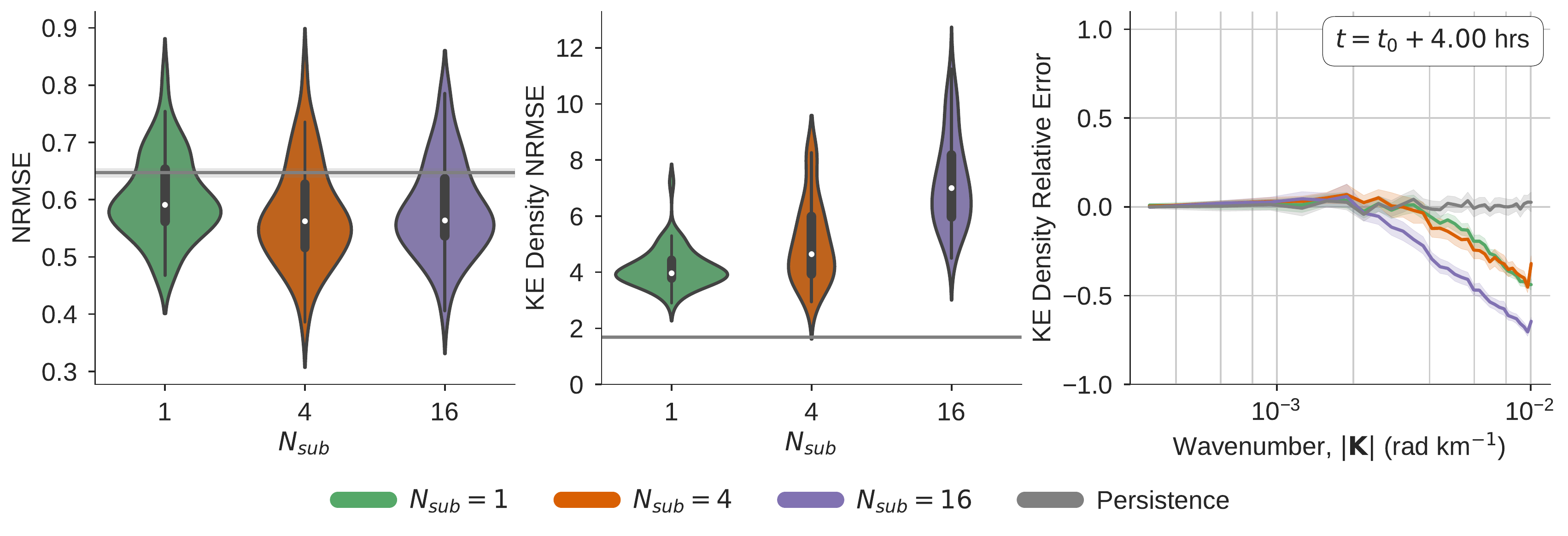}
    \caption{Quantitative comparison of ESN predictions, showing
        NRMSE (left), KE\_NRMSE (middle), and KE relative error (right), exactly as in
        \cref{fig:rc_quantiative_nsub01}, except here $\gamma$~=~$10^{-1}$ is fixed,
        and the temporal subsampling factor is varied: $\nsub$~=~\{1,~4,~16\}.
        As the temporal subsampling factor increases, spectral errors increase.
        In each plot, the solid gray line indicates the median skill of a persistent
        forecast.
    }
    \label{fig:rc_quantiative_gamma0.1}
\end{figure}

\cref{fig:rc_qualitative_gamma0.1} and \cref{fig:rc_quantiative_gamma0.1}
show that even when the macro-scale parameters are chosen to prioritize the KE
density representation (i.e., $\gamma = 10^{-1}$ is fixed),
temporal subsampling does lead to an apparently inescapable spectral bias.
This effect is shown qualitatively in \cref{fig:rc_qualitative_gamma0.1},
where the predictions become
smoother as the temporal subsampling factor, $\nsub$, increases.
The effect is similar to what was seen with NVAR except the blurring effect is
less pronounced.
Quantitatively, \cref{fig:rc_quantiative_gamma0.1}(b) shows that as $\nsub$
increases, error in KE density spectrum generally increases, while panel (c) shows that this KE
error is concentrated in the small spatial scales,
$|\mathbf{K}| > 2\cdot10^{-3}$~rad~km$^{-1}$.
We note that the degree of spectral bias at $\nsub=16$ is smaller than what was
achieved with NVAR for the same $\nsub$ value, cf. \cref{fig:nvar_ke_vs_lag},
indicating that the optimization was successful in reducing the spectral bias.

\begin{figure}
    \centering
    \includegraphics[width=\textwidth]{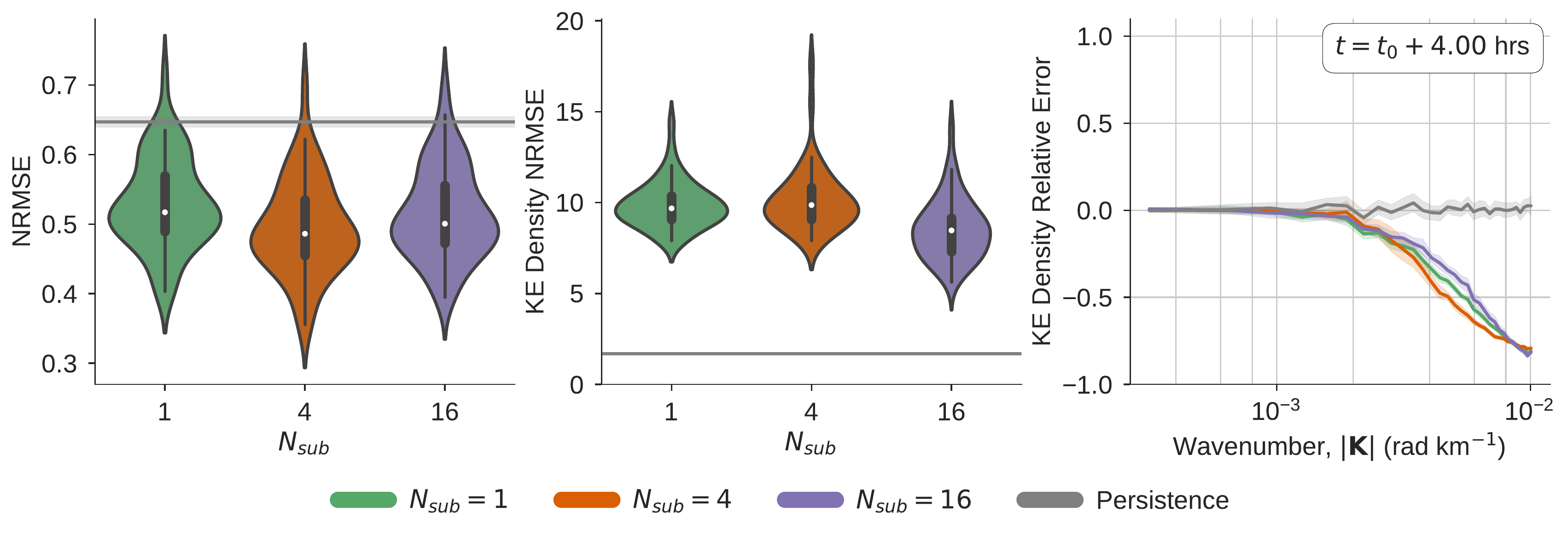}
    \caption{Same as \cref{fig:rc_quantiative_gamma0.1}, except here $\gamma$~=~0,
        indicating that only NRMSE is penalized in the cost function.
        The error is relatively similar, indicating that NRMSE alone is a
        suboptimal penalty for model selection.
        In each plot, the solid gray line indicates the median skill of a persistent
        forecast.
    }
    \label{fig:rc_quantiative_gamma0.0}
\end{figure}

Interestingly, there is little difference between NRMSE obtained by the ESNs at
different $\nsub$ values.
Additionally, \cref{fig:rc_quantiative_gamma0.0} shows that there is little
difference in both NRMSE and KE\_NRMSE when $\gamma=0$, i.e., when NRMSE is the only criterion
for parameter selection.
This result shows that NRMSE alone is not a good criterion for model selection, given
that we have shown success in reducing spectral errors by prioritizing the spectrum appropriately.

\subsection{Impact of the Hidden Layer Dimension}
\label{subsec:esn-size}

The dimension of the hidden layer, $\nhidden$, also known as the reservoir size, determines the
memory capacity available to the ESN
\citep{jaeger_echo_2001,lukosevicius_practical_2012}.
For systems with high dimensional input signals, it is crucial to use a sufficiently large hidden layer to afford the memory capacity necessary for accurate
predictions \citep{hermans_memory_2010}.
In all of the preceding sections we fixed $\nhidden=6,000$ for each local group,
where for reference each local group has an input dimension of
$\nlocalinputstate=200$ and an output dimension of $\nlocalstate=128$.
Here, we briefly address the effect of doubling the hidden layer dimension, while keeping the input and output dimensions constant, in order to test how
sensitive our conclusions are on this crucial hyperparameter.
Due to the computational expense of the parameter optimization discussed in
\cref{subsec:esn-ego}, we only perform this experiment for $\nsub=16$.

\begin{figure}
    \centering
    \includegraphics[width=\textwidth]{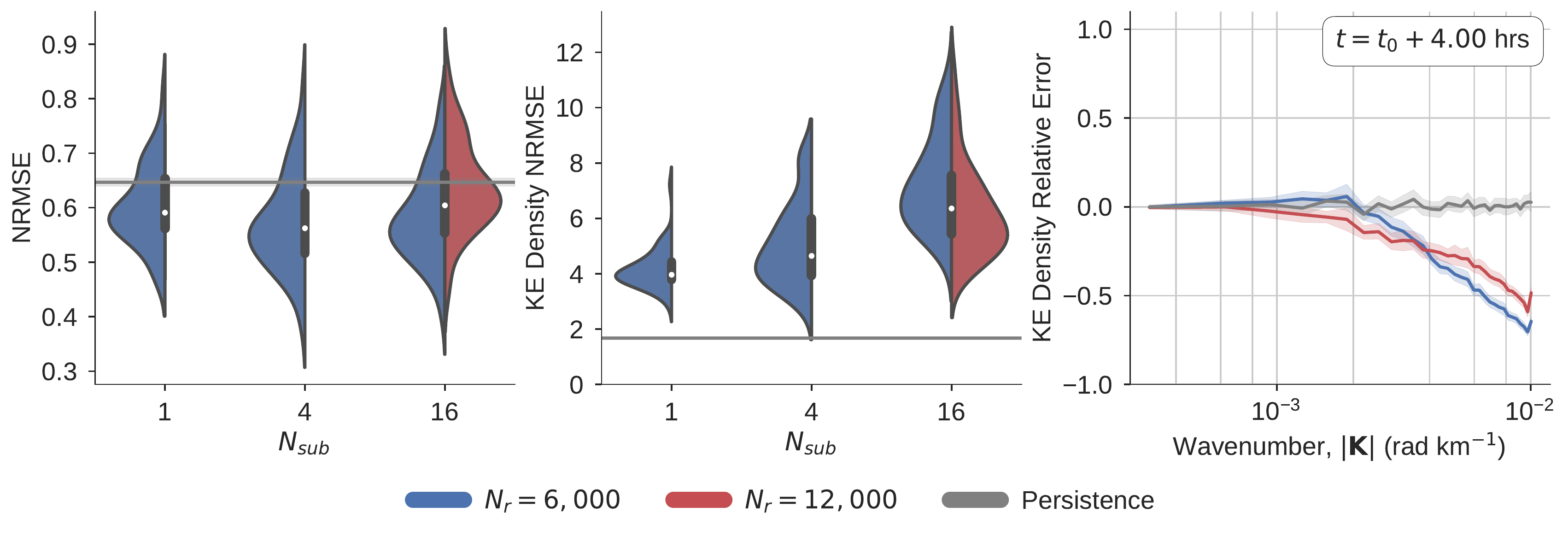}
    \caption{The impact of doubling the hidden layer dimension from
        $\nhidden$~=~6,000 to
        $\nhidden$~=~12,000 on NRMSE (left), KE\_NRMSE (middle), and KE relative
        error (right).
        Increasing the hidden layer dimension is relatively proportional to reducing
        the temporal subsampling factor, indicating a potential brute force
        approach to reducing the subsampling related spectral errors.
        Here $\gamma$~=~$10^{-1}$, and the solid gray line indicates the
        median skill of a persistent forecast.
    }
    \label{fig:esn-size}
\end{figure}

The impact of doubling $\nhidden$ on prediction skill is shown in
\cref{fig:esn-size}, where for the sake of brevity we only show results for the
case when $\gamma=10^{-1}$ in \cref{eq:macro-cost}.
The left panel shows that the larger hidden layer actually increases the NRMSE slightly.
However, the middle and right panels show that this increase is due to the improved
spectral representation.
The improvement in KE\_NRMSE is nearly proportional to the improvement achieved by increasing
the temporal resolution of the data.
That is, doubling the hidden layer width reduces the average KE\_NRMSE by 14\%,
while increasing the temporal resolution of the data by a factor of 4 reduces
the KE\_NRMSE by 30\%.
These results indicate a potential brute force approach to overcoming the
subsampling related spectral errors.
However, the larger hidden layer dimension has to be constrained
with enough training data, and requires more
computational resources.

\subsection{Impact of Training Dataset Size}
\label{subsec:esn-fixed-steps}

In all of the preceding experiments, the length of training time was fixed to
15~years, meaning that there are fewer training samples when the data are
subsampled, i.e., as $\nsub$ grows.
Specifically, 15~years of data at an original model timestep of 5~minutes means
that there are approximately
$1.6\cdot10^{6}$, $3.9\cdot10^5$, and $9.72\cdot10^4$ samples
for each case previously shown: $\nsub=1$, 4, and 16, respectively.
Here, we show that even when the number of training samples is fixed, the
subsampling related spectral errors are still present.

\begin{figure}
    \centering
    \includegraphics[width=\textwidth]{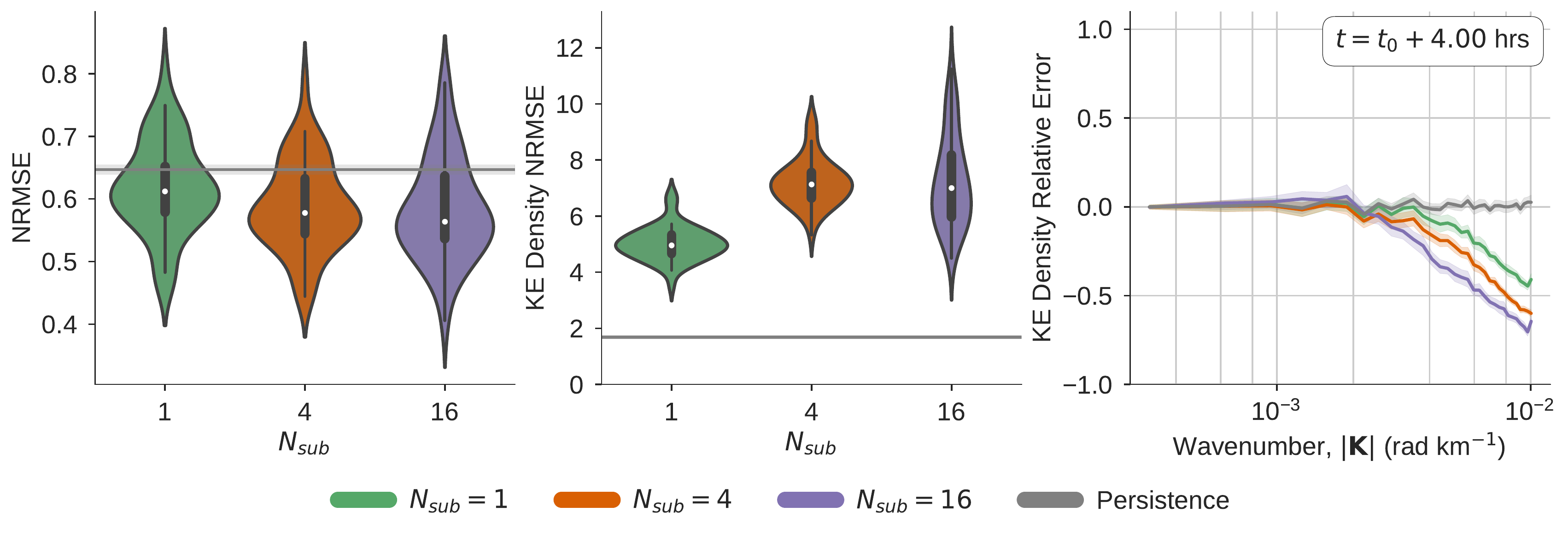}
    \caption{Subsampling related spectral errors persist even when the number of
        training samples is fixed. Here, the number of samples is fixed to
        $9.72\times10^{4}$ for all cases, and yet the temporal subsampling
        related spectral errors remain.
        Here, $\gamma$~=~$10^{-1}$ and the solid gray line indicates the median skill of a persistent
        forecast.
    }
    \label{fig:esn-fixed-steps}
\end{figure}

\cref{fig:esn-fixed-steps} shows the prediction skill in terms of NRMSE and
spectral errors when the number of training samples is fixed to $9.72\cdot10^4$.
With this number of samples, the training data is exactly the same for
$\nsub=16$, but only spans $3.75$ and $0.94$~years for $\nsub=4$ and $\nsub=1$,
respectively.
However, we see the same general trend as before: subsampling the data improves
NRMSE slightly but increases the KE\_NRMSE.
As before, the spectral error is largest in the higher wavenumbers,
$|\mathbf{K}| > 2\cdot10^{-3}$~rad~km$^{-1}$.
We note that the difference in performance between $\nsub=4$ and $\nsub=16$ is
marginal.
The only notable difference between these two cases is that the ESN is less
consistent, i.e., the KE\_NRMSE distribution is broader, when $\nsub=16$.
However, it is clear that spectral error is lowest when the data are not
subsampled at all, even though less than a year of data is used.
This result indicates that there could be a benefit to training a RNN on a
relatively shorter model trajectory that is untouched, rather than a longer
dataset that is subsampled in time.

\section{Discussion}
\label{sec:discussion}

Weather and climate forecasting necessitates the integration of expensive
numerical models to make accurate predictions and projections.
The computational cost of these models often results in tradeoffs, where
practitioners must balance the spatial resolution of their model with other
factors, such as the number of integrated model components or the ensemble
size that can be afforded in the system.
Model emulation or surrogate modeling aims to enable such predictions
by emulating the dynamical system with adequate accuracy at a much lower
computational expense.
In this study, our primary interest was to shed light on the spatial scales
that can be resolved by single layer autoregressive and recurrent
neural network emulators in order to better
understand the effective resolution that could be achieved in weather and
climate applications.
We used two relatively simple, single layer autoregressive and recurrent neural
network architectures, mainly because it has been shown that they can
successfully emulate low dimensional chaotic dynamics over multiple Lyapunov
timescales
\citep{pathak_using_2017,vlachas_backpropagation_2020,gauthier_next_2021,platt_systematic_2022}.
We implemented a multi-dimensional parallelization scheme based on the concept
introduced by \citet{pathak_model-free_2018} and similar to that of
\citet{arcomano_machine_2020}
in order to scale up these architectures and test them in high dimensional
systems.
We note that an in-depth discussion of our software implementation using the
task based scheduling system in python, Dask \citep{dask_2016}, will be covered in a
forthcoming paper.

\subsection{Main Result and Connections to Previous Work}
\label{subsec:discuss-past-work}
Our main result is that we observe an inherent spectral bias that occurs when
training data are subsampled in time, such that as the temporal resolution is
reduced, the resolution of small scale features in NVAR and ESN predictions is diminished.
High wavenumber spectral bias is a phenomenon
that has been studied in the context of training feed forward neural
networks \citep<see>[for a comprehensive review on the topic]{xu_overview_2022}.
The authors show that while
numerical Partial Differential Equation (PDE) solvers typically resolve small spatial scales first
and iteratively refine the larger spatial scales,
spectral biases arise while training neural
networks because the reverse happens: the large scales are uncovered first and
small spatial scales are slowly refined.

Here, we showed a similar bias that arises in NVAR and ESN architectures in relation
to their temporal resolution.
Given the sensitivity to model time step, this phenomenon bears
resemblance to the Courant-Friedrich-Lewy (CFL) condition, which poses an upper
bound on the time step size that can be used in the numerical solution of PDEs.
The CFL condition is therefore a barrier to weather and climate model efficiency.
However, sensitivity to the time step size manifests very
differently in neural networks and numerical PDEs.
While violating the CFL condition with too large of a time step leads to
fundamental issues of numerical instability in numerical PDEs,
here we see that increasing the time step adds a sort of numerical dissipation,
which can actually stabilize an otherwise unstable model architecture
(\cref{subsec:nvar-subsampling}).
We suggest that this occurs because the small scales are ``lost'' within the recurrent and
autoregressive time stepping relations.
Because of this, the models are trained to take on
an interpolated or spatially averaged view of the intermediate dynamical
behavior, which generates a blurred prediction.

We note that \citet{bi_pangu-weather_2022} discuss a similar phenomenon relating
to the timestepping of their autoregressive transformer model.
Specifically, they devise a ``Hierarchical Temporal Aggregation'' scheme to make
more stable and accurate forecasts (in terms of RMSE) over longer periods of time than they would
potentially be able to if they were to use the original 1~hour cadence of the ERA5 dataset.
However, it is not clear how well small scale features are preserved with this
approach.
This is unclear first because they use a cost function that is purely based on
RMSE.
Secondly, the approach requires training multiple models at successively larger
time intervals, and a forecast is made using the largest interval possibly
available first.
For instance, with trained models operating on 1 and 6 hourly increments, a
7~hour forecast would be made by first a 6 and then 1~hour prediction.
Our results indicate that this could be problematic, as the model making the 6~hour
prediction would filter out small scale features that would otherwise be
captured by the second model, operating on a 1~hour timestep.

Finally, \citet{chattopadhyay_long-term_2023} show the connection between
high wavenumber spectral bias and instabilities in neural network predictions of
turbulent flows.
Their focus was on achieving long term stability in neural network time
stepping for climate applications, while the focus in our work has been on short
term forecasting for weather applications -
capturing the long term, climate statistics in turbulent geophysical fluid
dynamics with an ESN or NVAR is future work.
However, both works (1) draw some
connection between high frequency spectral bias and the time stepping
of the neural network, and (2) offer potential solutions by penalizing the
solution's spectrum.
In our work, we show that some of the spectral bias stems from the timestep size of
the data used for training, while \citet{chattopadhyay_long-term_2023}
devise a Runge-Kutta scheme to reduce the bias on subsampled data.
Additionally, they use a spectral loss to train the internal weights of the
network, along with the addition of a ``corrector'' network to make predictions
of only the small scales.
On the other hand, we use a spectral loss to guide the optimization of 5
``macro-scale'' parameters, but the training of the network weights and
operation of the network remain the same.
Despite the differences in approach,
the similarity of these two works indicates that the details of neural network
time stepping schemes are crucial to their stability and accuracy in
representing small scale processes.
Additionally, it is clear that these small scale processes must be prioritized
in some way, for instance through a loss function, and potentially additional
``corrector'' networks that propagate the small scales explicitly.

\subsection{Implications for Training Datasets in Weather and Climate}
This result has important implications for the rapidly developing field of
neural network emulation for weather and climate forecasting because it shows a
potential limit to the effective resolution of an emulator relative to the
original training data.
If an emulator is used as a parameterization scheme for subgrid-scale dynamics,
then a high wavenumber spectral
bias will be detrimental to performance.
Additionally, we surmise that such errors will reduce ensemble spread within
data assimilation algorithms, which could limit their usefulness within a
forecasting system \citep<e.g.,>[]{kalnay_ensemble_2006}.
Our findings are pertinent to the field of neural network emulation development
because of the
widespread usage of reanalysis datasets for training.
Currently, most existing neural network emulators in this field use the ERA5
reanalysis dataset \citep{hersbach_era5_2020} for training
\citep<e.g.,>[]{lam_graphcast_2022,bi_accurate_2023,pathak_fourcastnet_2022,keisler_forecasting_2022,weyn_sub-seasonal_2021,arcomano_machine_2020}.
Of course, reanalyses like ERA5 are an obvious choice for many reasons:
the datasets are made freely available, they present a multi-decadal view of
weather and climate, and, most importantly, they are constrained to observational
data.
However, we note that reanalysis products are imperfect for at least the following reasons:
they contain jumps in the system state at the start of each DA cycle,
they may contain inconsistencies reflective of changes in observational
coverage,
and they are only made available at large time intervals relative to the time step of the underlying integrated numerical model dynamics, due to the massive size of the data.
Our study only addressed the latter of these issues, and showed that this simple
space-saving step can have a negative impact on data-driven prediction methods.
While we showed that adding spectral error as a weak constraint in the neural
network training can reduce this time step related spectral bias, our results
indicate that the underlying issue persists (\cref{subsec:esn-subsampling}).
Moreover, as long as the data are not subsampled, we showed that ESNs
perform only slightly worse when $<1$~year of data are used, compared to
15~years of training data (\cref{subsec:esn-fixed-steps}).
This result suggests that it may be more effective to design an RNN-based emulator
with a relatively short model trajectory that is not subsampled, rather than a long
trajectory that is subsampled.
In contrast to training the emulator on a reanalysis dataset,
a pure model-based emulator could then be used within a data assimilation system
as by
\citet{penny_integrating_2022}
in order to additionally benefit from observational constraints.

\subsection{Implications and Future Work Relating to Model Architecture}
Due to the fact that RNNs require long, sequential data streams in order to
learn the governing dynamics, it could be the case that RNNs suffer most
dramatically from temporal subsampling.
This hypothesis could be one reason for why the RNNs used by
\citet{agarwal_comparison_2021} performed worse than other models on data that
were subsampled every 10~days.
Additionally, if RNNs are most dramatically affected by temporal subsampling, then
they could be a suboptimal
architecture choice for model emulators in cases where representing small scale
dynamics is important but a coarse time step is required.
This requirement is especially true when designing a
parameterization scheme for subgrid-scale dynamics, where the emulator should ideally run at the same time step as the
``large-scale'' model.

However, given that we can qualitatively observe some degree of spectral error in
a wide variety of neural network architectures that use subsampled data for
training
\citep<e.g.,>[]{lam_graphcast_2022,bi_accurate_2023,pathak_fourcastnet_2022,keisler_forecasting_2022},
the issue could be more general to other neural network
architectures.
Moreover, the similarities between our work and
\citet{chattopadhyay_long-term_2023}
as well as the reasons behind the
hierarchical time stepping scheme introduced by \citet{bi_pangu-weather_2022} (both
discussed in \cref{subsec:discuss-past-work}) imply that
the time stepping related spectral bias is a general issue.
Therefore, we suggest that future work should be directed at understanding the degree to which
temporal resolution affects architectures other than RNNs.
Potential avenues
could include exploring how attention mechanisms
\citep{vaswani_attention_2017,dosovitskiy_image_2021}
handle this phenomenon.
Additionally, in light of our results indicating that wider networks can
mitigate the spectral bias at least to some degree (\cref{subsec:esn-size}), it would be
instructive to understand how successively adding layers to a neural
network affects the spectral bias.
Finally, we note the work of \citet{duncan_generative_2022} who show success in
using adversarial training to mitigate the spectral bias observed in
FourCastNet, and suggest that such techniques deserve additional study to
understand their robustness.

Of course, our neural network implementations are imperfect, and we suggest some
future avenues to improve their predictive capabilities.
Both of the architectures relied on a mean-squared error micro-scale cost function to learn
the readout matrix weights, even in the ESN models where the spectral errors
were penalized in the macro-scale cost function.
However, even when the spectrum was penalized and the data were not subsampled,
the ESNs maintained a high wavenumber bias that
resulted in KE\_NRMSE far greater than that of a persistent forecast.
While additional testing shows that a periodic sine activation function can
reduce the high frequency bias in KE\_NRMSE, following work by
\citet{sitzmann_implicit_2020},
the underlying problem still remains (see additional analysis in the
Supplemental Materials).
Therefore, in order to further reduce the high frequency bias, it may be necessary to
move the spectral penalties to the micro-scale
cost function, i.e., to learn the readout matrix weights in the case of
reservoir computing.
The time stepping, spectral loss, and ``small scale corrector
network'' employed by \citet{chattopadhyay_long-term_2023} would be appropriate
starting points for such future work.

The NVAR architecture that we employed is incredibly simple.
While we supposed that the local quadratic feature vector could learn
quantities like derivatives and fluxes necessary to step the model forward in
time, it is apparently not robust enough
given the dramatic sensitivity to time step used.
Future work could explore the possibility of using a larger library of analytic functions to improve the nonlinear expressions in the model,
with the caution that this will lead to very high dimensional feature vectors. Such developments must sufficiently address the ``Catch-22''
described by \citet{zhang_catch-22_2022}, who show that NVAR is inherently
sensitive to the types of nonlinearity chosen.
It is entirely possible, though, that an appropriate set of such basis functions exist for weather and climate emulation.

The ESN architecture that we employed is also relatively straightforward, and
can undoubtedly be improved.
In this work we took a somewhat brute force approach to emulate arbitrarily
high dimensional systems by partitioning the system into subdomains and
deploying parallel ESNs on each group.
However, this process comes with overhead and can still lead to rather large
networks on each group.
The memory costs associated with these large networks coupled with any
additional computational costs associated with timestepping, either by
increasing the frequency or by using a more expensive method to represent small
scale processes, will likely make the ESN implementation shown here too
expensive to be considered for practical applications.
Future work could explore dimension reduction techniques involving
proper orthogonal decomposition \citep{jordanou_investigation_2022},
autoencoders \citep{heyder_generalizability_2022},
or approaches involving self-organizing or scale invariant maps
\citep{basterrech_self-organizing_2011}.
Similarly, \citet{whiteaker_reducing_2022}
show success in deriving a controllability matrix for the
ESN, which leads to a reduced network size with minimal reduction in
error.
Finally, a number of studies claim to have developed ESN architectures that
can capture dynamics occurring at many scales
\citep{moon_hierarchical_2021,ma_deepr-esn_2020,gallicchio_design_2018,gallicchio_deep_2017,malik_multilayered_2017},
and these could be
explored for geophysical turbulence emulation as well.

\section{Conclusions}
\label{sec:conclusions}

Recent advances in neural network based emulators of Earth's weather and climate
indicate that forecasting centers could benefit greatly from incorporating
neural networks into their future prediction systems.
However, a common issue with these data-driven models is that they produce
relatively blurry predictions, and misrepresent the small spatial scale features
that can be resolved in traditional, physics-based forecasting models.
Here, we showed that the simple space saving step of subsampling the training
data used to generate recurrent neural network emulators accentuates this
small scale error.
While we show some success in mitigating the effects of this subsampling
related, high wavenumber bias through an inner/outer loop optimization
framework, the problem persists.
Many neural network emulators use subsampled datasets for training, including
most prominently the ERA5 Reanalysis.
While our work suggests that there could be a benefit to using a training
dataset based on a relatively
shorter model trajectory that is not subsampled, rather than a longer one that
is, addressing the subsampling issue would provide more confidence in using
already existing, freely available datasets like reanalyses.
We therefore suggest that future work should focus on how other architectures
and techniques like
attention or adversarial training can address this subsampling related bias at
the small spatial scales of turbulent geophysical fluid dynamics.

\appendix
\section{Matrix and Data Normalization for Echo State Networks}
\label{sec:new_methods}

Here we describe several aspects of our ESN implementation that are
unique with respect to previous works.
Additionally, we provide some empirical justification for these choices, using
the Lorenz96 model as a testbed \citep{lorenz_predictability_1996}, see
Appendix \cref{subsec:lorenz96} for a description of the datasets generated for these
tests.

Our testing framework follows the general procedure laid out
by \citet{platt_systematic_2022} to evaluate the architecture choices.
For each design choice, we compute the Valid Prediction Time (VPT) of
an ESN model over 100 randomly chosen initial conditions from a test dataset.
VPT is computed as
\begin{linenomath*}\begin{equation*}
    \begin{aligned}
        \text{VPT} &= \argmin_{n} \left\{ \text{NRMSE}(n) > \epsilon \right\} \\
        \text{NRMSE}(n) &= \sqrt{\dfrac{1}{\nstate}\sum_{i=1}^{\nstate}\left(
            \dfrac{\hat{v}_i(n) - v_i(n)}{SD_i}
            \right)^2
        } \, ,
    \end{aligned}
\end{equation*}\end{linenomath*}
where $n$ is a time index, $SD_i$ is the temporal standard deviation of the $i$-th dimension,
computed from the training data, and $\epsilon=0.2$.
To eliminate the dependence of the results on the randomly chosen adjacency and
input matrices, we repeat the process for 10 different adjacency and input
matrix pairs, initialized with different random number generator seeds.
In total, we compare each design choice with a VPT distribution from 1,000 test samples.
We note that we optimize the ESN parameters listed in
\cref{eq:rc-hyperparameters} for each design choice and each random matrix pair,
following the procedure described in \cref{subsec:esn-ego} with an NRMSE cost
function.
Of course, these tests are insufficient to definitively prove that these choices
will translate perfectly to the SQG system.
However, we consider this to be a bare minimum test that will catch downright bad
design choices, while saving the computing resources necessary to train an
emulator for larger problems.

\subsection{Input Matrix Scaling}
\label{subsec:input-scaling}

Typically, $\inputmatrix$ is filled with entries
\begin{linenomath*}\begin{equation*}
    \hat{w}_{i,j} \sim \mathcal{U}(-\sigma,\sigma) \qquad
    i = \{1, 2, ..., \nhidden\}, j=\{1,2, ..., \ninputstate\} \,
\end{equation*}\end{linenomath*}
where $\sigma$ determines the bounds of the uniform
distribution.
Here we found it to be advantageous to normalize the input matrix by the
largest singular value.
That is, we first compute $\hat{\mathbf{W}}_\text{in}$, with elements
\begin{linenomath*}\begin{equation*}
    \hat{w}_{i,j} \sim \mathcal{U}(-1,1) \qquad
    i = \{1, 2, ..., \nhidden\}, j=\{1,2, ..., \ninputstate\} \, .
\end{equation*}\end{linenomath*}
Then, we set $\inputmatrix$ as
\begin{linenomath*}\begin{equation*}
    \inputmatrix \coloneqq
    \dfrac{\sigma}{\sigma_{max}\left(\hat{\mathbf{W}}_\text{in}\right)}
    \hat{\mathbf{W}}_\text{in} \,
\end{equation*}\end{linenomath*}
where $\sigma_{max}\left(\cdot\right)$ is the largest singular value, and
the parameter $\sigma$ is the desired largest singular value of
$\inputmatrix$.

Our motivation for using this type of normalization is that we found it
necessary to use very wide parameter optimization bounds for $\sigma$ when
using the standard input scaling strategy.
Normalizing the matrix by the largest singular value compensates for the fact that
the amplitude of the contributions to the reservoir, i.e., the elements of the
vector
\begin{linenomath*}\begin{equation*}
    \mathbf{p} = \inputmatrix \inputstate =
    \begin{pmatrix}
        \mathbf{w}_1^T\inputstate \\
        \mathbf{w}_2^T\inputstate \\
        \vdots \\
        \mathbf{w}_{\nhidden}^T\inputstate
    \end{pmatrix}
\end{equation*}\end{linenomath*}
grow with $\ninputstate$.
By controlling for this growth, we were able to reduce the optimization search
space and achieve more consistent prediction skill with fewer iterations.

\begin{figure}
    \centering
    \includegraphics[width=.8\textwidth]{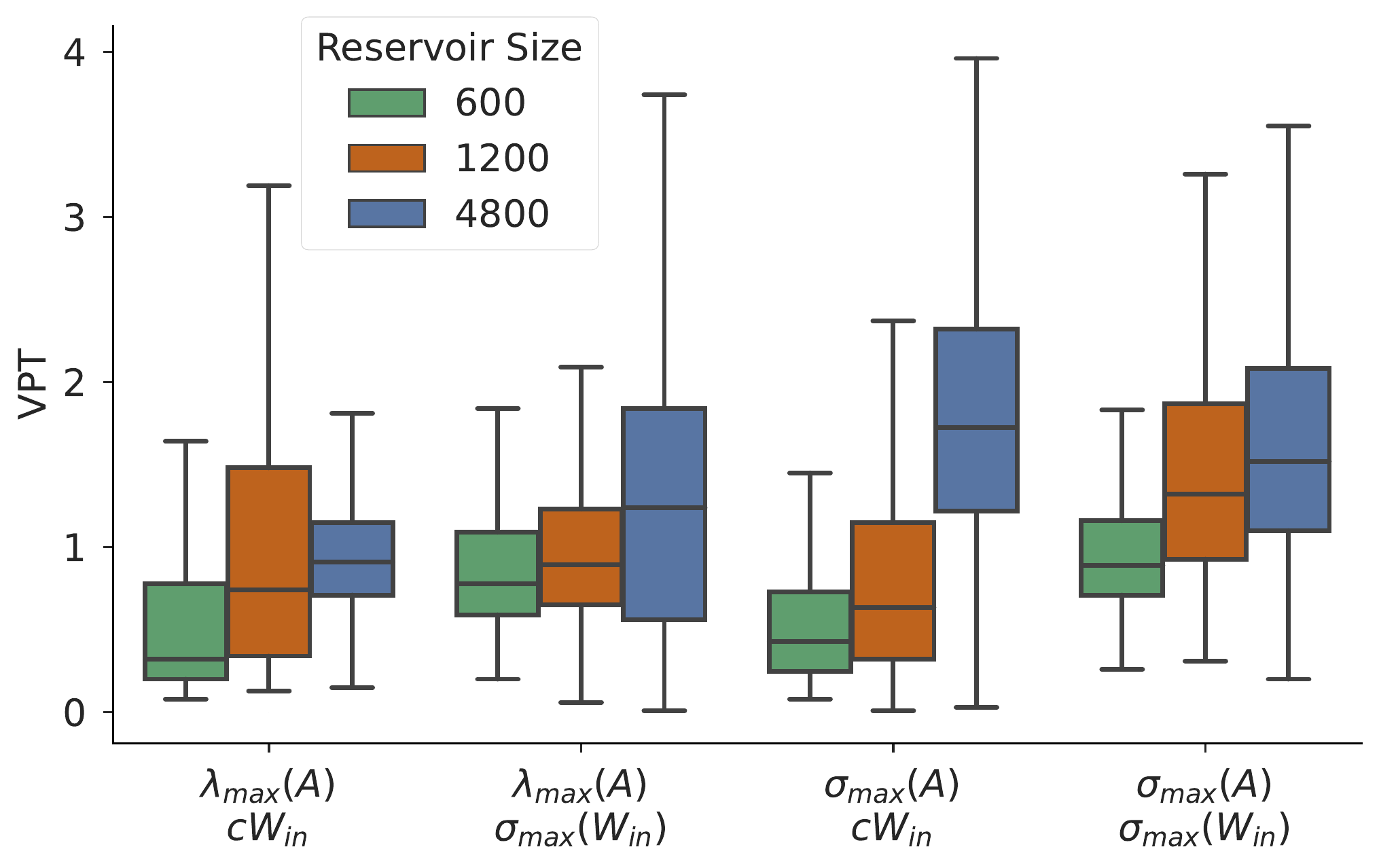}
    \caption{Valid Prediction Time (VPT) obtained with an ESN, using
        different normalization strategies for the adjacency and input matrices,
        $\adjacency$ and $\inputmatrix$. The normalization used for each matrix
        is indicated as follows:
        $\lambda_{max}(\cdot)$ refers to the largest eigenvalue (i.e.,
        spectral radius),
        $\sigma_{max}(\cdot)$ refers to the largest singular value (i.e.,
        induced 2 norm),
        while $c$ implies that no normalization was used.
        The results are computed with the 20D Lorenz96 system, described in
        Appendix \cref{subsec:lorenz96}.
        The boxplots indicate prediction skill from 10 different adjacency and
        input matrices, achieved by changing the random number generator seed,
        with 100 initial conditions randomly sampled from the test dataset for
        each set of matrices.
        The macro-scale parameters, including also the leak rate, bias, and
        Tikhonov parameter, were optimized for each unique matrix pair.
        Color indicates the size of the reservoir used.
    }
    \label{fig:simple-normalization}
\end{figure}

Additionally, we found empirical evidence to suggest that this normalization is
advantageous even for small systems.
\cref{fig:simple-normalization} shows the VPT achieved with
the 20-Dimensional Lorenz96 system (Appendix \cref{subsec:lorenz96}), using a variety of
normalization strategies for the input and adjacency matrices.
In \cref{fig:simple-normalization}, the two schemes used for the input matrix
are (1) no normalization (indicated by $c W_{in}$) and
(2) normalization by the largest singular value (indicated by
$\sigma_{max}(W_{in})$).
For a variety of reservoir sizes, $\nhidden$, we found that using the largest
singular value often performed better, usually by about 0.5~MTU.

\subsection{Adjacency Matrix Scaling}
\label{subsec:adjacency-scaling}

Typically, the reservoir adjacency matrix is normalized to achieve a desired
spectral radius.
That is, the matrix $\hat{\adjacency}$ is generated with elements
$\hat{a}_{i,j} \sim \mathcal{U}(-1,1)$, where $i,j$ are random indices in order
to satisfy the desired sparsity of the matrix (all other elements are 0).
Then, $\adjacency$ is set as
\begin{linenomath*}\begin{equation*}
    \adjacency \coloneqq
    \dfrac{\spectralradius}{\lambda_{max}\left(\hat{\adjacency}\right)}
    \hat{\adjacency} \, ,
\end{equation*}\end{linenomath*}
where $\lambda_{max}\left(\cdot\right)$ is the spectral radius, and $\spectralradius$
scales the matrix to achieve the desired spectral radius.
A common guideline is to set $\spectralradius \simeq 1$, as it is hypothesized
that this puts the reservoir on the ``edge of stability'' so that it performs
well in emulating nonlinear systems \citep<e.g., as recommended by>[]{lukosevicius_practical_2012}.
However, as originally described by \citet{jaeger_echo_2001},
the spectral radius provides only a necessary, but insufficient, means to satisfy
the required Echo State Property.
On the other hand, using the largest singular value is a sufficient condition
for satisfying the echo state property.

In our experimentation, we have found a slight benefit from using the largest
singular value to normalize the adjacency matrix.
\cref{fig:simple-normalization} shows that, for fixed input matrix
normalization, using the largest singular value rather than spectral radius
achieves similar and up to $\sim0.3$ longer valid predictions.
While the improvement may seem subtle, we note that using the largest singular value
has the following practical benefit for our python-based implementation:
the singular values can be computed directly on a Graphical Processing Unit
using CuPy \citep{cupy_learningsys2017}, while a general, non-symmetric eigenvalue
decomposition is not readily available.

\subsection{Data Normalization}
\label{subsec:data_normalization}

A key aspect in machine learning is normalizing input data before passing it to
the model.
Experiments from \citet{platt_systematic_2022} showed, however, that the
standard approach to normalizing data
can be detrimental to prediction skill.
By ``standard  approach'', we mean
\begin{linenomath*}\begin{equation*}
    v_i(n) = \dfrac{v_i(n) - \bar{v}_i}{SD_i} \qquad
    i = \{1, 2, ... \nstate\} \, ,
\end{equation*}\end{linenomath*}
where
\begin{linenomath*}\begin{equation*}
    \bar{v}_i =
    \dfrac{1}{\ntrain}\sum_{n=1}^{\ntrain} v_i(n) \, ,
    \qquad\qquad
    SD_i = \dfrac{1}{\ntrain-1}\sqrt{\sum_{n=1}^{\ntrain}\left(v_i(n) - \bar{v}_i\right)^2}
\end{equation*}\end{linenomath*}
i.e., $\bar{v}_i$ and $SD_i$ are the mean and standard deviation taken from the
training data separately over each channel of data, indexed by $i$.
The key takeaway from \citet{platt_systematic_2022} is that by using separate
normalization values for each channel, the covarying relationships between the
data are destroyed and the reservoir cannot learn the true dynamics.
The authors propose to normalize with the average and range of the data,
computed over the length of the training data and over all channels
\begin{linenomath*}\begin{equation}
    v_i(n) = \dfrac{v_i(n) - \mu}{
        v_\text{max} - v_\text{min}}
    \qquad i = \{1,2, ... \nstate\} \, ,
    \label{eq:maxmin}
\end{equation}\end{linenomath*}
where
\begin{linenomath*}\begin{equation}
    \mu =
    \dfrac{1}{\nstate}\sum_{i=1}^{\nstate}\bar{v}_i \, ,
    \qquad
    v_\text{max} = \max_{
        \substack{
            i=\{1, ..., \nstate\}
            \\
            n=\{1, ..., \ntrain\}}}
            \Big(v_i(n)\Big)\, ,
    \qquad
    v_\text{min} = \min_{
        \substack{
            i=\{1, ..., \nstate\}
            \\
            n=\{1, ..., \ntrain\}}}
            \Big(v_i(n)\Big) \, .
\end{equation}\end{linenomath*}
Here, we propose to replace the range in the denominator with the
standard deviation computed over all channels and timesteps in the training
data,
\begin{linenomath*}\begin{equation}
    v_i(n) = \dfrac{v_i(n) - \mu}{SD}
    \qquad i = \{1,2, ... \nstate\} \, ,
    \label{eq:sdnorm}
\end{equation}\end{linenomath*}
with
\begin{linenomath*}\begin{equation*}
    SD = \dfrac{1}{(\ntrain-1)(\nstate-1)}
    \sqrt{
        \sum_{i=1}^{\nstate}\sum_{n=1}^{\ntrain}\left(v_i(n) - \mu\right)^2
    }\, .
\end{equation*}\end{linenomath*}

\begin{figure}
    \centering
    \includegraphics[width=.4\textwidth]{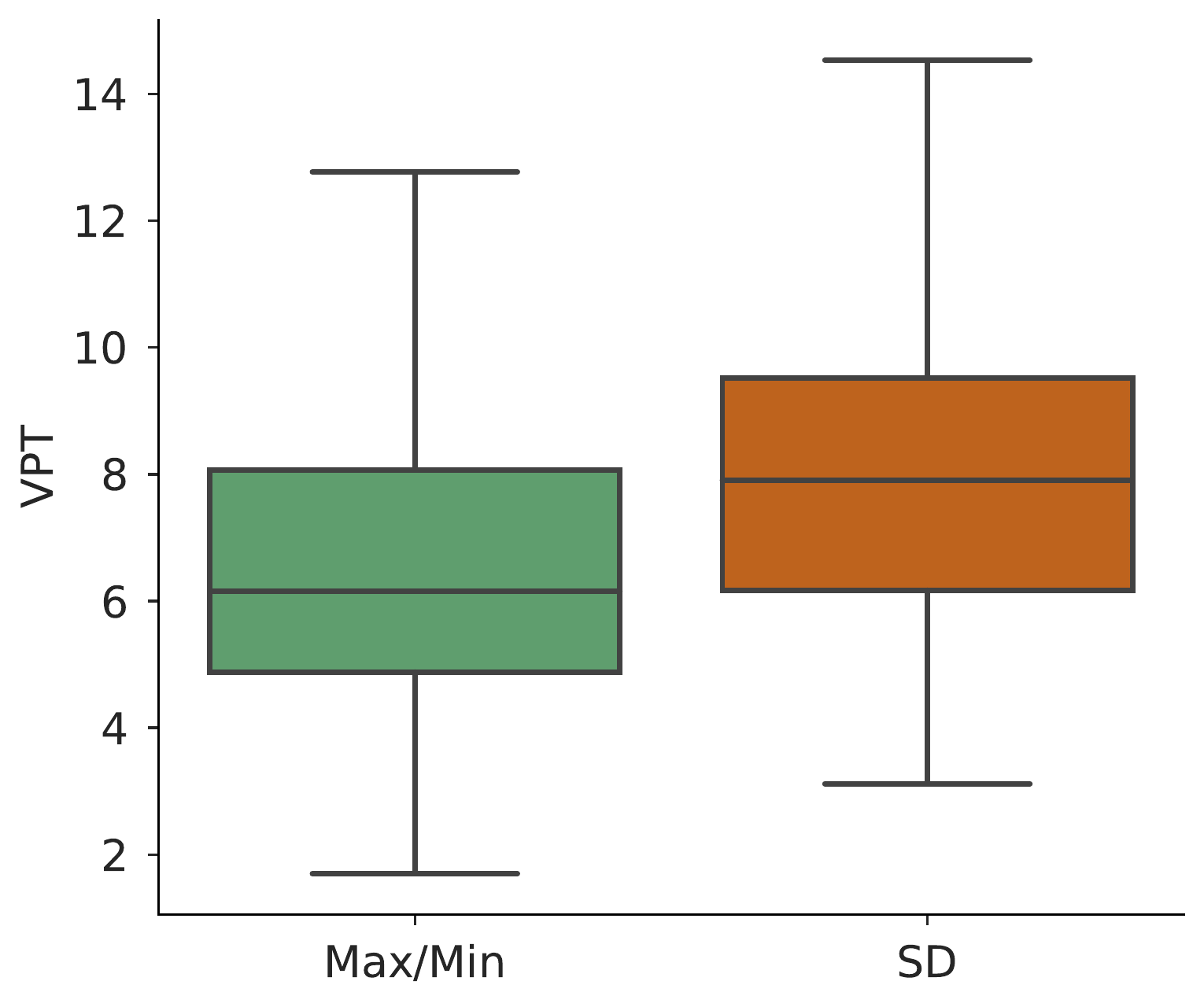}
    \caption{Valid Prediction Time (VPT) with an ESN, using the
        Max/Min normalization strategy shown in \cref{eq:maxmin} and standard
        deviation (SD)
        normalization strategy as in \cref{eq:sdnorm}.
        The results are computed with the 6D Lorenz96 system, described in
        Appendix \cref{subsec:lorenz96}.
        The boxplots indicate prediction skill from 10 different adjacency and
        input matricesm, achieved by changing the random number generator seed,
        with 100 initial conditions randomly sampled from the test dataset for
        each set of matrices.
        All macro-scale parameters were optimized for each unique matrix pair.
    }
    \label{fig:data-norm}
\end{figure}

\cref{fig:data-norm} compares the prediction skill when these two normalization
strategies are used.
Using the standard deviation normalization as in \cref{eq:sdnorm} leads to an
average VPT increase of 2~MTU.
We suggest that this improvement is due to the fact that when the data are
normalized by the full range, then all values are in the range $[-1,1]$.
In this case, once the input is mapped into the hidden space, it is more likely
to lie on the linear regime of the $\tanh(\cdot)$ activation function.
While a large enough input scaling could eliminate this problem, it is
apparently not easily obtained during the Bayesian optimization.

\subsection{Lorenz96 Datasets}
\label{subsec:lorenz96}

The Lorenz96 dataset used for these supplemental experiments were generated by
the following set of equations introduced by \citet{lorenz_predictability_1996},
\begin{linenomath*}\begin{equation*}
    \frac{dv_i(t)}{dt} = v_{i-1}(t)(v_{i+1}(t) - v_{i-2}(t)) - v_i(t) + F \, ,
    \label{eq:lorenz96}
\end{equation*}\end{linenomath*}
where $i=1,2,...,N_{l}$, and the domain is periodic.
$F=8$ is a fixed parameter that generates chaotic dynamics.
We use $N_l = 20$ for the tests in
Appendices \cref{subsec:input-scaling,subsec:adjacency-scaling} and $N_l = 6$ for the tests
in Appendix \cref{subsec:data_normalization}.
Each dataset was generated by stepping the model forward with a 4th order
Runge-Kutta scheme with $\Delta t = 0.01$~Model~Time~Units (MTU).
Each dataset consisted of a 10~MTU spinup period that was discarded, 420~MTU of
training data, a 60~MTU validation period, and a 120~MTU test
period.
Each randomly chosen validation and test trajectory were 1~MTU and 15~MTU,
respectively, and the ESN spinup period was 5~MTU.

\section{Gulf of Mexico Dataset and ESN Prediction}
\label{sec:gom}

The Gulf of Mexico reanalysis dataset used to generate the prediction in
\cref{fig:gom_sst} was provided by \citet{gom_dataset}.
The data consists of 6~hourly snapshots of 2D sea surface height and 3D
temperature, salinity, and zonal and meridional velocities,
covering 1993-2012 (inclusive).
We used only the top level of temperature, and used the first 18~years as
training, and the last two years as test data.
Here we apply a parallelized ESN architecture, using
$\nlocalx=\nlocaly=4$, $\noverlap=1$, and $\nhidden=6,000$.
Because we use only the top level of temperature, $\nvertical=1$, and therefore
$\nlocalinputstate=25$, $\nlocalstate=16$.
The grid cells that represent continental land are ignored in the input and
output vectors, and in the corresponding rows of $\Wout$.
Therefore, the effect of the boundary conditions on the neighboring grid cells
is implicitly learned from the data.

\section*{Open Research}

The model configurations used to generate the results in this manuscript can be
found at \citet{smith_rcgfd_2023}.


\acknowledgments

T.A. Smith and S.G. Penny acknowledge support from NOAA grant NA20OAR4600277.
S.G. Penny and J.A. Platt acknowledge support from the Office of Naval Research
(ONR) grants N00014-19-1-2522 and N00014-20-1-2580.
T.A. Smith thanks Nora Loose for comments and discussion that improved the manuscript.
The authors thank three anonymous reviewers and the handling editor for comments
that improved the manuscript.


%
\bibliography{references}

\begin{thebibliography}{}

\bibitem [\protect \citeauthoryear {%
Agarwal%
, Kondrashov%
, Dueben%
, Ryzhov%
\BCBL {}\ \BBA {} Berloff%
}{%
Agarwal%
\ \protect \BOthers {.}}{%
{\protect \APACyear {2021}}%
}]{%
agarwal_comparison_2021}
\APACinsertmetastar {%
agarwal_comparison_2021}%
\begin{APACrefauthors}%
Agarwal, N.%
, Kondrashov, D.%
, Dueben, P.%
, Ryzhov, E.%
\BCBL {}\ \BBA {} Berloff, P.%
\end{APACrefauthors}%
\unskip\
\newblock
\APACrefYearMonthDay{2021}{}{}.
\newblock
{\BBOQ}\APACrefatitle {A {Comparison} of {Data}-{Driven} {Approaches} to
  {Build} {Low}-{Dimensional} {Ocean} {Models}} {A {Comparison} of
  {Data}-{Driven} {Approaches} to {Build} {Low}-{Dimensional} {Ocean}
  {Models}}.{\BBCQ}
\newblock
\APACjournalVolNumPages{Journal of Advances in Modeling Earth
  Systems}{13}{9}{e2021MS002537}.
\newblock
\begin{APACrefURL}
  [{2021-12-01}]\url{http://onlinelibrary.wiley.com/doi/abs/10.1029/2021MS002537}
  \end{APACrefURL}
\newblock
\APACrefnote{\_eprint:
  https://agupubs.onlinelibrary.wiley.com/doi/pdf/10.1029/2021MS002537}
\newblock
\begin{APACrefDOI} \doi{10.1029/2021MS002537} \end{APACrefDOI}
\PrintBackRefs{\CurrentBib}

\bibitem [\protect \citeauthoryear {%
Arcomano%
\ \protect \BOthers {.}}{%
Arcomano%
\ \protect \BOthers {.}}{%
{\protect \APACyear {2020}}%
}]{%
arcomano_machine_2020}
\APACinsertmetastar {%
arcomano_machine_2020}%
\begin{APACrefauthors}%
Arcomano, T.%
, Szunyogh, I.%
, Pathak, J.%
, Wikner, A.%
, Hunt, B\BPBI R.%
\BCBL {}\ \BBA {} Ott, E.%
\end{APACrefauthors}%
\unskip\
\newblock
\APACrefYearMonthDay{2020}{}{}.
\newblock
{\BBOQ}\APACrefatitle {A {Machine} {Learning}-{Based} {Global} {Atmospheric}
  {Forecast} {Model}} {A {Machine} {Learning}-{Based} {Global} {Atmospheric}
  {Forecast} {Model}}.{\BBCQ}
\newblock
\APACjournalVolNumPages{Geophysical Research Letters}{47}{9}{e2020GL087776}.
\newblock
\begin{APACrefURL}
  [{2020-05-14}]\url{http://agupubs.onlinelibrary.wiley.com/doi/abs/10.1029/2020GL087776}
  \end{APACrefURL}
\newblock
\APACrefnote{\_eprint:
  https://onlinelibrary.wiley.com/doi/pdf/10.1029/2020GL087776}
\newblock
\begin{APACrefDOI} \doi{10.1029/2020GL087776} \end{APACrefDOI}
\PrintBackRefs{\CurrentBib}

\bibitem [\protect \citeauthoryear {%
Barbosa%
\ \BBA {} Gauthier%
}{%
Barbosa%
\ \BBA {} Gauthier%
}{%
{\protect \APACyear {2022}}%
}]{%
barbosa_learning_2022}
\APACinsertmetastar {%
barbosa_learning_2022}%
\begin{APACrefauthors}%
Barbosa, W\BPBI A\BPBI S.%
\BCBT {}\ \BBA {} Gauthier, D\BPBI J.%
\end{APACrefauthors}%
\unskip\
\newblock
\APACrefYearMonthDay{2022}{{\APACmonth{03}}}{}.
\newblock
{\BBOQ}\APACrefatitle {Learning {Spatiotemporal} {Chaos} {Using}
  {Next}-{Generation} {Reservoir} {Computing}} {Learning {Spatiotemporal}
  {Chaos} {Using} {Next}-{Generation} {Reservoir} {Computing}}.{\BBCQ}
\newblock
\APACjournalVolNumPages{arXiv:2203.13294 [nlin]}{}{}{}.
\newblock
\begin{APACrefURL} [{2022-04-04}]\url{http://arxiv.org/abs/2203.13294}
  \end{APACrefURL}
\newblock
\APACrefnote{arXiv: 2203.13294}
\PrintBackRefs{\CurrentBib}

\bibitem [\protect \citeauthoryear {%
Basterrech%
, Fyfe%
\BCBL {}\ \BBA {} Rubino%
}{%
Basterrech%
\ \protect \BOthers {.}}{%
{\protect \APACyear {2011}}%
}]{%
basterrech_self-organizing_2011}
\APACinsertmetastar {%
basterrech_self-organizing_2011}%
\begin{APACrefauthors}%
Basterrech, S.%
, Fyfe, C.%
\BCBL {}\ \BBA {} Rubino, G.%
\end{APACrefauthors}%
\unskip\
\newblock
\APACrefYearMonthDay{2011}{{\APACmonth{11}}}{}.
\newblock
{\BBOQ}\APACrefatitle {Self-{Organizing} {Maps} and {Scale}-{Invariant} {Maps}
  in {Echo} {State} {Networks}} {Self-{Organizing} {Maps} and
  {Scale}-{Invariant} {Maps} in {Echo} {State} {Networks}}.{\BBCQ}
\newblock
\BIn{} \APACrefbtitle {2011 11th {International} {Conference} on {Intelligent}
  {Systems} {Design} and {Applications}} {2011 11th {International}
  {Conference} on {Intelligent} {Systems} {Design} and {Applications}}\ (\BPGS\
  94--99).
\newblock
\APACrefnote{ISSN: 2164-7151}
\newblock
\begin{APACrefDOI} \doi{10.1109/ISDA.2011.6121637} \end{APACrefDOI}
\PrintBackRefs{\CurrentBib}

\bibitem [\protect \citeauthoryear {%
Bi%
\ \protect \BOthers {.}}{%
Bi%
\ \protect \BOthers {.}}{%
{\protect \APACyear {2022}}%
}]{%
bi_pangu-weather_2022}
\APACinsertmetastar {%
bi_pangu-weather_2022}%
\begin{APACrefauthors}%
Bi, K.%
, Xie, L.%
, Zhang, H.%
, Chen, X.%
, Gu, X.%
\BCBL {}\ \BBA {} Tian, Q.%
\end{APACrefauthors}%
\unskip\
\newblock
\APACrefYearMonthDay{2022}{{\APACmonth{11}}}{}.
\newblock
\APACrefbtitle {Pangu-{Weather}: {A} {3D} {High}-{Resolution} {Model} for
  {Fast} and {Accurate} {Global} {Weather} {Forecast}.} {Pangu-{Weather}: {A}
  {3D} {High}-{Resolution} {Model} for {Fast} and {Accurate} {Global} {Weather}
  {Forecast}.}
\newblock
\APACaddressPublisher{}{arXiv}.
\newblock
\begin{APACrefURL} [{2022-11-23}]\url{http://arxiv.org/abs/2211.02556}
  \end{APACrefURL}
\newblock
\APACrefnote{arXiv:2211.02556 [physics]}
\PrintBackRefs{\CurrentBib}

\bibitem [\protect \citeauthoryear {%
Bi%
\ \protect \BOthers {.}}{%
Bi%
\ \protect \BOthers {.}}{%
{\protect \APACyear {2023}}%
}]{%
bi_accurate_2023}
\APACinsertmetastar {%
bi_accurate_2023}%
\begin{APACrefauthors}%
Bi, K.%
, Xie, L.%
, Zhang, H.%
, Chen, X.%
, Gu, X.%
\BCBL {}\ \BBA {} Tian, Q.%
\end{APACrefauthors}%
\unskip\
\newblock
\APACrefYearMonthDay{2023}{{\APACmonth{07}}}{}.
\newblock
{\BBOQ}\APACrefatitle {Accurate medium-range global weather forecasting with
  {3D} neural networks} {Accurate medium-range global weather forecasting with
  {3D} neural networks}.{\BBCQ}
\newblock
\APACjournalVolNumPages{Nature}{619}{7970}{533--538}.
\newblock
\begin{APACrefURL}
  [{2023-09-14}]\url{https://www.nature.com/articles/s41586-023-06185-3}
  \end{APACrefURL}
\newblock
\APACrefnote{Number: 7970 Publisher: Nature Publishing Group}
\newblock
\begin{APACrefDOI} \doi{10.1038/s41586-023-06185-3} \end{APACrefDOI}
\PrintBackRefs{\CurrentBib}

\bibitem [\protect \citeauthoryear {%
Blumen%
}{%
Blumen%
}{%
{\protect \APACyear {1978}}%
{\protect \APACexlab {{\protect \BCnt {1}}}}}]{%
blumen_uniform_1978}
\APACinsertmetastar {%
blumen_uniform_1978}%
\begin{APACrefauthors}%
Blumen, W.%
\end{APACrefauthors}%
\unskip\
\newblock
\APACrefYearMonthDay{1978{\protect \BCnt {1}}}{{\APACmonth{05}}}{}.
\newblock
{\BBOQ}\APACrefatitle {{\noopsort{A}{Uniform}} {Potential} {Vorticity} {Flow}:
  {Part} {I}. {Theory} of {Wave} {Interactions} and {Two}-{Dimensional}
  {Turbulence}} {{\noopsort{A}{Uniform}} {Potential} {Vorticity} {Flow}: {Part}
  {I}. {Theory} of {Wave} {Interactions} and {Two}-{Dimensional}
  {Turbulence}}.{\BBCQ}
\newblock
\APACjournalVolNumPages{Journal of the Atmospheric Sciences}{35}{5}{774--783}.
\newblock
\begin{APACrefURL}
  [{2021-11-24}]\url{https://journals.ametsoc.org/view/journals/atsc/35/5/1520-0469_1978_035_0774_upvfpi_2_0_co_2.xml}
  \end{APACrefURL}
\newblock
\APACrefnote{Publisher: American Meteorological Society Section: Journal of the
  Atmospheric Sciences}
\newblock
\begin{APACrefDOI} \doi{10.1175/1520-0469(1978)035<0774:UPVFPI>2.0.CO;2}
  \end{APACrefDOI}
\PrintBackRefs{\CurrentBib}

\bibitem [\protect \citeauthoryear {%
Blumen%
}{%
Blumen%
}{%
{\protect \APACyear {1978}}%
{\protect \APACexlab {{\protect \BCnt {2}}}}}]{%
blumen_uniform_1978-1}
\APACinsertmetastar {%
blumen_uniform_1978-1}%
\begin{APACrefauthors}%
Blumen, W.%
\end{APACrefauthors}%
\unskip\
\newblock
\APACrefYearMonthDay{1978{\protect \BCnt {2}}}{{\APACmonth{05}}}{}.
\newblock
{\BBOQ}\APACrefatitle {{\noopsort{B}{Uniform}} {Potential} {Vorticity} {Flow}:
  {Part} {II}. {A} {Model} of {Wave} {Interacions}} {{\noopsort{B}{Uniform}}
  {Potential} {Vorticity} {Flow}: {Part} {II}. {A} {Model} of {Wave}
  {Interacions}}.{\BBCQ}
\newblock
\APACjournalVolNumPages{Journal of the Atmospheric Sciences}{35}{5}{784--789}.
\newblock
\begin{APACrefURL}
  [{2022-12-08}]\url{https://journals.ametsoc.org/view/journals/atsc/35/5/1520-0469_1978_035_0784_upvfpi_2_0_co_2.xml}
  \end{APACrefURL}
\newblock
\APACrefnote{Publisher: American Meteorological Society Section: Journal of the
  Atmospheric Sciences}
\newblock
\begin{APACrefDOI} \doi{10.1175/1520-0469(1978)035<0784:UPVFPI>2.0.CO;2}
  \end{APACrefDOI}
\PrintBackRefs{\CurrentBib}

\bibitem [\protect \citeauthoryear {%
Bollt%
}{%
Bollt%
}{%
{\protect \APACyear {2021}}%
}]{%
bollt_explaining_2021}
\APACinsertmetastar {%
bollt_explaining_2021}%
\begin{APACrefauthors}%
Bollt, E.%
\end{APACrefauthors}%
\unskip\
\newblock
\APACrefYearMonthDay{2021}{{\APACmonth{01}}}{}.
\newblock
{\BBOQ}\APACrefatitle {On explaining the surprising success of reservoir
  computing forecaster of chaos? {The} universal machine learning dynamical
  system with contrast to {VAR} and {DMD}} {On explaining the surprising
  success of reservoir computing forecaster of chaos? {The} universal machine
  learning dynamical system with contrast to {VAR} and {DMD}}.{\BBCQ}
\newblock
\APACjournalVolNumPages{Chaos: An Interdisciplinary Journal of Nonlinear
  Science}{31}{1}{013108}.
\newblock
\begin{APACrefURL}
  [{2021-11-03}]\url{https://aip-scitation-org.colorado.idm.oclc.org/doi/10.1063/5.0024890}
  \end{APACrefURL}
\newblock
\APACrefnote{Publisher: American Institute of Physics}
\newblock
\begin{APACrefDOI} \doi{10.1063/5.0024890} \end{APACrefDOI}
\PrintBackRefs{\CurrentBib}

\bibitem [\protect \citeauthoryear {%
Bouhlel%
, He%
\BCBL {}\ \BBA {} Martins%
}{%
Bouhlel%
\ \protect \BOthers {.}}{%
{\protect \APACyear {2020}}%
}]{%
bouhlel_scalable_2020}
\APACinsertmetastar {%
bouhlel_scalable_2020}%
\begin{APACrefauthors}%
Bouhlel, M\BPBI A.%
, He, S.%
\BCBL {}\ \BBA {} Martins, J\BPBI R\BPBI R\BPBI A.%
\end{APACrefauthors}%
\unskip\
\newblock
\APACrefYearMonthDay{2020}{{\APACmonth{04}}}{}.
\newblock
{\BBOQ}\APACrefatitle {Scalable gradient–enhanced artificial neural networks
  for airfoil shape design in the subsonic and transonic regimes} {Scalable
  gradient–enhanced artificial neural networks for airfoil shape design in
  the subsonic and transonic regimes}.{\BBCQ}
\newblock
\APACjournalVolNumPages{Structural and Multidisciplinary
  Optimization}{61}{4}{1363--1376}.
\newblock
\begin{APACrefURL}
  [{2023-04-03}]\url{https://doi.org/10.1007/s00158-020-02488-5}
  \end{APACrefURL}
\newblock
\begin{APACrefDOI} \doi{10.1007/s00158-020-02488-5} \end{APACrefDOI}
\PrintBackRefs{\CurrentBib}

\bibitem [\protect \citeauthoryear {%
Bouhlel%
\ \protect \BOthers {.}}{%
Bouhlel%
\ \protect \BOthers {.}}{%
{\protect \APACyear {2019}}%
}]{%
bouhlel_python_2019}
\APACinsertmetastar {%
bouhlel_python_2019}%
\begin{APACrefauthors}%
Bouhlel, M\BPBI A.%
, Hwang, J\BPBI T.%
, Bartoli, N.%
, Lafage, R.%
, Morlier, J.%
\BCBL {}\ \BBA {} Martins, J\BPBI R\BPBI R\BPBI A.%
\end{APACrefauthors}%
\unskip\
\newblock
\APACrefYearMonthDay{2019}{{\APACmonth{09}}}{}.
\newblock
{\BBOQ}\APACrefatitle {A {Python} surrogate modeling framework with
  derivatives} {A {Python} surrogate modeling framework with
  derivatives}.{\BBCQ}
\newblock
\APACjournalVolNumPages{Advances in Engineering Software}{135}{}{102662}.
\newblock
\begin{APACrefURL}
  [{2023-02-11}]\url{https://www.sciencedirect.com/science/article/pii/S0965997818309360}
  \end{APACrefURL}
\newblock
\begin{APACrefDOI} \doi{10.1016/j.advengsoft.2019.03.005} \end{APACrefDOI}
\PrintBackRefs{\CurrentBib}

\bibitem [\protect \citeauthoryear {%
Chattopadhyay%
\ \BBA {} Hassanzadeh%
}{%
Chattopadhyay%
\ \BBA {} Hassanzadeh%
}{%
{\protect \APACyear {2023}}%
}]{%
chattopadhyay_long-term_2023}
\APACinsertmetastar {%
chattopadhyay_long-term_2023}%
\begin{APACrefauthors}%
Chattopadhyay, A.%
\BCBT {}\ \BBA {} Hassanzadeh, P.%
\end{APACrefauthors}%
\unskip\
\newblock
\APACrefYearMonthDay{2023}{{\APACmonth{04}}}{}.
\newblock
\APACrefbtitle {Long-term instabilities of deep learning-based digital twins of
  the climate system: {The} cause and a solution.} {Long-term instabilities of
  deep learning-based digital twins of the climate system: {The} cause and a
  solution.}
\newblock
\APACaddressPublisher{}{arXiv}.
\newblock
\begin{APACrefURL} [{2023-04-26}]\url{http://arxiv.org/abs/2304.07029}
  \end{APACrefURL}
\newblock
\APACrefnote{arXiv:2304.07029 [physics]}
\PrintBackRefs{\CurrentBib}

\bibitem [\protect \citeauthoryear {%
T\BHBI C.~Chen%
, Penny%
, Smith%
\BCBL {}\ \BBA {} Platt%
}{%
T\BHBI C.~Chen%
\ \protect \BOthers {.}}{%
{\protect \APACyear {2022}}%
}]{%
chen_next_2022}
\APACinsertmetastar {%
chen_next_2022}%
\begin{APACrefauthors}%
Chen, T\BHBI C.%
, Penny, S\BPBI G.%
, Smith, T\BPBI A.%
\BCBL {}\ \BBA {} Platt, J\BPBI A.%
\end{APACrefauthors}%
\unskip\
\newblock
\APACrefYearMonthDay{2022}{{\APACmonth{01}}}{}.
\newblock
\APACrefbtitle {`{Next} {Generation}' {Reservoir} {Computing}: an {Empirical}
  {Data}-{Driven} {Expression} of {Dynamical} {Equations} in {Time}-{Stepping}
  {Form}.} {`{Next} {Generation}' {Reservoir} {Computing}: an {Empirical}
  {Data}-{Driven} {Expression} of {Dynamical} {Equations} in {Time}-{Stepping}
  {Form}.}
\newblock
\APACaddressPublisher{}{arXiv}.
\newblock
\begin{APACrefURL} [{2022-12-05}]\url{http://arxiv.org/abs/2201.05193}
  \end{APACrefURL}
\newblock
\APACrefnote{arXiv:2201.05193 [cs, math]}
\newblock
\begin{APACrefDOI} \doi{10.48550/arXiv.2201.05193} \end{APACrefDOI}
\PrintBackRefs{\CurrentBib}

\bibitem [\protect \citeauthoryear {%
X.~Chen%
, Nadiga%
\BCBL {}\ \BBA {} Timofeyev%
}{%
X.~Chen%
\ \protect \BOthers {.}}{%
{\protect \APACyear {2021}}%
}]{%
chen_predicting_2021}
\APACinsertmetastar {%
chen_predicting_2021}%
\begin{APACrefauthors}%
Chen, X.%
, Nadiga, B\BPBI T.%
\BCBL {}\ \BBA {} Timofeyev, I.%
\end{APACrefauthors}%
\unskip\
\newblock
\APACrefYearMonthDay{2021}{{\APACmonth{12}}}{}.
\newblock
{\BBOQ}\APACrefatitle {Predicting {Shallow} {Water} {Dynamics} using
  {Echo}-{State} {Networks} with {Transfer} {Learning}} {Predicting {Shallow}
  {Water} {Dynamics} using {Echo}-{State} {Networks} with {Transfer}
  {Learning}}.{\BBCQ}
\newblock
\APACjournalVolNumPages{arXiv:2112.09182 [physics]}{}{}{}.
\newblock
\begin{APACrefURL} [{2022-01-11}]\url{http://arxiv.org/abs/2112.09182}
  \end{APACrefURL}
\newblock
\APACrefnote{arXiv: 2112.09182}
\PrintBackRefs{\CurrentBib}

\bibitem [\protect \citeauthoryear {%
Cressie%
}{%
Cressie%
}{%
{\protect \APACyear {1993}}%
}]{%
cressie_statistics_1993}
\APACinsertmetastar {%
cressie_statistics_1993}%
\begin{APACrefauthors}%
Cressie, N.%
\end{APACrefauthors}%
\unskip\
\newblock
\APACrefYearMonthDay{1993}{}{}.
\newblock
{\BBOQ}\APACrefatitle {Statistics for spatial data} {Statistics for spatial
  data}.{\BBCQ}
\newblock

\PrintBackRefs{\CurrentBib}

\bibitem [\protect \citeauthoryear {%
{Dask Development Team}%
}{%
{Dask Development Team}%
}{%
{\protect \APACyear {2016}}%
}]{%
dask_2016}
\APACinsertmetastar {%
dask_2016}%
\begin{APACrefauthors}%
{Dask Development Team}.%
\end{APACrefauthors}%
\unskip\
\newblock
\APACrefYearMonthDay{2016}{}{}.
\newblock
{\BBOQ}\APACrefatitle {Dask: Library for dynamic task scheduling} {Dask:
  Library for dynamic task scheduling}{\BBCQ}\ [\bibcomputersoftwaremanual].
\newblock
\begin{APACrefURL} \url{https://dask.org} \end{APACrefURL}
\PrintBackRefs{\CurrentBib}

\bibitem [\protect \citeauthoryear {%
Dosovitskiy%
\ \protect \BOthers {.}}{%
Dosovitskiy%
\ \protect \BOthers {.}}{%
{\protect \APACyear {2021}}%
}]{%
dosovitskiy_image_2021}
\APACinsertmetastar {%
dosovitskiy_image_2021}%
\begin{APACrefauthors}%
Dosovitskiy, A.%
, Beyer, L.%
, Kolesnikov, A.%
, Weissenborn, D.%
, Zhai, X.%
, Unterthiner, T.%
\BDBL {}Houlsby, N.%
\end{APACrefauthors}%
\unskip\
\newblock
\APACrefYearMonthDay{2021}{{\APACmonth{06}}}{}.
\newblock
\APACrefbtitle {An {Image} is {Worth} 16x16 {Words}: {Transformers} for {Image}
  {Recognition} at {Scale}.} {An {Image} is {Worth} 16x16 {Words}:
  {Transformers} for {Image} {Recognition} at {Scale}.}
\newblock
\APACaddressPublisher{}{arXiv}.
\newblock
\begin{APACrefURL} [{2022-11-23}]\url{http://arxiv.org/abs/2010.11929}
  \end{APACrefURL}
\newblock
\APACrefnote{arXiv:2010.11929 [cs]}
\PrintBackRefs{\CurrentBib}

\bibitem [\protect \citeauthoryear {%
Dueben%
\ \BBA {} Bauer%
}{%
Dueben%
\ \BBA {} Bauer%
}{%
{\protect \APACyear {2018}}%
}]{%
dueben_challenges_2018}
\APACinsertmetastar {%
dueben_challenges_2018}%
\begin{APACrefauthors}%
Dueben, P\BPBI D.%
\BCBT {}\ \BBA {} Bauer, P.%
\end{APACrefauthors}%
\unskip\
\newblock
\APACrefYearMonthDay{2018}{{\APACmonth{10}}}{}.
\newblock
{\BBOQ}\APACrefatitle {Challenges and design choices for global weather and
  climate models based on machine learning} {Challenges and design choices for
  global weather and climate models based on machine learning}.{\BBCQ}
\newblock
\APACjournalVolNumPages{Geoscientific Model Development}{11}{10}{3999--4009}.
\newblock
\begin{APACrefURL}
  [{2021-02-09}]\url{https://gmd.copernicus.org/articles/11/3999/2018/}
  \end{APACrefURL}
\newblock
\APACrefnote{Publisher: Copernicus GmbH}
\newblock
\begin{APACrefDOI} \doi{https://doi.org/10.5194/gmd-11-3999-2018}
  \end{APACrefDOI}
\PrintBackRefs{\CurrentBib}

\bibitem [\protect \citeauthoryear {%
Duncan%
, Subramanian%
\BCBL {}\ \BBA {} Harrington%
}{%
Duncan%
\ \protect \BOthers {.}}{%
{\protect \APACyear {2022}}%
}]{%
duncan_generative_2022}
\APACinsertmetastar {%
duncan_generative_2022}%
\begin{APACrefauthors}%
Duncan, J.%
, Subramanian, S.%
\BCBL {}\ \BBA {} Harrington, P.%
\end{APACrefauthors}%
\unskip\
\newblock
\APACrefYearMonthDay{2022}{{\APACmonth{10}}}{}.
\newblock
\APACrefbtitle {Generative {Modeling} of {High}-resolution {Global}
  {Precipitation} {Forecasts}.} {Generative {Modeling} of {High}-resolution
  {Global} {Precipitation} {Forecasts}.}
\newblock
\APACaddressPublisher{}{arXiv}.
\newblock
\begin{APACrefURL} [{2023-01-06}]\url{http://arxiv.org/abs/2210.12504}
  \end{APACrefURL}
\newblock
\APACrefnote{arXiv:2210.12504 [physics]}
\newblock
\begin{APACrefDOI} \doi{10.48550/arXiv.2210.12504} \end{APACrefDOI}
\PrintBackRefs{\CurrentBib}

\bibitem [\protect \citeauthoryear {%
Eady%
}{%
Eady%
}{%
{\protect \APACyear {1949}}%
}]{%
eady_long_1949}
\APACinsertmetastar {%
eady_long_1949}%
\begin{APACrefauthors}%
Eady, E\BPBI T.%
\end{APACrefauthors}%
\unskip\
\newblock
\APACrefYearMonthDay{1949}{}{}.
\newblock
{\BBOQ}\APACrefatitle {Long {Waves} and {Cyclone} {Waves}} {Long {Waves} and
  {Cyclone} {Waves}}.{\BBCQ}
\newblock
\APACjournalVolNumPages{Tellus}{1}{3}{33--52}.
\newblock
\begin{APACrefURL}
  [{2023-01-04}]\url{https://onlinelibrary.wiley.com/doi/abs/10.1111/j.2153-3490.1949.tb01265.x}
  \end{APACrefURL}
\newblock
\APACrefnote{\_eprint:
  https://onlinelibrary.wiley.com/doi/pdf/10.1111/j.2153-3490.1949.tb01265.x}
\newblock
\begin{APACrefDOI} \doi{10.1111/j.2153-3490.1949.tb01265.x} \end{APACrefDOI}
\PrintBackRefs{\CurrentBib}

\bibitem [\protect \citeauthoryear {%
Evensen%
, Vossepoel%
\BCBL {}\ \BBA {} van Leeuwen%
}{%
Evensen%
\ \protect \BOthers {.}}{%
{\protect \APACyear {2022}}%
}]{%
evensen_data_2022}
\APACinsertmetastar {%
evensen_data_2022}%
\begin{APACrefauthors}%
Evensen, G.%
, Vossepoel, F\BPBI C.%
\BCBL {}\ \BBA {} van Leeuwen, P\BPBI J.%
\end{APACrefauthors}%
\unskip\
\newblock
\APACrefYear{2022}.
\newblock
\APACrefbtitle {Data {Assimilation} {Fundamentals}: {A} {Unified} {Formulation}
  of the {State} and {Parameter} {Estimation} {Problem}} {Data {Assimilation}
  {Fundamentals}: {A} {Unified} {Formulation} of the {State} and {Parameter}
  {Estimation} {Problem}}.
\newblock
\APACaddressPublisher{Cham}{Springer International Publishing}.
\newblock
\begin{APACrefURL}
  [{2022-06-08}]\url{https://link.springer.com/10.1007/978-3-030-96709-3}
  \end{APACrefURL}
\newblock
\begin{APACrefDOI} \doi{10.1007/978-3-030-96709-3} \end{APACrefDOI}
\PrintBackRefs{\CurrentBib}

\bibitem [\protect \citeauthoryear {%
Gallicchio%
, Micheli%
\BCBL {}\ \BBA {} Pedrelli%
}{%
Gallicchio%
\ \protect \BOthers {.}}{%
{\protect \APACyear {2017}}%
}]{%
gallicchio_deep_2017}
\APACinsertmetastar {%
gallicchio_deep_2017}%
\begin{APACrefauthors}%
Gallicchio, C.%
, Micheli, A.%
\BCBL {}\ \BBA {} Pedrelli, L.%
\end{APACrefauthors}%
\unskip\
\newblock
\APACrefYearMonthDay{2017}{{\APACmonth{12}}}{}.
\newblock
{\BBOQ}\APACrefatitle {Deep reservoir computing: {A} critical experimental
  analysis} {Deep reservoir computing: {A} critical experimental
  analysis}.{\BBCQ}
\newblock
\APACjournalVolNumPages{Neurocomputing}{268}{}{87--99}.
\newblock
\begin{APACrefURL}
  [{2023-03-30}]\url{https://www.sciencedirect.com/science/article/pii/S0925231217307567}
  \end{APACrefURL}
\newblock
\begin{APACrefDOI} \doi{10.1016/j.neucom.2016.12.089} \end{APACrefDOI}
\PrintBackRefs{\CurrentBib}

\bibitem [\protect \citeauthoryear {%
Gallicchio%
, Micheli%
\BCBL {}\ \BBA {} Pedrelli%
}{%
Gallicchio%
\ \protect \BOthers {.}}{%
{\protect \APACyear {2018}}%
}]{%
gallicchio_design_2018}
\APACinsertmetastar {%
gallicchio_design_2018}%
\begin{APACrefauthors}%
Gallicchio, C.%
, Micheli, A.%
\BCBL {}\ \BBA {} Pedrelli, L.%
\end{APACrefauthors}%
\unskip\
\newblock
\APACrefYearMonthDay{2018}{{\APACmonth{12}}}{}.
\newblock
{\BBOQ}\APACrefatitle {Design of deep echo state networks} {Design of deep echo
  state networks}.{\BBCQ}
\newblock
\APACjournalVolNumPages{Neural Networks}{108}{}{33--47}.
\newblock
\begin{APACrefURL}
  [{2023-03-30}]\url{https://www.sciencedirect.com/science/article/pii/S0893608018302223}
  \end{APACrefURL}
\newblock
\begin{APACrefDOI} \doi{10.1016/j.neunet.2018.08.002} \end{APACrefDOI}
\PrintBackRefs{\CurrentBib}

\bibitem [\protect \citeauthoryear {%
Gauthier%
, Bollt%
, Griffith%
\BCBL {}\ \BBA {} Barbosa%
}{%
Gauthier%
\ \protect \BOthers {.}}{%
{\protect \APACyear {2021}}%
}]{%
gauthier_next_2021}
\APACinsertmetastar {%
gauthier_next_2021}%
\begin{APACrefauthors}%
Gauthier, D\BPBI J.%
, Bollt, E.%
, Griffith, A.%
\BCBL {}\ \BBA {} Barbosa, W\BPBI A\BPBI S.%
\end{APACrefauthors}%
\unskip\
\newblock
\APACrefYearMonthDay{2021}{{\APACmonth{09}}}{}.
\newblock
{\BBOQ}\APACrefatitle {Next generation reservoir computing} {Next generation
  reservoir computing}.{\BBCQ}
\newblock
\APACjournalVolNumPages{Nature Communications}{12}{1}{5564}.
\newblock
\begin{APACrefURL}
  [{2021-10-21}]\url{https://www.nature.com/articles/s41467-021-25801-2}
  \end{APACrefURL}
\newblock
\APACrefnote{Bandiera\_abtest: a Cc\_license\_type: cc\_by Cg\_type: Nature
  Research Journals Number: 1 Primary\_atype: Research Publisher: Nature
  Publishing Group Subject\_term: Computational science;Electrical and
  electronic engineering Subject\_term\_id:
  computational-science;electrical-and-electronic-engineering}
\newblock
\begin{APACrefDOI} \doi{10.1038/s41467-021-25801-2} \end{APACrefDOI}
\PrintBackRefs{\CurrentBib}

\bibitem [\protect \citeauthoryear {%
Goodfellow%
, Yoshua%
\BCBL {}\ \BBA {} Aaron%
}{%
Goodfellow%
\ \protect \BOthers {.}}{%
{\protect \APACyear {2016}}%
}]{%
goodfellow_sequence_2016}
\APACinsertmetastar {%
goodfellow_sequence_2016}%
\begin{APACrefauthors}%
Goodfellow, I.%
, Yoshua, B.%
\BCBL {}\ \BBA {} Aaron, C.%
\end{APACrefauthors}%
\unskip\
\newblock
\APACrefYearMonthDay{2016}{}{}.
\newblock
{\BBOQ}\APACrefatitle {Sequence {Modeling}: {Recurrent} and {Recursive} {Nets}}
  {Sequence {Modeling}: {Recurrent} and {Recursive} {Nets}}.{\BBCQ}
\newblock
\BIn{} \APACrefbtitle {Deep {Learning}.} {Deep {Learning}.}
\newblock
\APACaddressPublisher{}{MIT Press}.
\newblock
\begin{APACrefURL} \url{https://www.deeplearningbook.org/} \end{APACrefURL}
\PrintBackRefs{\CurrentBib}

\bibitem [\protect \citeauthoryear {%
Griffith%
, Pomerance%
\BCBL {}\ \BBA {} Gauthier%
}{%
Griffith%
\ \protect \BOthers {.}}{%
{\protect \APACyear {2019}}%
}]{%
griffith_forecasting_2019}
\APACinsertmetastar {%
griffith_forecasting_2019}%
\begin{APACrefauthors}%
Griffith, A.%
, Pomerance, A.%
\BCBL {}\ \BBA {} Gauthier, D\BPBI J.%
\end{APACrefauthors}%
\unskip\
\newblock
\APACrefYearMonthDay{2019}{{\APACmonth{12}}}{}.
\newblock
{\BBOQ}\APACrefatitle {Forecasting chaotic systems with very low connectivity
  reservoir computers} {Forecasting chaotic systems with very low connectivity
  reservoir computers}.{\BBCQ}
\newblock
\APACjournalVolNumPages{Chaos: An Interdisciplinary Journal of Nonlinear
  Science}{29}{12}{123108}.
\newblock
\begin{APACrefURL}
  [{2021-11-29}]\url{https://aip.scitation.org/doi/10.1063/1.5120710}
  \end{APACrefURL}
\newblock
\APACrefnote{Publisher: American Institute of Physics}
\newblock
\begin{APACrefDOI} \doi{10.1063/1.5120710} \end{APACrefDOI}
\PrintBackRefs{\CurrentBib}

\bibitem [\protect \citeauthoryear {%
Hasselmann%
}{%
Hasselmann%
}{%
{\protect \APACyear {1988}}%
}]{%
hasselmann_pips_1988}
\APACinsertmetastar {%
hasselmann_pips_1988}%
\begin{APACrefauthors}%
Hasselmann, K.%
\end{APACrefauthors}%
\unskip\
\newblock
\APACrefYearMonthDay{1988}{}{}.
\newblock
{\BBOQ}\APACrefatitle {{PIPs} and {POPs}: {The} reduction of complex dynamical
  systems using principal interaction and oscillation patterns} {{PIPs} and
  {POPs}: {The} reduction of complex dynamical systems using principal
  interaction and oscillation patterns}.{\BBCQ}
\newblock
\APACjournalVolNumPages{Journal of Geophysical Research:
  Atmospheres}{93}{D9}{11015--11021}.
\newblock
\begin{APACrefURL}
  [{2022-10-11}]\url{https://onlinelibrary.wiley.com/doi/abs/10.1029/JD093iD09p11015}
  \end{APACrefURL}
\newblock
\APACrefnote{\_eprint:
  https://onlinelibrary.wiley.com/doi/pdf/10.1029/JD093iD09p11015}
\newblock
\begin{APACrefDOI} \doi{10.1029/JD093iD09p11015} \end{APACrefDOI}
\PrintBackRefs{\CurrentBib}

\bibitem [\protect \citeauthoryear {%
Held%
, Pierrehumbert%
, Garner%
\BCBL {}\ \BBA {} Swanson%
}{%
Held%
\ \protect \BOthers {.}}{%
{\protect \APACyear {1995}}%
}]{%
held_surface_1995}
\APACinsertmetastar {%
held_surface_1995}%
\begin{APACrefauthors}%
Held, I\BPBI M.%
, Pierrehumbert, R\BPBI T.%
, Garner, S\BPBI T.%
\BCBL {}\ \BBA {} Swanson, K\BPBI L.%
\end{APACrefauthors}%
\unskip\
\newblock
\APACrefYearMonthDay{1995}{{\APACmonth{01}}}{}.
\newblock
{\BBOQ}\APACrefatitle {Surface quasi-geostrophic dynamics} {Surface
  quasi-geostrophic dynamics}.{\BBCQ}
\newblock
\APACjournalVolNumPages{Journal of Fluid Mechanics}{282}{}{1--20}.
\newblock
\begin{APACrefURL}
  [{2021-11-24}]\url{http://www.cambridge.org/core/journals/journal-of-fluid-mechanics/article/surface-quasigeostrophic-dynamics/81CC9FC82F189A1E59E7816F47D3260F}
  \end{APACrefURL}
\newblock
\APACrefnote{Publisher: Cambridge University Press}
\newblock
\begin{APACrefDOI} \doi{10.1017/S0022112095000012} \end{APACrefDOI}
\PrintBackRefs{\CurrentBib}

\bibitem [\protect \citeauthoryear {%
Hermans%
\ \BBA {} Schrauwen%
}{%
Hermans%
\ \BBA {} Schrauwen%
}{%
{\protect \APACyear {2010}}%
}]{%
hermans_memory_2010}
\APACinsertmetastar {%
hermans_memory_2010}%
\begin{APACrefauthors}%
Hermans, M.%
\BCBT {}\ \BBA {} Schrauwen, B.%
\end{APACrefauthors}%
\unskip\
\newblock
\APACrefYearMonthDay{2010}{{\APACmonth{07}}}{}.
\newblock
{\BBOQ}\APACrefatitle {Memory in reservoirs for high dimensional input} {Memory
  in reservoirs for high dimensional input}.{\BBCQ}
\newblock
\BIn{} \APACrefbtitle {The 2010 {International} {Joint} {Conference} on
  {Neural} {Networks} ({IJCNN})} {The 2010 {International} {Joint} {Conference}
  on {Neural} {Networks} ({IJCNN})}\ (\BPGS\ 1--7).
\newblock
\APACaddressPublisher{Barcelona, Spain}{IEEE}.
\newblock
\begin{APACrefURL}
  [{2022-03-18}]\url{http://ieeexplore.ieee.org/document/5596884/}
  \end{APACrefURL}
\newblock
\begin{APACrefDOI} \doi{10.1109/IJCNN.2010.5596884} \end{APACrefDOI}
\PrintBackRefs{\CurrentBib}

\bibitem [\protect \citeauthoryear {%
Hersbach%
\ \protect \BOthers {.}}{%
Hersbach%
\ \protect \BOthers {.}}{%
{\protect \APACyear {2020}}%
}]{%
hersbach_era5_2020}
\APACinsertmetastar {%
hersbach_era5_2020}%
\begin{APACrefauthors}%
Hersbach, H.%
, Bell, B.%
, Berrisford, P.%
, Hirahara, S.%
, Horányi, A.%
, Muñoz-Sabater, J.%
\BDBL {}Thépaut, J\BHBI N.%
\end{APACrefauthors}%
\unskip\
\newblock
\APACrefYearMonthDay{2020}{}{}.
\newblock
{\BBOQ}\APACrefatitle {The {ERA5} global reanalysis} {The {ERA5} global
  reanalysis}.{\BBCQ}
\newblock
\APACjournalVolNumPages{Quarterly Journal of the Royal Meteorological
  Society}{146}{730}{1999--2049}.
\newblock
\begin{APACrefURL}
  [{2022-10-11}]\url{https://onlinelibrary.wiley.com/doi/abs/10.1002/qj.3803}
  \end{APACrefURL}
\newblock
\APACrefnote{\_eprint: https://onlinelibrary.wiley.com/doi/pdf/10.1002/qj.3803}
\newblock
\begin{APACrefDOI} \doi{10.1002/qj.3803} \end{APACrefDOI}
\PrintBackRefs{\CurrentBib}

\bibitem [\protect \citeauthoryear {%
Hewitt%
\ \protect \BOthers {.}}{%
Hewitt%
\ \protect \BOthers {.}}{%
{\protect \APACyear {2016}}%
}]{%
hewitt_impact_2016}
\APACinsertmetastar {%
hewitt_impact_2016}%
\begin{APACrefauthors}%
Hewitt, H\BPBI T.%
, Roberts, M\BPBI J.%
, Hyder, P.%
, Graham, T.%
, Rae, J.%
, Belcher, S\BPBI E.%
\BDBL {}Wood, R\BPBI A.%
\end{APACrefauthors}%
\unskip\
\newblock
\APACrefYearMonthDay{2016}{{\APACmonth{10}}}{}.
\newblock
{\BBOQ}\APACrefatitle {The impact of resolving the {Rossby} radius at
  mid-latitudes in the ocean: results from a high-resolution version of the
  {Met} {Office} {GC2} coupled model} {The impact of resolving the {Rossby}
  radius at mid-latitudes in the ocean: results from a high-resolution version
  of the {Met} {Office} {GC2} coupled model}.{\BBCQ}
\newblock
\APACjournalVolNumPages{Geoscientific Model Development}{9}{10}{3655--3670}.
\newblock
\begin{APACrefURL}
  [{2022-02-08}]\url{https://gmd.copernicus.org/articles/9/3655/2016/}
  \end{APACrefURL}
\newblock
\APACrefnote{Publisher: Copernicus GmbH}
\newblock
\begin{APACrefDOI} \doi{10.5194/gmd-9-3655-2016} \end{APACrefDOI}
\PrintBackRefs{\CurrentBib}

\bibitem [\protect \citeauthoryear {%
Heyder%
, Mellado%
\BCBL {}\ \BBA {} Schumacher%
}{%
Heyder%
\ \protect \BOthers {.}}{%
{\protect \APACyear {2022}}%
}]{%
heyder_generalizability_2022}
\APACinsertmetastar {%
heyder_generalizability_2022}%
\begin{APACrefauthors}%
Heyder, F.%
, Mellado, J\BPBI P.%
\BCBL {}\ \BBA {} Schumacher, J.%
\end{APACrefauthors}%
\unskip\
\newblock
\APACrefYearMonthDay{2022}{{\APACmonth{11}}}{}.
\newblock
{\BBOQ}\APACrefatitle {Generalizability of reservoir computing for flux-driven
  two-dimensional convection} {Generalizability of reservoir computing for
  flux-driven two-dimensional convection}.{\BBCQ}
\newblock
\APACjournalVolNumPages{Physical Review E}{106}{5}{055303}.
\newblock
\begin{APACrefURL}
  [{2022-11-23}]\url{https://link.aps.org/doi/10.1103/PhysRevE.106.055303}
  \end{APACrefURL}
\newblock
\APACrefnote{Publisher: American Physical Society}
\newblock
\begin{APACrefDOI} \doi{10.1103/PhysRevE.106.055303} \end{APACrefDOI}
\PrintBackRefs{\CurrentBib}

\bibitem [\protect \citeauthoryear {%
{HYCOM}%
}{%
{HYCOM}%
}{%
{\protect \APACyear {2016}}%
}]{%
gom_dataset}
\APACinsertmetastar {%
gom_dataset}%
\begin{APACrefauthors}%
{HYCOM}.%
\end{APACrefauthors}%
\unskip\
\newblock
\APACrefYearMonthDay{2016}{}{}.
\newblock
\APACrefbtitle {{HYCOM} + {NCODA} {Gulf} of {Mexico} {1/25°} {Reanalysis},
  ({GOM}u0.04/expt{\_}50.1).} {{HYCOM} + {NCODA} {Gulf} of {Mexico} {1/25°}
  {Reanalysis}, ({GOM}u0.04/expt{\_}50.1).}
\newblock
\APACrefnote{Data retrieved from HYCOM,
  \url{https://www.hycom.org/data/gomu0pt04/expt-50pt1}}
\PrintBackRefs{\CurrentBib}

\bibitem [\protect \citeauthoryear {%
Jaeger%
}{%
Jaeger%
}{%
{\protect \APACyear {2001}}%
}]{%
jaeger_echo_2001}
\APACinsertmetastar {%
jaeger_echo_2001}%
\begin{APACrefauthors}%
Jaeger, H.%
\end{APACrefauthors}%
\unskip\
\newblock
\APACrefYearMonthDay{2001}{}{}.
\newblock
{\BBOQ}\APACrefatitle {The "echo state” approach to analysing and training
  recurrent neural networks – with an {Erratum} note} {The "echo state”
  approach to analysing and training recurrent neural networks – with an
  {Erratum} note}.{\BBCQ}
\newblock
\APACjournalVolNumPages{Bonn, Germany: German National Research Center for
  Information Technology GMD Technical Report}{148}{34}{13}.
\PrintBackRefs{\CurrentBib}

\bibitem [\protect \citeauthoryear {%
Jones%
, Schonlau%
\BCBL {}\ \BBA {} Welch%
}{%
Jones%
\ \protect \BOthers {.}}{%
{\protect \APACyear {1998}}%
}]{%
jones_efficient_1998}
\APACinsertmetastar {%
jones_efficient_1998}%
\begin{APACrefauthors}%
Jones, D\BPBI R.%
, Schonlau, M.%
\BCBL {}\ \BBA {} Welch, W\BPBI J.%
\end{APACrefauthors}%
\unskip\
\newblock
\APACrefYearMonthDay{1998}{{\APACmonth{12}}}{}.
\newblock
{\BBOQ}\APACrefatitle {Efficient {Global} {Optimization} of {Expensive}
  {Black}-{Box} {Functions}} {Efficient {Global} {Optimization} of {Expensive}
  {Black}-{Box} {Functions}}.{\BBCQ}
\newblock
\APACjournalVolNumPages{Journal of Global Optimization}{13}{4}{455--492}.
\newblock
\begin{APACrefURL} [{2023-02-11}]\url{https://doi.org/10.1023/A:1008306431147}
  \end{APACrefURL}
\newblock
\begin{APACrefDOI} \doi{10.1023/A:1008306431147} \end{APACrefDOI}
\PrintBackRefs{\CurrentBib}

\bibitem [\protect \citeauthoryear {%
Jordanou%
, Antonelo%
, Camponogara%
\BCBL {}\ \BBA {} Gildin%
}{%
Jordanou%
\ \protect \BOthers {.}}{%
{\protect \APACyear {2022}}%
}]{%
jordanou_investigation_2022}
\APACinsertmetastar {%
jordanou_investigation_2022}%
\begin{APACrefauthors}%
Jordanou, J\BPBI P.%
, Antonelo, E\BPBI A.%
, Camponogara, E.%
\BCBL {}\ \BBA {} Gildin, E.%
\end{APACrefauthors}%
\unskip\
\newblock
\APACrefYearMonthDay{2022}{{\APACmonth{12}}}{}.
\newblock
\APACrefbtitle {Investigation of {Proper} {Orthogonal} {Decomposition} for
  {Echo} {State} {Networks}.} {Investigation of {Proper} {Orthogonal}
  {Decomposition} for {Echo} {State} {Networks}.}
\newblock
\APACaddressPublisher{}{arXiv}.
\newblock
\begin{APACrefURL} [{2022-12-20}]\url{http://arxiv.org/abs/2211.17179}
  \end{APACrefURL}
\newblock
\APACrefnote{arXiv:2211.17179 [cs, eess]}
\PrintBackRefs{\CurrentBib}

\bibitem [\protect \citeauthoryear {%
Kalnay%
, Hunt%
, Ott%
\BCBL {}\ \BBA {} Szunyogh%
}{%
Kalnay%
\ \protect \BOthers {.}}{%
{\protect \APACyear {2006}}%
}]{%
kalnay_ensemble_2006}
\APACinsertmetastar {%
kalnay_ensemble_2006}%
\begin{APACrefauthors}%
Kalnay, E.%
, Hunt, B.%
, Ott, E.%
\BCBL {}\ \BBA {} Szunyogh, I.%
\end{APACrefauthors}%
\unskip\
\newblock
\APACrefYearMonthDay{2006}{}{}.
\newblock
{\BBOQ}\APACrefatitle {Ensemble forecasting and data assimilation: two problems
  with the same solution} {Ensemble forecasting and data assimilation: two
  problems with the same solution}.{\BBCQ}
\newblock
\APACjournalVolNumPages{Predictability of weather and climate}{157}{}{180}.
\PrintBackRefs{\CurrentBib}

\bibitem [\protect \citeauthoryear {%
Keisler%
}{%
Keisler%
}{%
{\protect \APACyear {2022}}%
}]{%
keisler_forecasting_2022}
\APACinsertmetastar {%
keisler_forecasting_2022}%
\begin{APACrefauthors}%
Keisler, R.%
\end{APACrefauthors}%
\unskip\
\newblock
\APACrefYearMonthDay{2022}{{\APACmonth{02}}}{}.
\newblock
{\BBOQ}\APACrefatitle {Forecasting {Global} {Weather} with {Graph} {Neural}
  {Networks}} {Forecasting {Global} {Weather} with {Graph} {Neural}
  {Networks}}.{\BBCQ}
\newblock
\APACjournalVolNumPages{arXiv:2202.07575 [physics]}{}{}{}.
\newblock
\begin{APACrefURL} [{2022-02-17}]\url{http://arxiv.org/abs/2202.07575}
  \end{APACrefURL}
\newblock
\APACrefnote{arXiv: 2202.07575}
\PrintBackRefs{\CurrentBib}

\bibitem [\protect \citeauthoryear {%
Lam%
\ \protect \BOthers {.}}{%
Lam%
\ \protect \BOthers {.}}{%
{\protect \APACyear {2022}}%
}]{%
lam_graphcast_2022}
\APACinsertmetastar {%
lam_graphcast_2022}%
\begin{APACrefauthors}%
Lam, R.%
, Sanchez-Gonzalez, A.%
, Willson, M.%
, Wirnsberger, P.%
, Fortunato, M.%
, Pritzel, A.%
\BDBL {}Battaglia, P.%
\end{APACrefauthors}%
\unskip\
\newblock
\APACrefYearMonthDay{2022}{{\APACmonth{12}}}{}.
\newblock
\APACrefbtitle {{GraphCast}: {Learning} skillful medium-range global weather
  forecasting.} {{GraphCast}: {Learning} skillful medium-range global weather
  forecasting.}
\newblock
\APACaddressPublisher{}{arXiv}.
\newblock
\begin{APACrefURL} [{2023-01-03}]\url{http://arxiv.org/abs/2212.12794}
  \end{APACrefURL}
\newblock
\APACrefnote{arXiv:2212.12794 [physics]}
\newblock
\begin{APACrefDOI} \doi{10.48550/arXiv.2212.12794} \end{APACrefDOI}
\PrintBackRefs{\CurrentBib}

\bibitem [\protect \citeauthoryear {%
Li%
, Bouhlel%
\BCBL {}\ \BBA {} Martins%
}{%
Li%
\ \protect \BOthers {.}}{%
{\protect \APACyear {2019}}%
}]{%
li_data-based_2019}
\APACinsertmetastar {%
li_data-based_2019}%
\begin{APACrefauthors}%
Li, J.%
, Bouhlel, M\BPBI A.%
\BCBL {}\ \BBA {} Martins, J\BPBI R\BPBI R\BPBI A.%
\end{APACrefauthors}%
\unskip\
\newblock
\APACrefYearMonthDay{2019}{{\APACmonth{02}}}{}.
\newblock
{\BBOQ}\APACrefatitle {Data-{Based} {Approach} for {Fast} {Airfoil} {Analysis}
  and {Optimization}} {Data-{Based} {Approach} for {Fast} {Airfoil} {Analysis}
  and {Optimization}}.{\BBCQ}
\newblock
\APACjournalVolNumPages{AIAA Journal}{57}{2}{581--596}.
\newblock
\begin{APACrefURL}
  [{2023-04-03}]\url{https://arc.aiaa.org/doi/10.2514/1.J057129}
  \end{APACrefURL}
\newblock
\APACrefnote{Publisher: American Institute of Aeronautics and Astronautics}
\newblock
\begin{APACrefDOI} \doi{10.2514/1.J057129} \end{APACrefDOI}
\PrintBackRefs{\CurrentBib}

\bibitem [\protect \citeauthoryear {%
Lorenz%
}{%
Lorenz%
}{%
{\protect \APACyear {1996}}%
}]{%
lorenz_predictability_1996}
\APACinsertmetastar {%
lorenz_predictability_1996}%
\begin{APACrefauthors}%
Lorenz, E.%
\end{APACrefauthors}%
\unskip\
\newblock
\APACrefYearMonthDay{1996}{}{}.
\newblock
{\BBOQ}\APACrefatitle {Predictability - a problem partly solved}
  {Predictability - a problem partly solved}.{\BBCQ}
\newblock
\BIn{} \APACrefbtitle {Proceedings of a {Seminar} {Held} at {ECMWF} on
  {Predictability}.} {Proceedings of a {Seminar} {Held} at {ECMWF} on
  {Predictability}.}
\PrintBackRefs{\CurrentBib}

\bibitem [\protect \citeauthoryear {%
Lu%
, Hunt%
\BCBL {}\ \BBA {} Ott%
}{%
Lu%
\ \protect \BOthers {.}}{%
{\protect \APACyear {2018}}%
}]{%
lu_attractor_2018}
\APACinsertmetastar {%
lu_attractor_2018}%
\begin{APACrefauthors}%
Lu, Z.%
, Hunt, B\BPBI R.%
\BCBL {}\ \BBA {} Ott, E.%
\end{APACrefauthors}%
\unskip\
\newblock
\APACrefYearMonthDay{2018}{{\APACmonth{06}}}{}.
\newblock
{\BBOQ}\APACrefatitle {Attractor reconstruction by machine learning} {Attractor
  reconstruction by machine learning}.{\BBCQ}
\newblock
\APACjournalVolNumPages{Chaos: An Interdisciplinary Journal of Nonlinear
  Science}{28}{6}{061104}.
\newblock
\begin{APACrefURL}
  [{2020-12-29}]\url{http://aip.scitation.org/doi/10.1063/1.5039508}
  \end{APACrefURL}
\newblock
\APACrefnote{Publisher: American Institute of Physics}
\newblock
\begin{APACrefDOI} \doi{10.1063/1.5039508} \end{APACrefDOI}
\PrintBackRefs{\CurrentBib}

\bibitem [\protect \citeauthoryear {%
Lukoševičius%
}{%
Lukoševičius%
}{%
{\protect \APACyear {2012}}%
}]{%
lukosevicius_practical_2012}
\APACinsertmetastar {%
lukosevicius_practical_2012}%
\begin{APACrefauthors}%
Lukoševičius, M.%
\end{APACrefauthors}%
\unskip\
\newblock
\APACrefYearMonthDay{2012}{}{}.
\newblock
{\BBOQ}\APACrefatitle {A {Practical} {Guide} to {Applying} {Echo} {State}
  {Networks}} {A {Practical} {Guide} to {Applying} {Echo} {State}
  {Networks}}.{\BBCQ}
\newblock
\BIn{} G.~Montavon, G\BPBI B.~Orr\BCBL {}\ \BBA {} K\BHBI R.~Müller\ (\BEDS),
  \APACrefbtitle {Neural {Networks}: {Tricks} of the {Trade}: {Second}
  {Edition}} {Neural {Networks}: {Tricks} of the {Trade}: {Second} {Edition}}\
  (\BPGS\ 659--686).
\newblock
\APACaddressPublisher{Berlin, Heidelberg}{Springer}.
\newblock
\begin{APACrefURL}
  [{2021-01-04}]\url{https://doi.org/10.1007/978-3-642-35289-8_36}
  \end{APACrefURL}
\newblock
\begin{APACrefDOI} \doi{10.1007/978-3-642-35289-8_36} \end{APACrefDOI}
\PrintBackRefs{\CurrentBib}

\bibitem [\protect \citeauthoryear {%
Ma%
, Shen%
\BCBL {}\ \BBA {} Cottrell%
}{%
Ma%
\ \protect \BOthers {.}}{%
{\protect \APACyear {2020}}%
}]{%
ma_deepr-esn_2020}
\APACinsertmetastar {%
ma_deepr-esn_2020}%
\begin{APACrefauthors}%
Ma, Q.%
, Shen, L.%
\BCBL {}\ \BBA {} Cottrell, G\BPBI W.%
\end{APACrefauthors}%
\unskip\
\newblock
\APACrefYearMonthDay{2020}{{\APACmonth{02}}}{}.
\newblock
{\BBOQ}\APACrefatitle {{DeePr}-{ESN}: {A} deep projection-encoding echo-state
  network} {{DeePr}-{ESN}: {A} deep projection-encoding echo-state
  network}.{\BBCQ}
\newblock
\APACjournalVolNumPages{Information Sciences}{511}{}{152--171}.
\newblock
\begin{APACrefURL}
  [{2022-10-03}]\url{https://www.sciencedirect.com/science/article/pii/S0020025519309053}
  \end{APACrefURL}
\newblock
\begin{APACrefDOI} \doi{10.1016/j.ins.2019.09.049} \end{APACrefDOI}
\PrintBackRefs{\CurrentBib}

\bibitem [\protect \citeauthoryear {%
Maass%
, Natschläger%
\BCBL {}\ \BBA {} Markram%
}{%
Maass%
\ \protect \BOthers {.}}{%
{\protect \APACyear {2002}}%
}]{%
maass_real-time_2002}
\APACinsertmetastar {%
maass_real-time_2002}%
\begin{APACrefauthors}%
Maass, W.%
, Natschläger, T.%
\BCBL {}\ \BBA {} Markram, H.%
\end{APACrefauthors}%
\unskip\
\newblock
\APACrefYearMonthDay{2002}{{\APACmonth{11}}}{}.
\newblock
{\BBOQ}\APACrefatitle {Real-{Time} {Computing} {Without} {Stable} {States}: {A}
  {New} {Framework} for {Neural} {Computation} {Based} on {Perturbations}}
  {Real-{Time} {Computing} {Without} {Stable} {States}: {A} {New} {Framework}
  for {Neural} {Computation} {Based} on {Perturbations}}.{\BBCQ}
\newblock
\APACjournalVolNumPages{Neural Computation}{14}{11}{2531--2560}.
\newblock
\begin{APACrefURL}
  [{2022-12-05}]\url{https://direct.mit.edu/neco/article/14/11/2531-2560/6650}
  \end{APACrefURL}
\newblock
\begin{APACrefDOI} \doi{10.1162/089976602760407955} \end{APACrefDOI}
\PrintBackRefs{\CurrentBib}

\bibitem [\protect \citeauthoryear {%
Malik%
, Hussain%
\BCBL {}\ \BBA {} Wu%
}{%
Malik%
\ \protect \BOthers {.}}{%
{\protect \APACyear {2017}}%
}]{%
malik_multilayered_2017}
\APACinsertmetastar {%
malik_multilayered_2017}%
\begin{APACrefauthors}%
Malik, Z\BPBI K.%
, Hussain, A.%
\BCBL {}\ \BBA {} Wu, Q\BPBI J.%
\end{APACrefauthors}%
\unskip\
\newblock
\APACrefYearMonthDay{2017}{{\APACmonth{04}}}{}.
\newblock
{\BBOQ}\APACrefatitle {Multilayered {Echo} {State} {Machine}: {A} {Novel}
  {Architecture} and {Algorithm}} {Multilayered {Echo} {State} {Machine}: {A}
  {Novel} {Architecture} and {Algorithm}}.{\BBCQ}
\newblock
\APACjournalVolNumPages{IEEE Transactions on Cybernetics}{47}{4}{946--959}.
\newblock
\begin{APACrefURL}
  [{2023-03-30}]\url{https://ieeexplore.ieee.org/document/7494974/}
  \end{APACrefURL}
\newblock
\begin{APACrefDOI} \doi{10.1109/TCYB.2016.2533545} \end{APACrefDOI}
\PrintBackRefs{\CurrentBib}

\bibitem [\protect \citeauthoryear {%
Moon%
, Wu%
\BCBL {}\ \BBA {} Lu%
}{%
Moon%
\ \protect \BOthers {.}}{%
{\protect \APACyear {2021}}%
}]{%
moon_hierarchical_2021}
\APACinsertmetastar {%
moon_hierarchical_2021}%
\begin{APACrefauthors}%
Moon, J.%
, Wu, Y.%
\BCBL {}\ \BBA {} Lu, W\BPBI D.%
\end{APACrefauthors}%
\unskip\
\newblock
\APACrefYearMonthDay{2021}{{\APACmonth{08}}}{}.
\newblock
{\BBOQ}\APACrefatitle {Hierarchical architectures in reservoir computing
  systems} {Hierarchical architectures in reservoir computing systems}.{\BBCQ}
\newblock
\APACjournalVolNumPages{Neuromorphic Computing and Engineering}{1}{1}{014006}.
\newblock
\begin{APACrefURL} [{2022-10-03}]\url{https://doi.org/10.1088/2634-4386/ac1b75}
  \end{APACrefURL}
\newblock
\APACrefnote{Publisher: IOP Publishing}
\newblock
\begin{APACrefDOI} \doi{10.1088/2634-4386/ac1b75} \end{APACrefDOI}
\PrintBackRefs{\CurrentBib}

\bibitem [\protect \citeauthoryear {%
Moore%
, Fiechter%
\BCBL {}\ \BBA {} Edwards%
}{%
Moore%
\ \protect \BOthers {.}}{%
{\protect \APACyear {2022}}%
}]{%
moore_linear_2022}
\APACinsertmetastar {%
moore_linear_2022}%
\begin{APACrefauthors}%
Moore, A\BPBI M.%
, Fiechter, J.%
\BCBL {}\ \BBA {} Edwards, C\BPBI A.%
\end{APACrefauthors}%
\unskip\
\newblock
\APACrefYearMonthDay{2022}{{\APACmonth{06}}}{}.
\newblock
{\BBOQ}\APACrefatitle {A linear stochastic emulator of the {California}
  {Current} system using balanced truncation} {A linear stochastic emulator of
  the {California} {Current} system using balanced truncation}.{\BBCQ}
\newblock
\APACjournalVolNumPages{Ocean Modelling}{174}{}{102023}.
\newblock
\begin{APACrefURL}
  [{2022-06-21}]\url{https://linkinghub.elsevier.com/retrieve/pii/S1463500322000610}
  \end{APACrefURL}
\newblock
\begin{APACrefDOI} \doi{10.1016/j.ocemod.2022.102023} \end{APACrefDOI}
\PrintBackRefs{\CurrentBib}

\bibitem [\protect \citeauthoryear {%
Nadiga%
}{%
Nadiga%
}{%
{\protect \APACyear {2021}}%
}]{%
nadiga_reservoir_2021}
\APACinsertmetastar {%
nadiga_reservoir_2021}%
\begin{APACrefauthors}%
Nadiga, B\BPBI T.%
\end{APACrefauthors}%
\unskip\
\newblock
\APACrefYearMonthDay{2021}{}{}.
\newblock
{\BBOQ}\APACrefatitle {Reservoir {Computing} as a {Tool} for {Climate}
  {Predictability} {Studies}} {Reservoir {Computing} as a {Tool} for {Climate}
  {Predictability} {Studies}}.{\BBCQ}
\newblock
\APACjournalVolNumPages{Journal of Advances in Modeling Earth
  Systems}{13}{4}{e2020MS002290}.
\newblock
\begin{APACrefURL}
  [{2021-11-17}]\url{https://onlinelibrary.wiley.com/doi/abs/10.1029/2020MS002290}
  \end{APACrefURL}
\newblock
\APACrefnote{\_eprint:
  https://onlinelibrary.wiley.com/doi/pdf/10.1029/2020MS002290}
\newblock
\begin{APACrefDOI} \doi{10.1029/2020MS002290} \end{APACrefDOI}
\PrintBackRefs{\CurrentBib}

\bibitem [\protect \citeauthoryear {%
Najm%
}{%
Najm%
}{%
{\protect \APACyear {2009}}%
}]{%
najm_uncertainty_2009}
\APACinsertmetastar {%
najm_uncertainty_2009}%
\begin{APACrefauthors}%
Najm, H\BPBI N.%
\end{APACrefauthors}%
\unskip\
\newblock
\APACrefYearMonthDay{2009}{}{}.
\newblock
{\BBOQ}\APACrefatitle {Uncertainty {Quantification} and {Polynomial} {Chaos}
  {Techniques} in {Computational} {Fluid} {Dynamics}} {Uncertainty
  {Quantification} and {Polynomial} {Chaos} {Techniques} in {Computational}
  {Fluid} {Dynamics}}.{\BBCQ}
\newblock
\APACjournalVolNumPages{Annual Review of Fluid Mechanics}{41}{1}{35--52}.
\newblock
\begin{APACrefURL}
  [{2022-10-11}]\url{https://doi.org/10.1146/annurev.fluid.010908.165248}
  \end{APACrefURL}
\newblock
\APACrefnote{\_eprint: https://doi.org/10.1146/annurev.fluid.010908.165248}
\newblock
\begin{APACrefDOI} \doi{10.1146/annurev.fluid.010908.165248} \end{APACrefDOI}
\PrintBackRefs{\CurrentBib}

\bibitem [\protect \citeauthoryear {%
Okuta%
, Unno%
, Nishino%
, Hido%
\BCBL {}\ \BBA {} Loomis%
}{%
Okuta%
\ \protect \BOthers {.}}{%
{\protect \APACyear {2017}}%
}]{%
cupy_learningsys2017}
\APACinsertmetastar {%
cupy_learningsys2017}%
\begin{APACrefauthors}%
Okuta, R.%
, Unno, Y.%
, Nishino, D.%
, Hido, S.%
\BCBL {}\ \BBA {} Loomis, C.%
\end{APACrefauthors}%
\unskip\
\newblock
\APACrefYearMonthDay{2017}{}{}.
\newblock
{\BBOQ}\APACrefatitle {CuPy: A NumPy-Compatible Library for NVIDIA GPU
  Calculations} {Cupy: A numpy-compatible library for nvidia gpu
  calculations}.{\BBCQ}
\newblock
\BIn{} \APACrefbtitle {Proceedings of Workshop on Machine Learning Systems
  (LearningSys) in The Thirty-first Annual Conference on Neural Information
  Processing Systems (NIPS).} {Proceedings of workshop on machine learning
  systems (learningsys) in the thirty-first annual conference on neural
  information processing systems (nips).}
\newblock
\begin{APACrefURL}
  \url{http://learningsys.org/nips17/assets/papers/paper_16.pdf}
  \end{APACrefURL}
\PrintBackRefs{\CurrentBib}

\bibitem [\protect \citeauthoryear {%
Orlanski%
}{%
Orlanski%
}{%
{\protect \APACyear {1975}}%
}]{%
orlanski_rational_1975}
\APACinsertmetastar {%
orlanski_rational_1975}%
\begin{APACrefauthors}%
Orlanski, I.%
\end{APACrefauthors}%
\unskip\
\newblock
\APACrefYearMonthDay{1975}{}{}.
\newblock
{\BBOQ}\APACrefatitle {A {Rational} {Subdivision} of {Scales} for {Atmospheric}
  {Processes}} {A {Rational} {Subdivision} of {Scales} for {Atmospheric}
  {Processes}}.{\BBCQ}
\newblock
\APACjournalVolNumPages{Bulletin of the American Meteorological
  Society}{56}{5}{527--530}.
\newblock
\begin{APACrefURL} [{2023-04-17}]\url{https://www.jstor.org/stable/26216020}
  \end{APACrefURL}
\newblock
\APACrefnote{Publisher: American Meteorological Society}
\PrintBackRefs{\CurrentBib}

\bibitem [\protect \citeauthoryear {%
Pathak%
, Hunt%
, Girvan%
, Lu%
\BCBL {}\ \BBA {} Ott%
}{%
Pathak%
\ \protect \BOthers {.}}{%
{\protect \APACyear {2018}}%
}]{%
pathak_model-free_2018}
\APACinsertmetastar {%
pathak_model-free_2018}%
\begin{APACrefauthors}%
Pathak, J.%
, Hunt, B.%
, Girvan, M.%
, Lu, Z.%
\BCBL {}\ \BBA {} Ott, E.%
\end{APACrefauthors}%
\unskip\
\newblock
\APACrefYearMonthDay{2018}{{\APACmonth{01}}}{}.
\newblock
{\BBOQ}\APACrefatitle {Model-{Free} {Prediction} of {Large} {Spatiotemporally}
  {Chaotic} {Systems} from {Data}: {A} {Reservoir} {Computing} {Approach}}
  {Model-{Free} {Prediction} of {Large} {Spatiotemporally} {Chaotic} {Systems}
  from {Data}: {A} {Reservoir} {Computing} {Approach}}.{\BBCQ}
\newblock
\APACjournalVolNumPages{Physical Review Letters}{120}{2}{024102}.
\newblock
\begin{APACrefURL}
  [{2020-12-29}]\url{https://link.aps.org/doi/10.1103/PhysRevLett.120.024102}
  \end{APACrefURL}
\newblock
\APACrefnote{Publisher: American Physical Society}
\newblock
\begin{APACrefDOI} \doi{10.1103/PhysRevLett.120.024102} \end{APACrefDOI}
\PrintBackRefs{\CurrentBib}

\bibitem [\protect \citeauthoryear {%
Pathak%
, Lu%
, Hunt%
, Girvan%
\BCBL {}\ \BBA {} Ott%
}{%
Pathak%
\ \protect \BOthers {.}}{%
{\protect \APACyear {2017}}%
}]{%
pathak_using_2017}
\APACinsertmetastar {%
pathak_using_2017}%
\begin{APACrefauthors}%
Pathak, J.%
, Lu, Z.%
, Hunt, B\BPBI R.%
, Girvan, M.%
\BCBL {}\ \BBA {} Ott, E.%
\end{APACrefauthors}%
\unskip\
\newblock
\APACrefYearMonthDay{2017}{{\APACmonth{12}}}{}.
\newblock
{\BBOQ}\APACrefatitle {Using machine learning to replicate chaotic attractors
  and calculate {Lyapunov} exponents from data} {Using machine learning to
  replicate chaotic attractors and calculate {Lyapunov} exponents from
  data}.{\BBCQ}
\newblock
\APACjournalVolNumPages{Chaos: An Interdisciplinary Journal of Nonlinear
  Science}{27}{12}{121102}.
\newblock
\begin{APACrefURL}
  [{2022-10-21}]\url{https://aip.scitation.org/doi/10.1063/1.5010300}
  \end{APACrefURL}
\newblock
\APACrefnote{Publisher: American Institute of Physics}
\newblock
\begin{APACrefDOI} \doi{10.1063/1.5010300} \end{APACrefDOI}
\PrintBackRefs{\CurrentBib}

\bibitem [\protect \citeauthoryear {%
Pathak%
\ \protect \BOthers {.}}{%
Pathak%
\ \protect \BOthers {.}}{%
{\protect \APACyear {2022}}%
}]{%
pathak_fourcastnet_2022}
\APACinsertmetastar {%
pathak_fourcastnet_2022}%
\begin{APACrefauthors}%
Pathak, J.%
, Subramanian, S.%
, Harrington, P.%
, Raja, S.%
, Chattopadhyay, A.%
, Mardani, M.%
\BDBL {}Anandkumar, A.%
\end{APACrefauthors}%
\unskip\
\newblock
\APACrefYearMonthDay{2022}{{\APACmonth{02}}}{}.
\newblock
{\BBOQ}\APACrefatitle {{FourCastNet}: {A} {Global} {Data}-driven
  {High}-resolution {Weather} {Model} using {Adaptive} {Fourier} {Neural}
  {Operators}} {{FourCastNet}: {A} {Global} {Data}-driven {High}-resolution
  {Weather} {Model} using {Adaptive} {Fourier} {Neural} {Operators}}.{\BBCQ}
\newblock
\APACjournalVolNumPages{arXiv:2202.11214 [physics]}{}{}{}.
\newblock
\begin{APACrefURL} [{2022-03-24}]\url{http://arxiv.org/abs/2202.11214}
  \end{APACrefURL}
\newblock
\APACrefnote{arXiv: 2202.11214}
\PrintBackRefs{\CurrentBib}

\bibitem [\protect \citeauthoryear {%
Penland%
}{%
Penland%
}{%
{\protect \APACyear {1989}}%
}]{%
penland_random_1989}
\APACinsertmetastar {%
penland_random_1989}%
\begin{APACrefauthors}%
Penland, C.%
\end{APACrefauthors}%
\unskip\
\newblock
\APACrefYearMonthDay{1989}{{\APACmonth{10}}}{}.
\newblock
{\BBOQ}\APACrefatitle {Random {Forcing} and {Forecasting} {Using} {Principal}
  {Oscillation} {Pattern} {Analysis}} {Random {Forcing} and {Forecasting}
  {Using} {Principal} {Oscillation} {Pattern} {Analysis}}.{\BBCQ}
\newblock
\APACjournalVolNumPages{Monthly Weather Review}{117}{10}{2165--2185}.
\newblock
\begin{APACrefURL}
  [{2022-10-11}]\url{https://journals.ametsoc.org/view/journals/mwre/117/10/1520-0493_1989_117_2165_rfafup_2_0_co_2.xml}
  \end{APACrefURL}
\newblock
\APACrefnote{Publisher: American Meteorological Society Section: Monthly
  Weather Review}
\newblock
\begin{APACrefDOI} \doi{10.1175/1520-0493(1989)117<2165:RFAFUP>2.0.CO;2}
  \end{APACrefDOI}
\PrintBackRefs{\CurrentBib}

\bibitem [\protect \citeauthoryear {%
Penny%
\ \protect \BOthers {.}}{%
Penny%
\ \protect \BOthers {.}}{%
{\protect \APACyear {2017}}%
}]{%
penny_coupled_2017}
\APACinsertmetastar {%
penny_coupled_2017}%
\begin{APACrefauthors}%
Penny, S\BPBI G.%
, Akella, S.%
, Alves, O.%
, Craig, B.%
, Buehner, M.%
, Chevallier, M.%
\BDBL {}Wu, X.%
\end{APACrefauthors}%
\unskip\
\newblock
\APACrefYearMonthDay{2017}{}{}.
\newblock
\APACrefbtitle {Coupled {Data} {Assimilation} for {Integrated} {Earth} {System}
  {Analysis} and {Prediction}: {Goals}, {Challenges} and {Recommendations}}
  {Coupled {Data} {Assimilation} for {Integrated} {Earth} {System} {Analysis}
  and {Prediction}: {Goals}, {Challenges} and {Recommendations}}\
  \APACbVolEdTR{}{\BTR{}}.
\newblock
\APACaddressInstitution{}{Geneva: World Meteorological Organization.}
\PrintBackRefs{\CurrentBib}

\bibitem [\protect \citeauthoryear {%
Penny%
\ \protect \BOthers {.}}{%
Penny%
\ \protect \BOthers {.}}{%
{\protect \APACyear {2022}}%
}]{%
penny_integrating_2022}
\APACinsertmetastar {%
penny_integrating_2022}%
\begin{APACrefauthors}%
Penny, S\BPBI G.%
, Smith, T\BPBI A.%
, Chen, T\BHBI C.%
, Platt, J\BPBI A.%
, Lin, H\BHBI Y.%
, Goodliff, M.%
\BCBL {}\ \BBA {} Abarbanel, H\BPBI D\BPBI I.%
\end{APACrefauthors}%
\unskip\
\newblock
\APACrefYearMonthDay{2022}{}{}.
\newblock
{\BBOQ}\APACrefatitle {Integrating {Recurrent} {Neural} {Networks} {With}
  {Data} {Assimilation} for {Scalable} {Data}-{Driven} {State} {Estimation}}
  {Integrating {Recurrent} {Neural} {Networks} {With} {Data} {Assimilation} for
  {Scalable} {Data}-{Driven} {State} {Estimation}}.{\BBCQ}
\newblock
\APACjournalVolNumPages{Journal of Advances in Modeling Earth
  Systems}{14}{3}{e2021MS002843}.
\newblock
\begin{APACrefURL}
  [{2022-03-04}]\url{https://onlinelibrary.wiley.com/doi/abs/10.1029/2021MS002843}
  \end{APACrefURL}
\newblock
\APACrefnote{\_eprint:
  https://onlinelibrary.wiley.com/doi/pdf/10.1029/2021MS002843}
\newblock
\begin{APACrefDOI} \doi{10.1029/2021MS002843} \end{APACrefDOI}
\PrintBackRefs{\CurrentBib}

\bibitem [\protect \citeauthoryear {%
Platt%
, Penny%
, Smith%
, Chen%
\BCBL {}\ \BBA {} Abarbanel%
}{%
Platt%
\ \protect \BOthers {.}}{%
{\protect \APACyear {2022}}%
}]{%
platt_systematic_2022}
\APACinsertmetastar {%
platt_systematic_2022}%
\begin{APACrefauthors}%
Platt, J\BPBI A.%
, Penny, S\BPBI G.%
, Smith, T\BPBI A.%
, Chen, T\BHBI C.%
\BCBL {}\ \BBA {} Abarbanel, H\BPBI D\BPBI I.%
\end{APACrefauthors}%
\unskip\
\newblock
\APACrefYearMonthDay{2022}{{\APACmonth{09}}}{}.
\newblock
{\BBOQ}\APACrefatitle {A systematic exploration of reservoir computing for
  forecasting complex spatiotemporal dynamics} {A systematic exploration of
  reservoir computing for forecasting complex spatiotemporal dynamics}.{\BBCQ}
\newblock
\APACjournalVolNumPages{Neural Networks}{153}{}{530--552}.
\newblock
\begin{APACrefURL}
  [{2022-07-13}]\url{https://www.sciencedirect.com/science/article/pii/S0893608022002404}
  \end{APACrefURL}
\newblock
\begin{APACrefDOI} \doi{10.1016/j.neunet.2022.06.025} \end{APACrefDOI}
\PrintBackRefs{\CurrentBib}

\bibitem [\protect \citeauthoryear {%
Platt%
, Penny%
, Smith%
, Chen%
\BCBL {}\ \BBA {} Abarbanel%
}{%
Platt%
\ \protect \BOthers {.}}{%
{\protect \APACyear {2023}}%
}]{%
platt_constraining_2023}
\APACinsertmetastar {%
platt_constraining_2023}%
\begin{APACrefauthors}%
Platt, J\BPBI A.%
, Penny, S\BPBI G.%
, Smith, T\BPBI A.%
, Chen, T\BHBI C.%
\BCBL {}\ \BBA {} Abarbanel, H\BPBI D\BPBI I.%
\end{APACrefauthors}%
\unskip\
\newblock
\APACrefYearMonthDay{2023}{{\APACmonth{04}}}{}.
\newblock
\APACrefbtitle {Constraining {Chaos}: {Enforcing} dynamical invariants in the
  training of recurrent neural networks.} {Constraining {Chaos}: {Enforcing}
  dynamical invariants in the training of recurrent neural networks.}
\newblock
\APACaddressPublisher{}{arXiv}.
\newblock
\begin{APACrefURL} [{2023-04-26}]\url{http://arxiv.org/abs/2304.12865}
  \end{APACrefURL}
\newblock
\APACrefnote{arXiv:2304.12865 [physics]}
\newblock
\begin{APACrefDOI} \doi{10.48550/arXiv.2304.12865} \end{APACrefDOI}
\PrintBackRefs{\CurrentBib}

\bibitem [\protect \citeauthoryear {%
Rasp%
\ \BBA {} Thuerey%
}{%
Rasp%
\ \BBA {} Thuerey%
}{%
{\protect \APACyear {2021}}%
}]{%
rasp_data-driven_2021}
\APACinsertmetastar {%
rasp_data-driven_2021}%
\begin{APACrefauthors}%
Rasp, S.%
\BCBT {}\ \BBA {} Thuerey, N.%
\end{APACrefauthors}%
\unskip\
\newblock
\APACrefYearMonthDay{2021}{}{}.
\newblock
{\BBOQ}\APACrefatitle {Data-{Driven} {Medium}-{Range} {Weather} {Prediction}
  {With} a {Resnet} {Pretrained} on {Climate} {Simulations}: {A} {New} {Model}
  for {WeatherBench}} {Data-{Driven} {Medium}-{Range} {Weather} {Prediction}
  {With} a {Resnet} {Pretrained} on {Climate} {Simulations}: {A} {New} {Model}
  for {WeatherBench}}.{\BBCQ}
\newblock
\APACjournalVolNumPages{Journal of Advances in Modeling Earth
  Systems}{13}{2}{e2020MS002405}.
\newblock
\begin{APACrefURL}
  [{2022-10-03}]\url{https://onlinelibrary.wiley.com/doi/abs/10.1029/2020MS002405}
  \end{APACrefURL}
\newblock
\APACrefnote{\_eprint:
  https://onlinelibrary.wiley.com/doi/pdf/10.1029/2020MS002405}
\newblock
\begin{APACrefDOI} \doi{10.1029/2020MS002405} \end{APACrefDOI}
\PrintBackRefs{\CurrentBib}

\bibitem [\protect \citeauthoryear {%
Rossa%
, Nurmi%
\BCBL {}\ \BBA {} Ebert%
}{%
Rossa%
\ \protect \BOthers {.}}{%
{\protect \APACyear {2008}}%
}]{%
rossa_overview_2008}
\APACinsertmetastar {%
rossa_overview_2008}%
\begin{APACrefauthors}%
Rossa, A.%
, Nurmi, P.%
\BCBL {}\ \BBA {} Ebert, E.%
\end{APACrefauthors}%
\unskip\
\newblock
\APACrefYearMonthDay{2008}{}{}.
\newblock
{\BBOQ}\APACrefatitle {Overview of methods for the verification of quantitative
  precipitation forecasts} {Overview of methods for the verification of
  quantitative precipitation forecasts}.{\BBCQ}
\newblock
\BIn{} S.~Michaelides\ (\BED), \APACrefbtitle {Precipitation: {Advances} in
  {Measurement}, {Estimation} and {Prediction}} {Precipitation: {Advances} in
  {Measurement}, {Estimation} and {Prediction}}\ (\BPGS\ 419--452).
\newblock
\APACaddressPublisher{Berlin, Heidelberg}{Springer}.
\newblock
\begin{APACrefURL}
  [{2023-03-31}]\url{https://doi.org/10.1007/978-3-540-77655-0_16}
  \end{APACrefURL}
\newblock
\begin{APACrefDOI} \doi{10.1007/978-3-540-77655-0_16} \end{APACrefDOI}
\PrintBackRefs{\CurrentBib}

\bibitem [\protect \citeauthoryear {%
Scher%
}{%
Scher%
}{%
{\protect \APACyear {2018}}%
}]{%
scher_toward_2018}
\APACinsertmetastar {%
scher_toward_2018}%
\begin{APACrefauthors}%
Scher, S.%
\end{APACrefauthors}%
\unskip\
\newblock
\APACrefYearMonthDay{2018}{}{}.
\newblock
{\BBOQ}\APACrefatitle {Toward {Data}-{Driven} {Weather} and {Climate}
  {Forecasting}: {Approximating} a {Simple} {General} {Circulation} {Model}
  {With} {Deep} {Learning}} {Toward {Data}-{Driven} {Weather} and {Climate}
  {Forecasting}: {Approximating} a {Simple} {General} {Circulation} {Model}
  {With} {Deep} {Learning}}.{\BBCQ}
\newblock
\APACjournalVolNumPages{Geophysical Research Letters}{45}{22}{12,616--12,622}.
\newblock
\begin{APACrefURL}
  [{2022-10-11}]\url{https://onlinelibrary.wiley.com/doi/abs/10.1029/2018GL080704}
  \end{APACrefURL}
\newblock
\APACrefnote{\_eprint:
  https://onlinelibrary.wiley.com/doi/pdf/10.1029/2018GL080704}
\newblock
\begin{APACrefDOI} \doi{10.1029/2018GL080704} \end{APACrefDOI}
\PrintBackRefs{\CurrentBib}

\bibitem [\protect \citeauthoryear {%
Scher%
\ \BBA {} Messori%
}{%
Scher%
\ \BBA {} Messori%
}{%
{\protect \APACyear {2019}}%
}]{%
scher_weather_2019}
\APACinsertmetastar {%
scher_weather_2019}%
\begin{APACrefauthors}%
Scher, S.%
\BCBT {}\ \BBA {} Messori, G.%
\end{APACrefauthors}%
\unskip\
\newblock
\APACrefYearMonthDay{2019}{{\APACmonth{07}}}{}.
\newblock
{\BBOQ}\APACrefatitle {Weather and climate forecasting with neural networks:
  using general circulation models ({GCMs}) with different complexity as a
  study ground} {Weather and climate forecasting with neural networks: using
  general circulation models ({GCMs}) with different complexity as a study
  ground}.{\BBCQ}
\newblock
\APACjournalVolNumPages{Geoscientific Model Development}{12}{7}{2797--2809}.
\newblock
\begin{APACrefURL}
  [{2022-10-03}]\url{https://gmd.copernicus.org/articles/12/2797/2019/}
  \end{APACrefURL}
\newblock
\APACrefnote{Publisher: Copernicus GmbH}
\newblock
\begin{APACrefDOI} \doi{10.5194/gmd-12-2797-2019} \end{APACrefDOI}
\PrintBackRefs{\CurrentBib}

\bibitem [\protect \citeauthoryear {%
Schultz%
\ \protect \BOthers {.}}{%
Schultz%
\ \protect \BOthers {.}}{%
{\protect \APACyear {2021}}%
}]{%
schultz_can_2021}
\APACinsertmetastar {%
schultz_can_2021}%
\begin{APACrefauthors}%
Schultz, M\BPBI G.%
, Betancourt, C.%
, Gong, B.%
, Kleinert, F.%
, Langguth, M.%
, Leufen, L\BPBI H.%
\BDBL {}Stadtler, S.%
\end{APACrefauthors}%
\unskip\
\newblock
\APACrefYearMonthDay{2021}{{\APACmonth{04}}}{}.
\newblock
{\BBOQ}\APACrefatitle {Can deep learning beat numerical weather prediction?}
  {Can deep learning beat numerical weather prediction?}{\BBCQ}
\newblock
\APACjournalVolNumPages{Philosophical Transactions of the Royal Society A:
  Mathematical, Physical and Engineering Sciences}{379}{2194}{20200097}.
\newblock
\begin{APACrefURL}
  [{2022-10-07}]\url{https://royalsocietypublishing.org/doi/10.1098/rsta.2020.0097}
  \end{APACrefURL}
\newblock
\APACrefnote{Publisher: Royal Society}
\newblock
\begin{APACrefDOI} \doi{10.1098/rsta.2020.0097} \end{APACrefDOI}
\PrintBackRefs{\CurrentBib}

\bibitem [\protect \citeauthoryear {%
Sitzmann%
, Martel%
, Bergman%
, Lindell%
\BCBL {}\ \BBA {} Wetzstein%
}{%
Sitzmann%
\ \protect \BOthers {.}}{%
{\protect \APACyear {2020}}%
}]{%
sitzmann_implicit_2020}
\APACinsertmetastar {%
sitzmann_implicit_2020}%
\begin{APACrefauthors}%
Sitzmann, V.%
, Martel, J\BPBI N\BPBI P.%
, Bergman, A\BPBI W.%
, Lindell, D\BPBI B.%
\BCBL {}\ \BBA {} Wetzstein, G.%
\end{APACrefauthors}%
\unskip\
\newblock
\APACrefYearMonthDay{2020}{{\APACmonth{06}}}{}.
\newblock
\APACrefbtitle {Implicit {Neural} {Representations} with {Periodic}
  {Activation} {Functions}.} {Implicit {Neural} {Representations} with
  {Periodic} {Activation} {Functions}.}
\newblock
\APACaddressPublisher{}{arXiv}.
\newblock
\begin{APACrefURL} [{2023-08-22}]\url{http://arxiv.org/abs/2006.09661}
  \end{APACrefURL}
\newblock
\APACrefnote{arXiv:2006.09661 [cs, eess]}
\PrintBackRefs{\CurrentBib}

\bibitem [\protect \citeauthoryear {%
Smith%
}{%
Smith%
}{%
{\protect \APACyear {2023}}%
}]{%
smith_rcgfd_2023}
\APACinsertmetastar {%
smith_rcgfd_2023}%
\begin{APACrefauthors}%
Smith, T.%
\end{APACrefauthors}%
\unskip\
\newblock
\APACrefYearMonthDay{2023}{{\APACmonth{09}}}{}.
\newblock
\APACrefbtitle {timothyas/rc-gfd: Revision 1.} {timothyas/rc-gfd: Revision 1.}
\newblock
\APACaddressPublisher{}{Zenodo}.
\newblock
\begin{APACrefURL} \url{https://doi.org/10.5281/zenodo.8368225}
  \end{APACrefURL}
\newblock
\begin{APACrefDOI} \doi{10.5281/zenodo.8368225} \end{APACrefDOI}
\PrintBackRefs{\CurrentBib}

\bibitem [\protect \citeauthoryear {%
Steil%
}{%
Steil%
}{%
{\protect \APACyear {2004}}%
}]{%
steil_backpropagation-decorrelation_2004}
\APACinsertmetastar {%
steil_backpropagation-decorrelation_2004}%
\begin{APACrefauthors}%
Steil, J.%
\end{APACrefauthors}%
\unskip\
\newblock
\APACrefYearMonthDay{2004}{{\APACmonth{07}}}{}.
\newblock
{\BBOQ}\APACrefatitle {Backpropagation-decorrelation: online recurrent learning
  with {O}({N}) complexity} {Backpropagation-decorrelation: online recurrent
  learning with {O}({N}) complexity}.{\BBCQ}
\newblock
\BIn{} \APACrefbtitle {2004 {IEEE} {International} {Joint} {Conference} on
  {Neural} {Networks} ({IEEE} {Cat}. {No}.{04CH37541})} {2004 {IEEE}
  {International} {Joint} {Conference} on {Neural} {Networks} ({IEEE} {Cat}.
  {No}.{04CH37541})}\ (\BVOL~2, \BPGS\ 843--848 vol.2).
\newblock
\APACrefnote{ISSN: 1098-7576}
\newblock
\begin{APACrefDOI} \doi{10.1109/IJCNN.2004.1380039} \end{APACrefDOI}
\PrintBackRefs{\CurrentBib}

\bibitem [\protect \citeauthoryear {%
Tikhonov%
}{%
Tikhonov%
}{%
{\protect \APACyear {1963}}%
}]{%
tikhonov_solution_1963}
\APACinsertmetastar {%
tikhonov_solution_1963}%
\begin{APACrefauthors}%
Tikhonov, A\BPBI N.%
\end{APACrefauthors}%
\unskip\
\newblock
\APACrefYearMonthDay{1963}{}{}.
\newblock
{\BBOQ}\APACrefatitle {Solution of Incorrectly Formulated Problems and the
  Regularization Method} {Solution of incorrectly formulated problems and the
  regularization method}.{\BBCQ}
\newblock
\APACjournalVolNumPages{Soviet Math. Dokl}{}{}{}.
\PrintBackRefs{\CurrentBib}

\bibitem [\protect \citeauthoryear {%
Tulloch%
\ \BBA {} Smith%
}{%
Tulloch%
\ \BBA {} Smith%
}{%
{\protect \APACyear {2009}}%
}]{%
tulloch_note_2009}
\APACinsertmetastar {%
tulloch_note_2009}%
\begin{APACrefauthors}%
Tulloch, R.%
\BCBT {}\ \BBA {} Smith, K\BPBI S.%
\end{APACrefauthors}%
\unskip\
\newblock
\APACrefYearMonthDay{2009}{{\APACmonth{04}}}{}.
\newblock
{\BBOQ}\APACrefatitle {A {Note} on the {Numerical} {Representation} of
  {Surface} {Dynamics} in {Quasigeostrophic} {Turbulence}: {Application} to the
  {Nonlinear} {Eady} {Model}} {A {Note} on the {Numerical} {Representation} of
  {Surface} {Dynamics} in {Quasigeostrophic} {Turbulence}: {Application} to the
  {Nonlinear} {Eady} {Model}}.{\BBCQ}
\newblock
\APACjournalVolNumPages{Journal of the Atmospheric
  Sciences}{66}{4}{1063--1068}.
\newblock
\begin{APACrefURL}
  [{2021-11-24}]\url{https://journals.ametsoc.org/view/journals/atsc/66/4/2008jas2921.1.xml}
  \end{APACrefURL}
\newblock
\APACrefnote{Publisher: American Meteorological Society Section: Journal of the
  Atmospheric Sciences}
\newblock
\begin{APACrefDOI} \doi{10.1175/2008JAS2921.1} \end{APACrefDOI}
\PrintBackRefs{\CurrentBib}

\bibitem [\protect \citeauthoryear {%
Vaswani%
\ \protect \BOthers {.}}{%
Vaswani%
\ \protect \BOthers {.}}{%
{\protect \APACyear {2017}}%
}]{%
vaswani_attention_2017}
\APACinsertmetastar {%
vaswani_attention_2017}%
\begin{APACrefauthors}%
Vaswani, A.%
, Shazeer, N.%
, Parmar, N.%
, Uszkoreit, J.%
, Jones, L.%
, Gomez, A\BPBI N.%
\BDBL {}Polosukhin, I.%
\end{APACrefauthors}%
\unskip\
\newblock
\APACrefYearMonthDay{2017}{}{}.
\newblock
{\BBOQ}\APACrefatitle {Attention is {All} you {Need}} {Attention is {All} you
  {Need}}.{\BBCQ}
\newblock
\BIn{} \APACrefbtitle {Advances in {Neural} {Information} {Processing}
  {Systems}} {Advances in {Neural} {Information} {Processing} {Systems}}\
  (\BVOL~30).
\newblock
\APACaddressPublisher{}{Curran Associates, Inc.}
\newblock
\begin{APACrefURL}
  [{2023-03-13}]\url{https://proceedings.neurips.cc/paper/2017/hash/3f5ee243547dee91fbd053c1c4a845aa-Abstract.html}
  \end{APACrefURL}
\PrintBackRefs{\CurrentBib}

\bibitem [\protect \citeauthoryear {%
Virtanen%
\ \protect \BOthers {.}}{%
Virtanen%
\ \protect \BOthers {.}}{%
{\protect \APACyear {2020}}%
}]{%
scipy_2020}
\APACinsertmetastar {%
scipy_2020}%
\begin{APACrefauthors}%
Virtanen, P.%
, Gommers, R.%
, Oliphant, T\BPBI E.%
, Haberland, M.%
, Reddy, T.%
, Cournapeau, D.%
\BDBL {}{SciPy 1.0 Contributors}%
\end{APACrefauthors}%
\unskip\
\newblock
\APACrefYearMonthDay{2020}{}{}.
\newblock
{\BBOQ}\APACrefatitle {{{SciPy} 1.0: Fundamental Algorithms for Scientific
  Computing in Python}} {{{SciPy} 1.0: Fundamental Algorithms for Scientific
  Computing in Python}}.{\BBCQ}
\newblock
\APACjournalVolNumPages{Nature Methods}{17}{}{261--272}.
\newblock
\begin{APACrefDOI} \doi{10.1038/s41592-019-0686-2} \end{APACrefDOI}
\PrintBackRefs{\CurrentBib}

\bibitem [\protect \citeauthoryear {%
Vlachas%
\ \protect \BOthers {.}}{%
Vlachas%
\ \protect \BOthers {.}}{%
{\protect \APACyear {2020}}%
}]{%
vlachas_backpropagation_2020}
\APACinsertmetastar {%
vlachas_backpropagation_2020}%
\begin{APACrefauthors}%
Vlachas, P\BPBI R.%
, Pathak, J.%
, Hunt, B\BPBI R.%
, Sapsis, T\BPBI P.%
, Girvan, M.%
, Ott, E.%
\BCBL {}\ \BBA {} Koumoutsakos, P.%
\end{APACrefauthors}%
\unskip\
\newblock
\APACrefYearMonthDay{2020}{{\APACmonth{06}}}{}.
\newblock
{\BBOQ}\APACrefatitle {Backpropagation algorithms and {Reservoir} {Computing}
  in {Recurrent} {Neural} {Networks} for the forecasting of complex
  spatiotemporal dynamics} {Backpropagation algorithms and {Reservoir}
  {Computing} in {Recurrent} {Neural} {Networks} for the forecasting of complex
  spatiotemporal dynamics}.{\BBCQ}
\newblock
\APACjournalVolNumPages{Neural Networks}{126}{}{191--217}.
\newblock
\begin{APACrefURL}
  [{2021-04-30}]\url{https://www.sciencedirect.com/science/article/pii/S0893608020300708}
  \end{APACrefURL}
\newblock
\begin{APACrefDOI} \doi{10.1016/j.neunet.2020.02.016} \end{APACrefDOI}
\PrintBackRefs{\CurrentBib}

\bibitem [\protect \citeauthoryear {%
Weyn%
, Durran%
\BCBL {}\ \BBA {} Caruana%
}{%
Weyn%
\ \protect \BOthers {.}}{%
{\protect \APACyear {2019}}%
}]{%
weyn_can_2019}
\APACinsertmetastar {%
weyn_can_2019}%
\begin{APACrefauthors}%
Weyn, J\BPBI A.%
, Durran, D\BPBI R.%
\BCBL {}\ \BBA {} Caruana, R.%
\end{APACrefauthors}%
\unskip\
\newblock
\APACrefYearMonthDay{2019}{}{}.
\newblock
{\BBOQ}\APACrefatitle {Can {Machines} {Learn} to {Predict} {Weather}? {Using}
  {Deep} {Learning} to {Predict} {Gridded} 500-{hPa} {Geopotential} {Height}
  {From} {Historical} {Weather} {Data}} {Can {Machines} {Learn} to {Predict}
  {Weather}? {Using} {Deep} {Learning} to {Predict} {Gridded} 500-{hPa}
  {Geopotential} {Height} {From} {Historical} {Weather} {Data}}.{\BBCQ}
\newblock
\APACjournalVolNumPages{Journal of Advances in Modeling Earth
  Systems}{11}{8}{2680--2693}.
\newblock
\begin{APACrefURL}
  [{2022-10-03}]\url{https://onlinelibrary.wiley.com/doi/abs/10.1029/2019MS001705}
  \end{APACrefURL}
\newblock
\APACrefnote{\_eprint:
  https://onlinelibrary.wiley.com/doi/pdf/10.1029/2019MS001705}
\newblock
\begin{APACrefDOI} \doi{10.1029/2019MS001705} \end{APACrefDOI}
\PrintBackRefs{\CurrentBib}

\bibitem [\protect \citeauthoryear {%
Weyn%
, Durran%
\BCBL {}\ \BBA {} Caruana%
}{%
Weyn%
\ \protect \BOthers {.}}{%
{\protect \APACyear {2020}}%
}]{%
weyn_improving_2020}
\APACinsertmetastar {%
weyn_improving_2020}%
\begin{APACrefauthors}%
Weyn, J\BPBI A.%
, Durran, D\BPBI R.%
\BCBL {}\ \BBA {} Caruana, R.%
\end{APACrefauthors}%
\unskip\
\newblock
\APACrefYearMonthDay{2020}{}{}.
\newblock
{\BBOQ}\APACrefatitle {Improving {Data}-{Driven} {Global} {Weather}
  {Prediction} {Using} {Deep} {Convolutional} {Neural} {Networks} on a {Cubed}
  {Sphere}} {Improving {Data}-{Driven} {Global} {Weather} {Prediction} {Using}
  {Deep} {Convolutional} {Neural} {Networks} on a {Cubed} {Sphere}}.{\BBCQ}
\newblock
\APACjournalVolNumPages{Journal of Advances in Modeling Earth
  Systems}{12}{9}{e2020MS002109}.
\newblock
\begin{APACrefURL}
  [{2022-10-03}]\url{https://onlinelibrary.wiley.com/doi/abs/10.1029/2020MS002109}
  \end{APACrefURL}
\newblock
\APACrefnote{\_eprint:
  https://onlinelibrary.wiley.com/doi/pdf/10.1029/2020MS002109}
\newblock
\begin{APACrefDOI} \doi{10.1029/2020MS002109} \end{APACrefDOI}
\PrintBackRefs{\CurrentBib}

\bibitem [\protect \citeauthoryear {%
Weyn%
, Durran%
, Caruana%
\BCBL {}\ \BBA {} Cresswell-Clay%
}{%
Weyn%
\ \protect \BOthers {.}}{%
{\protect \APACyear {2021}}%
}]{%
weyn_sub-seasonal_2021}
\APACinsertmetastar {%
weyn_sub-seasonal_2021}%
\begin{APACrefauthors}%
Weyn, J\BPBI A.%
, Durran, D\BPBI R.%
, Caruana, R.%
\BCBL {}\ \BBA {} Cresswell-Clay, N.%
\end{APACrefauthors}%
\unskip\
\newblock
\APACrefYearMonthDay{2021}{}{}.
\newblock
{\BBOQ}\APACrefatitle {Sub-{Seasonal} {Forecasting} {With} a {Large} {Ensemble}
  of {Deep}-{Learning} {Weather} {Prediction} {Models}} {Sub-{Seasonal}
  {Forecasting} {With} a {Large} {Ensemble} of {Deep}-{Learning} {Weather}
  {Prediction} {Models}}.{\BBCQ}
\newblock
\APACjournalVolNumPages{Journal of Advances in Modeling Earth
  Systems}{13}{7}{e2021MS002502}.
\newblock
\begin{APACrefURL}
  [{2022-10-03}]\url{https://onlinelibrary.wiley.com/doi/abs/10.1029/2021MS002502}
  \end{APACrefURL}
\newblock
\APACrefnote{\_eprint:
  https://onlinelibrary.wiley.com/doi/pdf/10.1029/2021MS002502}
\newblock
\begin{APACrefDOI} \doi{10.1029/2021MS002502} \end{APACrefDOI}
\PrintBackRefs{\CurrentBib}

\bibitem [\protect \citeauthoryear {%
Whiteaker%
\ \BBA {} Gerstoft%
}{%
Whiteaker%
\ \BBA {} Gerstoft%
}{%
{\protect \APACyear {2022}}%
}]{%
whiteaker_reducing_2022}
\APACinsertmetastar {%
whiteaker_reducing_2022}%
\begin{APACrefauthors}%
Whiteaker, B.%
\BCBT {}\ \BBA {} Gerstoft, P.%
\end{APACrefauthors}%
\unskip\
\newblock
\APACrefYearMonthDay{2022}{{\APACmonth{07}}}{}.
\newblock
{\BBOQ}\APACrefatitle {Reducing echo state network size with controllability
  matrices} {Reducing echo state network size with controllability
  matrices}.{\BBCQ}
\newblock
\APACjournalVolNumPages{Chaos: An Interdisciplinary Journal of Nonlinear
  Science}{32}{7}{073116}.
\newblock
\begin{APACrefURL}
  [{2022-07-19}]\url{https://aip.scitation.org/doi/full/10.1063/5.0071926}
  \end{APACrefURL}
\newblock
\APACrefnote{Publisher: American Institute of Physics}
\newblock
\begin{APACrefDOI} \doi{10.1063/5.0071926} \end{APACrefDOI}
\PrintBackRefs{\CurrentBib}

\bibitem [\protect \citeauthoryear {%
Xu%
, Zhang%
\BCBL {}\ \BBA {} Luo%
}{%
Xu%
\ \protect \BOthers {.}}{%
{\protect \APACyear {2022}}%
}]{%
xu_overview_2022}
\APACinsertmetastar {%
xu_overview_2022}%
\begin{APACrefauthors}%
Xu, Z\BHBI Q\BPBI J.%
, Zhang, Y.%
\BCBL {}\ \BBA {} Luo, T.%
\end{APACrefauthors}%
\unskip\
\newblock
\APACrefYearMonthDay{2022}{{\APACmonth{01}}}{}.
\newblock
\APACrefbtitle {Overview frequency principle/spectral bias in deep learning.}
  {Overview frequency principle/spectral bias in deep learning.}
\newblock
\APACaddressPublisher{}{arXiv}.
\newblock
\begin{APACrefURL} [{2022-08-19}]\url{http://arxiv.org/abs/2201.07395}
  \end{APACrefURL}
\newblock
\APACrefnote{arXiv:2201.07395 [cs]}
\PrintBackRefs{\CurrentBib}

\bibitem [\protect \citeauthoryear {%
Zhang%
\ \BBA {} Cornelius%
}{%
Zhang%
\ \BBA {} Cornelius%
}{%
{\protect \APACyear {2022}}%
}]{%
zhang_catch-22_2022}
\APACinsertmetastar {%
zhang_catch-22_2022}%
\begin{APACrefauthors}%
Zhang, Y.%
\BCBT {}\ \BBA {} Cornelius, S\BPBI P.%
\end{APACrefauthors}%
\unskip\
\newblock
\APACrefYearMonthDay{2022}{{\APACmonth{10}}}{}.
\newblock
\APACrefbtitle {A {Catch}-22 of {Reservoir} {Computing}.} {A {Catch}-22 of
  {Reservoir} {Computing}.}
\newblock
\APACaddressPublisher{}{arXiv}.
\newblock
\begin{APACrefURL} [{2022-10-24}]\url{http://arxiv.org/abs/2210.10211}
  \end{APACrefURL}
\newblock
\APACrefnote{arXiv:2210.10211 [nlin]}
\PrintBackRefs{\CurrentBib}

\end{thebibliography}
\end{document}